\documentclass[letterpaper, 10 pt, conference]{ieeeconf}
\IEEEoverridecommandlockouts
\makeatletter
\let\NAT@parse\undefined
\makeatother
\usepackage[numbers,sort&compress]{natbib}
\usepackage{xspace}
\usepackage[dvipsnames]{xcolor}
\usepackage{graphicx}
\usepackage{amsmath}
\usepackage{amssymb}
\usepackage{booktabs}
\usepackage[hyphens]{url}
\usepackage[font=small]{caption}
\usepackage[font=footnotesize,skip=3pt,subrefformat=parens]{subcaption}

\usepackage[T1]{fontenc}

\usepackage{rpm_SIunits}
\usepackage{rpm_acronyms}
\usepackage{rpm_math}
\usepackage{rpm_misc}

\usepackage{multirow}
\usepackage{colortbl}
\usepackage{tabularx}
\definecolor{tabfirst}{RGB}{255,179,179}
\definecolor{tabsecond}{RGB}{255,223,179}
\newcommand{\cfirst}{\cellcolor{tabfirst}}
\newcommand{\csecond}{\cellcolor{tabsecond}}
\definecolor{dettight}{HTML}{2CA02C}
\definecolor{detloose}{HTML}{F39C12}
\definecolor{detmiss}{HTML}{D62728}
\usepackage{mathtools}
\acrodef{HCS}{hybrid cylindrical--spherical}
\acrodef{VAE}{variational autoencoder}
\acrodef{DiT}{diffusion transformer}
\acrodef{HD}{high-definition}
\acrodef{BEV}{bird's-eye view}
\acrodef{MLP}{multilayer perceptron}

\renewcommand{\tabref}[1]{Table~\ref{#1}}

\AtBeginDocument{\acused{LiDAR}}

\newcommand{\ours}{\textsc{Ours}\xspace}
\newcommand{\rayset}{\mathcal{R}}
\newcommand{\dd}[1]{\kern0.8pt{\color{black!78}\fontsize{5}{5.5}\selectfont(#1\%)}}
\usepackage{afterpage}
\definecolor{linkblue}{rgb}{0.21,0.49,0.74}
\usepackage[breaklinks,colorlinks,allcolors=linkblue]{hyperref}

\newcommand{\papertitle}{One Latent World for Heterogeneous LiDARs:\\
A Scan Pattern Agnostic 4D World Model via Direct Ray Decoding}
\newcommand{\paperauthors}{Minwoo Jung$^{1}$, Dongjae Lee$^{1}$, DongGuw Lee$^{1}$, Yeong Sang Park$^{2}$, Maurice Fallon$^{3}$, and Ayoung Kim$^{1}$}
\newcommand{\projecturl}{\href{https://minwoo0611.github.io/OneLatentWorld}{\texttt{https://minwoo0611.github.io/OneLatentWorld}}}

\title{\LARGE \bf \papertitle}
\author{\paperauthors%
\thanks{$^{1}$Minwoo Jung, Dongjae Lee, DongGuw Lee, and Ayoung Kim are with Seoul National University, Seoul, South Korea.}%
\thanks{$^{2}$Yeong Sang Park is with the Korea Institute of Machinery \& Materials, South Korea.}%
\thanks{$^{3}$Maurice Fallon is with the Oxford Robotics Institute, University of Oxford, UK.}%
}

\newcommand{\maketitlesupp}{%
  \twocolumn[{%
    \centering
    {\LARGE\bfseries \papertitle\par}
    \vspace{0.7em}{\Large Supplementary Material\par}
    \vspace{0.9em}{\large \paperauthors\par}
    \vspace{0.3em}{\small $^{1}$Seoul National University\quad $^{2}$Korea Institute of Machinery \& Materials\quad $^{3}$University of Oxford\par}
    \vspace{0.3em}{\small Project page: \projecturl\par}
    \vspace{1.4em}
  }]}

\begin{document}
\maketitle

\begin{abstract}
LiDAR world models generate and predict metric 3D observations in the form of LiDAR scans.
They typically synthesize scans following the scan pattern of their training sensor or render them from an intermediate 3D surface or field.
However, following one scan pattern limits transfer to LiDARs with other patterns, and rendering from intermediate geometry can cause incorrect depths and false returns.
In this paper, we propose a scan pattern agnostic 4D LiDAR world model that decouples the scene from sensor observation, so one spatial latent can be observed by heterogeneous LiDARs at the same mounting position without regeneration.
Our decoder predicts the hit probability and depth of each ray directly from the latent, combining global scene context with local features retrieved at a preliminary depth.
Without an intermediate surface or field, it reduces incorrect depths and false returns, and as each ray is decoded independently from one latent, LiDARs that share a ray receive the same return and observe one consistent scene.
Finally, a diffusion transformer generates scenes in this latent and predicts 4D sequences from observed latents warped by the given ego and object motion.
On KITTI-360 and nuScenes, our model generates realistic scenes and reduces the registration error between predicted frames by about 70\% relative to the best compared world model.
Its decoder yields 19 to 28\% lower depth error than the rendering used by other world models, and generated scans benefit several downstream tasks.
Our code will be open-sourced.
More details are available at \projecturl.
\end{abstract}
\section{Introduction}
\label{sec:intro}

\begin{figure}[t]
    \centering
    \includegraphics[width=\linewidth]{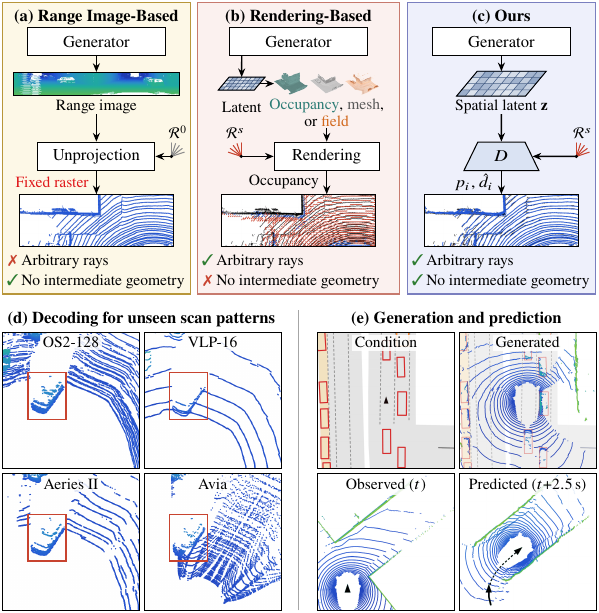}
    \caption{\textbf{One latent world for heterogeneous LiDARs.}
    (a)--(c) Scans produced on a measured KITTI-360 scene for training rays $\mathcal{R}^0$ or target rays $\mathcal{R}^s$.
    Measured points are black. Produced points are blue, or red if they lie over 0.5\,m in front of the measured surface.
    (d) One KITTI-360 latent decoded for four unseen HeLiPR~\cite{jung2024helipr} sensors, with boxes on the same car.
    (e) A nuScenes scene generated from 3D boxes and an HD map, and a prediction 2.5\,s ahead under the recorded ego and object motion.}
    \label{fig:teaser}
    \vspace{-6mm}
\end{figure}

World models learn representations that predict how an environment evolves under actions~\cite{lecun2022path}.
In autonomous driving, they simulate scenes and their evolution, allowing scenarios to be revisited without collecting new data~\cite{kong2025worldmodeling}.
Driving world models have been built on video~\cite{ni2025maskgwm}, semantic occupancy~\cite{zheng2023occworld}, and \ac{LiDAR} scans~\cite{zhang2024copilot4d,zyrianov2025lidardm,opendwm,che2026lidarverse}.
Among them, \ac{LiDAR} world models generate metric observations in the scan format that perception and localization modules use.

Since autonomous platforms carry \acp{LiDAR} with different scan patterns, from rotating to solid-state sensors~\cite{jung2024helipr}, a \ac{LiDAR} world model should be able to produce observations for each pattern.
A representative approach to \ac{LiDAR} scene generation uses range images~\cite{zyrianov2022lidargen,nakashima2024lidar,ran2024lidm,hu2024rangeldm} and produces ranges on the angular raster of the training sensor, as illustrated in \figref{fig:teaser}a.
However, this ties the representation to that scan pattern and makes non-repetitive scans hard to handle.
Range imaging at variable resolutions~\cite{nakashima2023generative} relaxes the fixed raster, but its multiscale processing depends on the requested grid. A ray shared by different grids can then have different ranges.

One way to avoid this dependence is to represent the scene in space and render observations along the requested rays, as illustrated in \figref{fig:teaser}b.
Several world models take this approach and decode a generated latent into meshes, semantic occupancy, or an implicit occupancy field~\cite{zyrianov2025lidardm,li2025uniscene,zhang2024copilot4d,opendwm}.
Rendering supports flexible scan patterns, and learned fields can be fitted to sensor measurements, but each range is still read from the intermediate geometry along the ray.
Spurious or coarsely discretized geometry in front of the visible surface pulls the range toward the sensor, and geometry on a ray without a measured return creates a false return.

These limitations motivate separating scene generation from sensor observation without an intermediate geometry step.
This requires a scene representation independent of the target scan pattern, from which the return and depth of each ray can be predicted directly.
Spatial latents such as \ac{BEV} and \ac{HCS}~\cite{xiong2023ultralidar,zhang2024copilot4d,liu2025listar} suit this role by placing heterogeneous scans in shared spatial coordinates.
However, for efficient generation, such latents use coarse grids and are often encoded from binary occupancy~\cite{xiong2023ultralidar,opendwm,liu2025listar}, which discards where points lie within each cell.
Both factors limit the geometric detail needed for accurate depth along each ray.

To tackle these issues, we present a \ac{LiDAR} world model built on a spatial latent decoupled from scan patterns, outlined in \figref{fig:teaser}c.
As a shared encoder maps scans of different patterns into this latent, the autoencoder learns from all of them.
To retain detail within each cell, the encoder input also stores the point count and the mean radial offset of the points.
Since only the decoder receives the rays of the target scan pattern, one generated scene can be observed by different \acp{LiDAR} (\figref{fig:teaser}d) without regeneration.
Furthermore, unlike rendering, our decoder predicts the hit probability and depth of each ray directly from the latent and learns from labeled hits and misses to reduce false returns.
Finally, a \ac{DiT} generates scenes in this latent with or without layouts and predicts future scenes from observed latents warped by the given motion (\figref{fig:teaser}e), so different \acp{LiDAR} can observe these 4D sequences.
Our contributions are as follows:

\begin{itemize}
    \item \textbf{Scan pattern agnostic LiDAR world modeling.} Our world model separates scene generation and prediction from scan patterns, enabling a single latent scene or sequence to be observed by heterogeneous LiDARs at the same mounting position without regeneration.
    \item \textbf{Direct latent-to-ray decoding.} A depth-guided ray decoder combines global scene context with local features retrieved at a preliminary depth to predict ray returns directly, reducing false returns and depth errors without reconstructing an intermediate surface or field.
    \item \textbf{Temporally consistent scene prediction.} The motion-aligned latent conditioning warps the observed latents to the predicted frames under given ego and object motion, which improves temporal consistency.
    \item \textbf{Generation quality and downstream utility.} Experiments on KITTI-360 and nuScenes show strong generation quality, 66--72\% lower registration errors between predicted frames than the compared world models, and 19--28\% lower depth errors than their rendering. Predicted scans for unseen sensors improve place recognition, moving object segmentation, and object detection.
\end{itemize}
\begin{figure*}[t]
    \centering
    \includegraphics[width=0.92\textwidth]{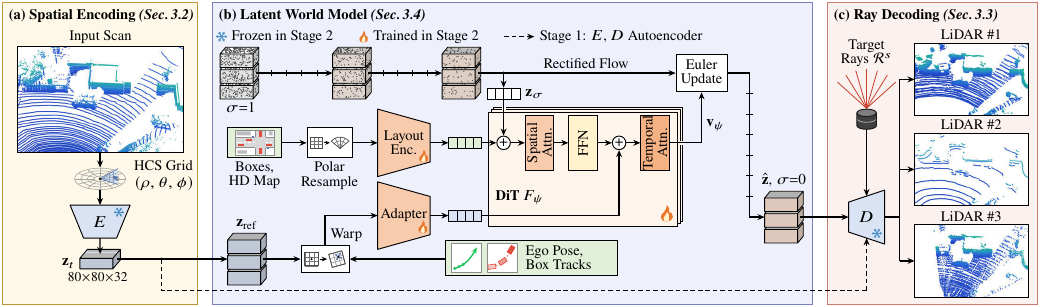}
    \caption{\textbf{Overview.}
    (a) $E$ maps a scan on the \ac{HCS} grid to a latent on the $(\rho,\theta)$ plane.
    (b) The \ac{DiT} $F_\psi$ receives layouts and, for prediction, reference latents warped by the given motion. Ticks mark the other Euler steps.
    (c) $D$ decodes the latent along target rays $\mathcal{R}^s$.}
    \label{fig:pipeline}
    \vspace{-5mm}
\end{figure*}

\section{Related Work}
\label{sec:relatedwork}

\noindent\textbf{Range image generation.}
Most LiDAR generative models represent scans as range images and apply diffusion directly to scans~\cite{zyrianov2022lidargen,nakashima2024lidar} or in an autoencoder's latent space~\cite{ran2024lidm,hu2024rangeldm}. Subsequent work improves sampling efficiency and quality~\cite{nakashima2025r2flow,liu2026l3dr,sereyjol2026r3dpa} and adds conditions such as semantic maps~\cite{ran2024lidm} or text~\cite{qu2026t2ldm,qu2026t2ldmpp}, while keeping the angular raster of the training sensor. Generative Range Imaging~\cite{nakashima2023generative} supports variable angular resolutions, but its predictions for a shared ray can change with the grid. As our model decodes each requested ray independently from a spatial latent, its scans are not tied to a raster, and a ray common to two scan patterns receives the same prediction.

\noindent\textbf{Observation from spatial representations.}
Spatial representations avoid angular indexing and enable flexible scan patterns. UltraLiDAR~\cite{xiong2023ultralidar} outputs grid-quantized points at occupied voxels without modeling scan patterns or ray returns, whereas simulators and world models render ranges along rays from meshes, occupancy volumes, or neural fields~\cite{manivasagam2020lidarsim,huang2023nfl,tao2024lidarnerf,zyrianov2025lidardm,li2025uniscene,zhang2024copilot4d,opendwm,zhang2024bevworld}, with several also modeling ray drop. Spurious intermediate geometry in front of the visible surface can bias these ranges toward the sensor. Predictors conditioned on rays estimate depth~\cite{guizilini2022depth} or refine geometric intersection distances~\cite{dai2025learning}. Neither these predictors nor the render heads of Copilot4D~\cite{zhang2024copilot4d} and OpenDWM~\cite{opendwm} are trained with an explicit loss on rays without a measured return. Occupancy labels instead supervise empty space everywhere outside fused surfaces~\cite{wei2023surroundocc} or along measured rays~\cite{tian2023occ3d,agro2024uno}. Building on prediction along rays, our decoder learns from labeled hits and misses and directly estimates the hit probability and depth of each ray from a spatial latent using depth-guided feature retrieval, without reconstructing an intermediate surface or field.

\noindent\textbf{4D LiDAR world models.}
Recent world models extend generation to temporal prediction through spatial token modeling~\cite{zhang2024copilot4d,opendwm}, occupancy forecasting~\cite{zheng2023occworld}, and motion-guided range image generation~\cite{liang2025lidarcrafter}. Other advances address uncertainty~\cite{xu2026u4d}, static--dynamic decomposition~\cite{wu2026gem}, and scan rendering from a reconstructed volume~\cite{liu2025listar}. LidarVerse~\cite{che2026lidarverse} generates range images of several beam configurations with one model, each chosen by a sensor condition. Thus, each configuration is a separate sample that need not depict the same scene, and its fixed raster cannot follow a non-repetitive pattern. In our model, only the decoder receives the target rays, so even non-repetitive \acp{LiDAR} observe the same latent sequence without regeneration.
\section{Method}
\label{sec:method}

\subsection{Overview}
\figref{fig:pipeline} gives an overview of our model, which separates the scene from the way a sensor observes it.
Given a \ac{LiDAR} point cloud $\mathbf{P}_t$, a shared encoder computes its spatial latent $\mathbf{z}_t=E(\mathbf{P}_t)$.
A latent world model built on a \ac{DiT} $F_\psi$ samples scene latents from noise, optionally conditioned on layouts $\mathbf{c}$ of 3D boxes and \ac{HD} maps.
For prediction, it samples a future latent $\hat{\mathbf{z}}_{t+1}$ conditioned on reference latents of past frames warped by the ego and object motion.
A decoder $D$ then decodes the scan of a target \ac{LiDAR},
\begin{equation}
    \hat{\mathbf{P}}^{s}_{t+1} = D\bigl(\hat{\mathbf{z}}_{t+1},\,\rayset^{s}\bigr),
    \label{eq:overview}
\end{equation}
where $\rayset^{s}=\{\mathbf{r}_i\}_{i=1}^{M_s}$ contains the unit directions of the $M_s$ rays of its scan pattern $s$, which all start at the sensor origin~$\mathbf{0}$.
Because only the decoder receives $\rayset^{s}$, the latent and the world model do not depend on the target scan pattern.

\subsection{Shared Spatial Encoding}
\label{sec:encoder}
Point clouds from different types of \acp{LiDAR} differ in scan pattern, coverage, and density.
To place them on a common grid, we voxelize every scan on an \ac{HCS} grid~\cite{chen2020every,liu2025listar} whose axes are the horizontal radius $\rho=\sqrt{x^2+y^2}$, the azimuth $\theta=\operatorname{atan2}(y,x)$, and the elevation $\phi=\operatorname{atan2}(z,\rho)$.
Each cell stores its binary occupancy, a capped and normalized log point count, and the mean radial offset of its points from the inner cell boundary, normalized by the radial size of the cell.
The count and offset retain how densely a cell was sampled and where its points lie in range, which occupancy alone discards.

The encoder maps this grid to a latent that keeps radius and azimuth as spatial axes and stores elevation in its channels. Elevation is compressed rather than averaged away.
The encoder groups neighboring elevation bins into bands and folds each band into channels. A convolution within each band lowers the azimuth resolution to keep the latent compact.
A $1\times1$ convolution then combines the bands, and shifted-window attention aggregates neighboring features.
For context beyond the windows, the encoder averages the features over all azimuths in each ring of neighboring radial rows and adds a gated transform of this average back to every cell of that ring.
These features define a diagonal Gaussian posterior~\cite{kingma2014autoencoding} over $\mathbf{z}\in\mathbb{R}^{H_z\times W_z\times C_z}$.
We sample $\mathbf{z}$ in autoencoder training and otherwise use its mean.
Since radius and azimuth remain spatial axes, each latent position indexes a fixed region around the sensor, not an input beam.
Appendix~\ref{sec:app_model} details the encoder.

\subsection{Depth-Guided Latent-to-Ray Decoding}
\label{sec:decoder}
\begin{figure}[t]
    \centering
    \includegraphics[width=0.88\linewidth]{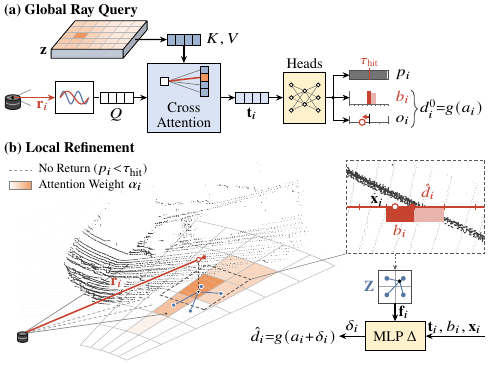}
    \caption{\textbf{Depth-guided ray decoding.}
    (a) A ray attends to all latent tokens and predicts a hit probability $p_i$, a bin $b_i$, and an offset $o_i$. The bin and offset give a preliminary depth $d^0_i$.
    (b) Features sampled from $\mathbf{Z}$ around $\mathbf{x}_i$ refine $d^0_i$ to $\hat{d}_i$.
    The colored grid illustrates the attention weights $\alpha_i$.}
    \label{fig:decoder}
    \vspace{-5mm}
\end{figure}

As illustrated in \figref{fig:decoder}, the decoder first estimates the hit probability and a preliminary depth of each ray from the whole scene and then refines the depth with local features retrieved at that depth.
For the global depth estimation, an embedding of the direction $\mathbf{r}_i$ queries the latent tokens through cross-attention with weights $\alpha_i$, as in direct depth prediction~\cite{guizilini2022depth}.
The resulting ray token $\mathbf{t}_i$ predicts a hit probability $p_i$, a distribution over $B$ depth bins spaced uniformly in log depth, and a continuous offset $o_i$.

To obtain a continuous preliminary depth, we select the most probable bin $b_i\in\{0,\ldots,B-1\}$.
We add the offset to the logit of its normalized center, $a_i=o_i+\operatorname{logit}\bigl((b_i+\tfrac12)/B\bigr)$, and map the result to $d^0_i=g(a_i)$.
The map $g(a)=\exp\bigl(\ell_{\min}+\operatorname{sigm}(a)\,\Delta\ell\bigr)$ covers depths strictly between $d_{\min}$ and $d_{\max}$ on a log scale, with $\ell_{\min}=\log d_{\min}$ and $\Delta\ell=\log d_{\max}-\ell_{\min}$.
The local query point $\mathbf{x}_i=d^0_i\,\mathbf{r}_i$ uses this continuous depth rather than the bin center.

For the local refinement, a $1\times1$ convolution maps $\mathbf{z}$ to a $C_f$-channel feature map $\mathbf{Z}$ on the same grid.
We sample it at the query point as $\mathbf{f}_i=\operatorname{Bilinear}\bigl(\mathbf{Z},\pi(\mathbf{x}_i)\bigr)$, where $\pi$ maps a point to its $(\rho,\theta)$ coordinates with wrap-around in azimuth.
Since $\mathbf{Z}$ has no elevation axis, the channels of $\mathbf{f}_i$ cover all elevations, and $\mathbf{t}_i$ and $\mathbf{x}_i$ supply the elevation of the ray.
A small \ac{MLP} $\Delta$ then predicts a correction $\delta_i=\Delta(\mathbf{t}_i,\mathbf{f}_i,b_i,\mathbf{x}_i)$.
The final depth is $\hat{d}_i=g(a_i+\delta_i)$.
Since $\delta_i$ is added before $g$, $\hat{d}_i$ can move beyond the selected bin.
A ray with $p_i$ above the hit threshold $\tau_{\mathrm{hit}}$ returns the point $\hat{d}_i\mathbf{r}_i$. The decoder outputs points without an intermediate surface or field.

\noindent\textbf{Consistency across query sets.}
Each ray's attention is normalized over the latent tokens, and every later operation acts on that ray alone without modifying $\mathbf{Z}$.
For a fixed latent, a ray's prediction depends only on its direction.
In exact arithmetic, it does not change when other rays are added, removed, or reordered, as Appendix~\ref{sec:app_rays_consist} confirms.

\subsection{Latent Scene Generation and Prediction}
\label{sec:worldmodel}
With the target scan pattern left to the decoder, scene generation and prediction operate entirely on the spatial latents.
A conditional \ac{DiT}~\cite{peebles2023scalable} $F_\psi$ splits each latent frame into patch tokens.
Each block applies spatial attention and a feed-forward network, adds motion-aligned reference features for prediction, and then applies temporal attention across frames.
Scene generation starts from noise without reference latents.
For prediction, $N_{\mathrm{ref}}$ observed frames provide clean reference latents $\mathbf{z}_{\mathrm{ref}}$, and the ego poses and object boxes of all frames are given.

To align layout conditions with the latent, the boxes and map elements of each frame are drawn as \ac{BEV} rasters and resampled onto a $(\rho,\theta)$ grid.
A small convolutional network encodes these rasters, and zero-initialized projections add its outputs to the tokens of every block.
Box and map conditions are dropped individually or together during training. Since some generation samples drop both, one model supports conditional and unconditional generation.

For prediction, ego and object motion shift scene content across the $(\rho,\theta)$ grid.
LiDARCrafter~\cite{liang2025lidarcrafter} moves points back-projected from its generated range images and reprojects them into a range image for the next frame. Our model instead warps the reference latents directly, without decoding them.
Because each latent cell has known horizontal coordinates, the warp aligns the latents with each target frame in radius and azimuth.
For this warp, each target cell is represented by the point $\mathbf{x}_{\rho\theta}=[\rho\cos\theta,\ \rho\sin\theta,\ 0]^\top$ at the sensor height.
This point is moved into the reference frame, where the reference latent is sampled bilinearly,
\begin{equation}
    \tilde{\mathbf{z}}_{\mathrm{ref}}(\rho,\theta)=\operatorname{Bilinear}\bigl(\mathbf{z}_{\mathrm{ref}},\,\pi(\mathbf{T}\,\mathbf{x}_{\rho\theta})\bigr).
    \label{eq:warp}
\end{equation}
Here $\mathbf{T}$ is the relative ego pose from the target to the reference frame.
Inside a target box matched to a reference box by track identity, $\mathbf{T}$ is replaced by the box motion $\mathbf{B}_{\mathrm{ref}}\mathbf{B}^{-1}$, letting object features follow the object.
The maps $\mathbf{B}$ and $\mathbf{B}_{\mathrm{ref}}$ take horizontal box coordinates, normalized by the box size, to the target and reference frames.
To avoid copying object features into the background, ego-warped features whose source lies inside a reference box are excluded.
The remaining aligned features are averaged over reference frames with weights decaying exponentially with the frame gap.
A zero-initialized adapter adds this average to the target tokens in every block.
Appendix~\ref{sec:app_model} illustrates this warp.

To generate latents under these conditions, we train $F_\psi$ with rectified flow~\cite{liu2023flow,lipman2023flow}.
At noise level $\sigma\in[0,1]$, a target latent is corrupted as $\mathbf{z}_\sigma=(1-\sigma)\mathbf{z}+\sigma\boldsymbol{\epsilon}$ with $\boldsymbol{\epsilon}\sim\mathcal{N}(\mathbf{0},\mathbf{I})$.
$F_\psi$ predicts the velocity $\mathbf{v}_\psi$, which gives the clean estimate $\hat{\mathbf{z}}_0=\mathbf{z}_\sigma-\sigma\,\mathbf{v}_\psi(\mathbf{z}_\sigma,\sigma;\mathbf{z}_{\mathrm{ref}},\mathbf{c})$.
At inference, target latents are integrated with Euler steps from Gaussian noise at $\sigma=1$ to $\sigma=0$, while the reference latents stay clean.

\subsection{Training}
\label{sec:training}
\noindent\textbf{Training scans.}
A single scan supervises the decoder only along the rays of its own sensor.
To supervise other directions as well, we add aggregated scans, which combine the target scan with the static points seen from other poses.
Aggregated scans add samples between the sensor's beams and can extend the angular coverage, while single scans keep the sparsity and occlusion of measured observations.
Each supervised ray is labeled with a return depth or a miss.
In an aggregated scan, the fused points are projected into fine angular cells around the target sensor, each defining a ray direction, and the nearest point in each cell gives a hit and its depth.
Cells without a hit may still hold unobserved surfaces.
Such cells are labeled misses when the fused scans are dense enough to rule this out or other evidence shows them to be empty, such as a ray of the target single scan in the same direction without a return.
Other cells are not used in training.
As a scan pattern augmentation, half of the aggregated scans are observed with a random pattern, never one taken from an existing sensor.
The encoder then receives only the hits along this pattern, and the decoder is supervised on the labeled cells along it.
Otherwise, the encoder receives all fused hits.
Appendix~\ref{sec:app_setup} explains the labeling rules and the pattern distribution in detail.

\noindent\textbf{Autoencoder objective.}
Hit prediction is trained with a focal loss $\mathcal{L}_{\mathrm{hit}}$~\cite{lin2017focal} on labeled hits and misses, and the depth bin of each hit with a cross-entropy $\mathcal{L}_{\mathrm{bin}}$.
The depth loss $\mathcal{L}_{\mathrm{depth}}$ applies a Smooth-$\ell_1$ penalty to the metric error $\hat{d}_i-d^\star_i$ of each hit with measured depth $d^\star_i$.
To train the refinement independently of errors in the selected bin, $\mathcal{L}_{\mathrm{depth}}$ also includes a term on the depth refined from the ground-truth bin, penalizing its log-depth error normalized by the bin width.
With a KL term $\mathcal{L}_{\mathrm{KL}}$ on the posterior, the objective is $\mathcal{L}_{\mathrm{AE}}=\lambda_h\mathcal{L}_{\mathrm{hit}}+\lambda_b\mathcal{L}_{\mathrm{bin}}+\lambda_d\mathcal{L}_{\mathrm{depth}}+\lambda_{\mathrm{KL}}\mathcal{L}_{\mathrm{KL}}$.

\noindent\textbf{World model objective.}
After $E$ and $D$ are trained, both are frozen, and $E$ provides the target latents $\mathbf{z}$.
All target frames of a sequence share one noise level, and reference frames stay clean.
The main loss is $\mathcal{L}_{\mathrm{WM}}=\mathbb{E}\bigl[\|\hat{\mathbf{z}}_0-\mathbf{z}\|_2^2\bigr]$ over the target frames.
Because a small latent error does not guarantee a small error in the decoded scan, we also supervise decoded outputs at low noise levels, $\sigma\leq\sigma_r$.
The frozen decoder decodes $\hat{\mathbf{z}}_0$ and $\mathbf{z}$ along the same rays, and the predictions from $\mathbf{z}$ serve as fixed targets.
The loss $\mathcal{L}_{\mathrm{ray}}$ matches the hit probabilities and depth bin distributions decoded from $\hat{\mathbf{z}}_0$ to these targets with cross-entropy, and the final objective is $\mathcal{L}_{\mathrm{WM}}+\lambda_r\mathcal{L}_{\mathrm{ray}}$.
\section{Experiments}
\label{sec:evaluation}

\subsection{Experimental Setup}
\label{sec:setup}

\noindent\textbf{Datasets.}
Training uses the same KITTI-360~\cite{liao2022kitti360} sequences as prior work~\cite{zyrianov2022lidargen,nakashima2025r2flow,opendwm} and the nuScenes~\cite{caesar2020nuscenes} training split.
Unlike ours, some compared methods train on sequence 0000 or 0002, so Tables~\ref{tab:uncond_generation} and~\ref{tab:vae_fidelity} use sequence 0018, which no compared method trains on.
The other KITTI-360 tables use sequences 0000 and 0002 with occupancy labels and 3D boxes, and nuScenes uses its validation split.
Appendix~\ref{sec:app_baselines} lists each method's training data.
\afterpage{\begin{table}[t]
    \centering
    \caption{Scene generation without any condition. The 3D MMD and the BEV JSD are in $10^{-2}$, and the BEV MMD is in $10^{-3}$. Time is the generation time per scan in seconds on one RTX 3090.}
    \label{tab:uncond_generation}
    \scriptsize
    \setlength{\tabcolsep}{0.8pt}
    \renewcommand{\arraystretch}{0.94}
    \begin{tabularx}{\linewidth}{@{}l*{6}{>{\centering\arraybackslash}X}@{}}
        \toprule
        & \multicolumn{3}{c}{3D} & \multicolumn{2}{c}{BEV} & \\
        \cmidrule(lr){2-4}\cmidrule(lr){5-6}
        Method & FSVD $\downarrow$ & FPVD $\downarrow$ & MMD $\downarrow$ & JSD $\downarrow$ & MMD $\downarrow$ & Time $\downarrow$ \\
        \midrule
        \multicolumn{7}{@{}l}{\textbf{KITTI-360} (sequence 0018, 2,000 scans per method)} \\
        LiDARGen~\cite{zyrianov2022lidargen} & 83.53 & 70.63 & 24.21 & 8.32 & 0.92 & 90.67 \\
        R2DM~\cite{nakashima2024lidar} & 32.90 & 31.13 & 9.18 & 8.56 & 0.90 & 3.92 \\
        LiDM~\cite{ran2024lidm} & 33.61 & 37.85 & 11.09 & 21.02 & 4.82 & \cfirst\textbf{0.76} \\
        R2Flow~\cite{nakashima2025r2flow} & 40.24 & 37.31 & 11.65 & \csecond 7.96 & \csecond 0.74 & 5.41 \\
        LidarDM~\cite{zyrianov2025lidardm} & 70.77 & 55.56 & 12.18 & 18.14 & 4.49 & 16.68 \\
        OpenDWM-DiT~\cite{opendwm} & 150.80 & 133.79 & 43.00 & 16.46 & 3.17 & 1.69 \\
        R3DPA~\cite{sereyjol2026r3dpa} & 32.95 & 32.14 & 9.48 & 8.50 & 0.82 & 1.40 \\
        T2LDM~\cite{qu2026t2ldm} & \cfirst\textbf{24.62} & 23.20 & 6.40 & 8.31 & 0.88 & 63.99 \\
        T2LDM++~\cite{qu2026t2ldmpp} & \csecond 25.35 & \csecond 22.89 & \csecond 6.22 & 8.15 & 0.81 & 59.60 \\
        \ours & 28.12 & \cfirst\textbf{17.78} & \cfirst\textbf{3.41} & \cfirst\textbf{7.66} & \cfirst\textbf{0.69} & \csecond 0.94 \\
        \midrule
        \multicolumn{7}{@{}l}{\textbf{nuScenes} (6,019 scans per method)} \\
        OpenDWM-DiT~\cite{opendwm} & 93.30 & 81.01 & 25.06 & 13.10 & 2.28 & 1.69 \\
        UniScene~\cite{li2025uniscene} & 56.33 & 49.79 & \csecond 13.45 & 15.81 & 3.51 & \csecond 1.37 \\
        T2LDM~\cite{qu2026t2ldm} & \csecond 43.49 & \csecond 48.43 & 15.66 & 9.07 & 1.93 & 31.75 \\
        T2LDM++~\cite{qu2026t2ldmpp} & 44.14 & 49.53 & 16.04 & \csecond 8.94 & \csecond 1.90 & 28.96 \\
        \ours & \cfirst\textbf{12.06} & \cfirst\textbf{10.57} & \cfirst\textbf{1.68} & \cfirst\textbf{5.09} & \cfirst\textbf{0.11} & \cfirst\textbf{0.86} \\
        \bottomrule
    \end{tabularx}
    \vspace{-4.5mm}
\end{table}}%

\noindent\textbf{Metrics.}
For scene generation, FSVD, FPVD~\cite{ran2024lidm}, and MMD~\cite{liang2025lidarcrafter} compare the feature distributions of generated and measured scans, and JSD and MMD compare their \ac{BEV} histograms of point counts.
We also measure the time to generate one scan on one RTX 3090.
Following LiDARCrafter~\cite{liang2025lidarcrafter}, object generation compares car crops by FPD~\cite{shu2019treegan} and MMD, scores detected cars by detector confidence (FDC), and scores conditioned objects by 3D detection AP (CDA$_{\mathrm{3D}}$) and by class and box recovery (CFCA, CFSC).
Sequences are scored by the translation error after registration (TTCE) and the cross-frame point distance (CTC), adapted from LiDARCrafter~\cite{liang2025lidarcrafter}.
Reconstruction uses the Chamfer distance (CD) and the F-score at 0.1\,m ($F_{0.1}$).
The mean, median, root-mean-square, and relative depth errors (dMAE, MedAE, RMSE, AbsRel) are measured on range image cells where both scans return, as in other LiDAR world models~\cite{khurana2023point,zhang2024copilot4d}.
Distances are in meters, AbsRel is in percent, and Appendix~\ref{sec:app_metrics} details the metrics.

\noindent\textbf{Baselines.}
We compare with the range image models LiDARGen~\cite{zyrianov2022lidargen}, R2DM~\cite{nakashima2024lidar}, LiDM~\cite{ran2024lidm}, R2Flow~\cite{nakashima2025r2flow}, R3DPA~\cite{sereyjol2026r3dpa}, T2LDM~\cite{qu2026t2ldm}, T2LDM++~\cite{qu2026t2ldmpp}, and GEM~\cite{wu2026gem}, and the rendering-based models LidarDM~\cite{zyrianov2025lidardm}, OpenDWM~\cite{opendwm}, and UniScene~\cite{li2025uniscene}.
For OpenDWM, its DiT and MaskGIT models use voxel heads, and Sec.~\ref{sec:transfer} also tests its render head.
All baselines use their released weights and are scored with the same evaluator.

\noindent\textbf{Implementation details.}
The latent of size $H_z\times W_z\times C_z=80\times80\times32$ spans an \ac{HCS} grid up to 80\,m.
One autoencoder is trained jointly on both datasets for 160k steps on four RTX A6000 GPUs.
The world model is trained for 150k steps on two H200 GPUs and predicts five frames from $N_{\mathrm{ref}}=3$ observed frames.
Sampling uses 10 Euler steps.
Further details are given in Appendices~\ref{sec:app_setup} and~\ref{sec:app_model}.

\subsection{Scene Generation and Prediction}
\label{sec:generation}
\label{sec:cond_generation}
\label{sec:prediction}
\afterpage{\begin{table}[t]
    \centering
    \caption{Object generation and box agreement. MMD is in $10^{-3}$.}
    \label{tab:object_generation}
    \scriptsize
    \setlength{\tabcolsep}{1.6pt}
    \renewcommand{\arraystretch}{0.94}
    \resizebox{\linewidth}{!}{\begin{tabular}{@{}lcccccc@{}}
        \toprule
        & \multicolumn{2}{c}{Object} & \multicolumn{2}{c}{Detection} & \multicolumn{2}{c}{Consistency} \\
        \cmidrule(lr){2-3}\cmidrule(lr){4-5}\cmidrule(l){6-7}
        Method & FPD $\downarrow$ & MMD $\downarrow$ & FDC $\uparrow$ & CDA$_{\mathrm{3D}}$ (\%) $\uparrow$ & CFCA (\%) $\uparrow$ & CFSC $\uparrow$ \\
        \midrule
        \multicolumn{7}{@{}l}{\textbf{KITTI-360} (sequences 0000 and 0002, 2,000 scans, 3D boxes)} \\
        R2DM (U)~\cite{nakashima2024lidar} & 1.89 & 5.03 & 0.37 & n/a & n/a & n/a \\
        R2Flow (U)~\cite{nakashima2025r2flow} & 2.49 & 6.84 & 0.37 & n/a & n/a & n/a \\
        R3DPA (U)~\cite{sereyjol2026r3dpa} & \csecond 1.54 & \csecond 3.99 & \cfirst\textbf{0.42} & n/a & n/a & n/a \\
        OpenDWM-MaskGIT (C)~\cite{opendwm} & 4.49 & 12.51 & \cfirst\textbf{0.42} & \csecond 4.50 & \csecond 47.5 & \csecond 0.718 \\
        OpenDWM-DiT (C)~\cite{opendwm} & 3.71 & 10.23 & 0.37 & 0.15 & 45.7 & 0.629 \\
        \ours\ (C) & \cfirst\textbf{1.23} & \cfirst\textbf{3.07} & \csecond 0.39 & \cfirst\textbf{20.41} & \cfirst\textbf{92.5} & \cfirst\textbf{0.726} \\
        \textit{Measured scans (Ref.)} & 0.22 & 0.17 & 0.50 & 22.86 & 98.7 & 0.724 \\
        \midrule
        \multicolumn{7}{@{}l}{\textbf{nuScenes} (2,400 keyframes, 3D boxes and HD maps)} \\
        OpenDWM-MaskGIT (C)~\cite{opendwm} & 1.39 & 2.14 & \csecond 0.81 & 3.71 & 31.3 & 0.711 \\
        OpenDWM-DiT (C)~\cite{opendwm} & \csecond 1.26 & \csecond 1.88 & \cfirst\textbf{0.84} & \csecond 5.09 & \csecond 45.6 & \cfirst\textbf{0.718} \\
        UniScene (C)~\cite{li2025uniscene} & 1.31 & 2.04 & 0.69 & 0.05 & 20.7 & 0.628 \\
        \ours\ (C) & \cfirst\textbf{0.91} & \cfirst\textbf{1.12} & \csecond 0.81 & \cfirst\textbf{17.03} & \cfirst\textbf{47.9} & \csecond 0.715 \\
        \textit{Measured scans (Ref.)} & 0.26 & 0.01 & 0.92 & 37.74 & 93.3 & 0.741 \\
        \bottomrule
    \end{tabular}}
    \par\vspace{2pt}\parbox{\linewidth}{\scriptsize U/C: unconditional/layout-conditioned. n/a: not applicable to unconditional generation. Unconditional baselines trained on the evaluation frames are omitted.}
    \vspace{-3mm}
\end{table}\begin{figure}[t]
    \centering
    \includegraphics[width=0.88\linewidth]{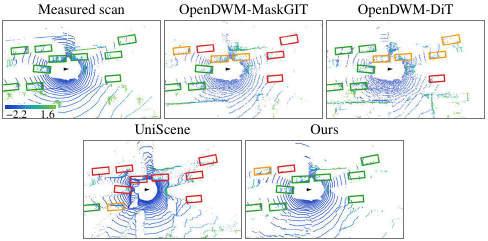}
    \caption{\textbf{Box agreement on nuScenes.}
    On the measured and generated scans of one frame, detections of the \tabref{tab:object_generation} detector scoring above 0.3 are matched one-to-one to the conditioning boxes of cars within 20\,m.
    These boxes are \textcolor{dettight}{tight} with a 3D IoU of at least 0.7, \textcolor{detloose}{loose} with only a \ac{BEV} IoU of at least 0.5, or \textcolor{detmiss}{unmatched}.
    Points are colored by height in meters.}
    \label{fig:det_cmp}
    \vspace{-5.5mm}
\end{figure}}%

\noindent\textbf{Unconditional generation.}
As shown in \tabref{tab:uncond_generation}, our model achieves the best values in all distribution metrics on nuScenes and in all except FSVD on KITTI-360.
On nuScenes, it lowers FPVD from 48.43 to 10.57 and the \ac{BEV} MMD from 1.90 to 0.11 relative to the best baseline for each metric.
It also generates a scan in less than 1\,s, the fastest on nuScenes, and on KITTI-360 it is more than 60 times faster than T2LDM and T2LDM++, which rank first and second in FSVD.
These results indicate that our generated scans match measured scans most closely in both 3D and \ac{BEV} while generation remains fast.
We attribute this fidelity in part to the decoder, which learns each ray's hit probability and depth directly from labeled hits and misses.

\noindent\textbf{Layout-conditioned generation.}
\tabref{tab:object_generation} evaluates the objects in scenes generated from the same layouts for all layout-conditioned methods.
Our model achieves the best FPD, MMD, CDA$_{\mathrm{3D}}$, and CFCA on both datasets and stays close to the best on the remaining metrics.
These results indicate that our generated objects resemble real objects and follow their conditioning boxes most closely.
OpenDWM and UniScene read returns from voxels that quantize object surfaces, whereas our decoder predicts a continuous depth for each ray.
As \figref{fig:det_cmp} shows for a nuScenes frame, the objects detected in our scans agree with the conditioning boxes most closely among the generated scans.
Over all evaluated frames, they match these boxes tightly in 3D about twice as often as detections on OpenDWM-DiT scans and also lead in \ac{BEV} agreement.
Appendix~\ref{sec:app_cond} reports these numbers, and Appendix~\ref{sec:app_qual} shows more qualitative results.

\afterpage{\begin{table}[t]
    \centering
    \caption{Temporal consistency. Subscripts are frame gaps of 0.1\,s on KITTI-360 and 0.5\,s on nuScenes, and all values are in meters. UniScene generates from layouts alone, without observed scans.}
    \label{tab:sequence_generation}
    \scriptsize
    \setlength{\tabcolsep}{3.0pt}
    \renewcommand{\arraystretch}{0.94}
    \resizebox{\linewidth}{!}{\begin{tabular}{@{}lcccccc@{}}
        \toprule
        Method & TTCE$_3$ $\downarrow$ & TTCE$_4$ $\downarrow$ & CTC$_1$ $\downarrow$ & CTC$_2$ $\downarrow$ & CTC$_3$ $\downarrow$ & CTC$_4$ $\downarrow$ \\
        \midrule
        \multicolumn{7}{@{}l}{\textbf{KITTI-360} (sequences 0000 and 0002, 250 windows)} \\
        OpenDWM-MaskGIT~\cite{opendwm} & \csecond 0.059 & \csecond 0.084 & \csecond 0.139 & \csecond 0.148 & \csecond 0.154 & \csecond 0.159 \\
        OpenDWM-DiT~\cite{opendwm} & 0.180 & 0.218 & 0.140 & 0.151 & 0.158 & 0.164 \\
        \ours & \cfirst\textbf{0.020} & \cfirst\textbf{0.025} & \cfirst\textbf{0.097} & \cfirst\textbf{0.107} & \cfirst\textbf{0.115} & \cfirst\textbf{0.120} \\
        \midrule
        \multicolumn{7}{@{}l}{\textbf{nuScenes} (326 windows)} \\
        OpenDWM-MaskGIT~\cite{opendwm} & 1.652 & 2.311 & 0.158 & 0.167 & 0.173 & 0.176 \\
        OpenDWM-DiT~\cite{opendwm} & 1.527 & 2.083 & 0.160 & 0.169 & 0.175 & 0.179 \\
        UniScene~\cite{li2025uniscene} & \csecond 1.492 & \csecond 2.081 & \csecond 0.152 & \csecond 0.164 & \csecond 0.169 & \csecond 0.174 \\
        \ours & \cfirst\textbf{0.413} & \cfirst\textbf{0.614} & \cfirst\textbf{0.139} & \cfirst\textbf{0.152} & \cfirst\textbf{0.159} & \cfirst\textbf{0.165} \\
        \bottomrule
    \end{tabular}}
    \vspace{-1.5mm}
\end{table}\begin{figure}[t]
    \centering
    \includegraphics[width=\linewidth]{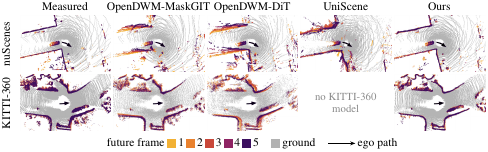}
    \caption{\textbf{Sequence prediction.}
    OpenDWM and ours receive the same three observed scans, future ego poses, and 3D boxes.
    All five future scans are moved to the first future frame with the given poses, and above-ground points are colored by frame.}
    \label{fig:seq_pred}
    \vspace{-1.5mm}
\end{figure}\begin{figure}[t]
    \centering
    \includegraphics[width=0.87\linewidth]{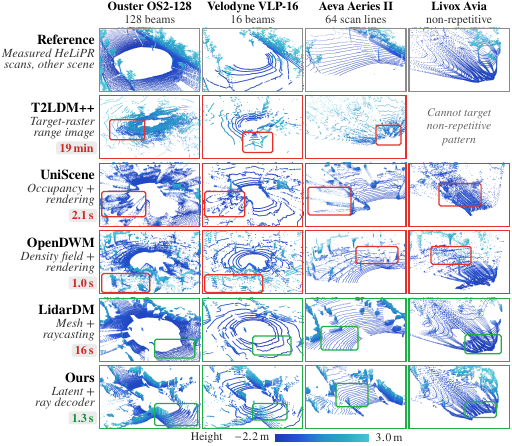}
    \caption{\textbf{Transfer across scan patterns.}
    Each method produces scans of one generated scene with four unseen HeLiPR scan patterns, except T2LDM++, which generates one sample per raster.
    Red frames mark broken patterns, and boxes mark regions to compare.
    Times below each method name cover generation and all scans on one RTX 3090.}
    \label{fig:transfer}
    \vspace{-6mm}
\end{figure}}%

\noindent\textbf{Motion-conditioned prediction.}
We check whether predicted sequences follow the given ego motion and keep the scene geometry consistent across frames.
As shown in \tabref{tab:sequence_generation}, our model achieves the lowest TTCE and CTC on both datasets.
We attribute this consistency to the motion-aligned conditioning, which warps the reference latents with the given ego and object motion.
In \figref{fig:seq_pred}, where the vehicle turns, the frames predicted by OpenDWM and those generated by UniScene from layouts alone leave several copies of the same wall or curb once the given poses are applied.
Our predicted frames overlap as in the measured scans and keep the sensor's scan lines, whereas OpenDWM predictions on KITTI-360 show the regular pattern of a voxel grid.
Appendix~\ref{sec:app_pred} compares the predictions with measured future scans and shows layout editing.

\subsection{Observation across Scan Patterns}
\label{sec:transfer}
\label{sec:ae_recon}
\afterpage{\begin{figure}[t]

    \centering
    \captionof{table}{Returns from scene representations of measured scans. Only the inputs of OpenDWM and ours omit the masked rays. KITTI-360 blocks use the 362 scans with SSCBench labels.}
    \label{tab:occ_oracle}
    \scriptsize
    \setlength{\tabcolsep}{2.2pt}
    \renewcommand{\arraystretch}{0.94}
    \resizebox{\linewidth}{!}{\begin{tabular}{@{}llcccccccc@{}}
        \toprule
        & & \multicolumn{4}{c}{KITTI-360, VLP-16 rows masked} & \multicolumn{4}{c}{KITTI-360, Avia cells masked} \\
        \cmidrule(lr){3-6}\cmidrule(l){7-10}
        Scene (resolution) & Method & CD $\downarrow$ & $F_{0.1}$ $\uparrow$ & dMAE $\downarrow$ & AbsRel $\downarrow$ & CD $\downarrow$ & $F_{0.1}$ $\uparrow$ & dMAE $\downarrow$ & AbsRel $\downarrow$ \\
        \midrule
        Occ., SSCBench 0.2\,m & Ray marching & 0.180 & 0.085 & 1.51 & 12.14 & 0.173 & 0.085 & 1.76 & 10.99 \\
        Mesh, NKSR 0.15\,m & Raycasting & \csecond 0.086 & \csecond 0.663 & 0.55 & 4.49 & \cfirst\textbf{0.077} & \cfirst\textbf{0.707} & \csecond 0.49 & \csecond 2.57 \\
        Mesh, NKSR 0.15\,m & LidarDM~\cite{zyrianov2025lidardm} & 0.134 & 0.371 & 0.86 & 5.84 & 0.131 & 0.359 & 0.97 & 5.20 \\
        Latent, OpenDWM~\cite{opendwm} & Render head & 0.101 & 0.564 & \csecond 0.48 & \csecond 3.81 & 0.102 & 0.563 & 0.77 & 3.86 \\
        Latent, \ours & Ray decoder & \cfirst\textbf{0.083} & \cfirst\textbf{0.665} & \cfirst\textbf{0.35} & \cfirst\textbf{2.73} & \csecond 0.083 & \csecond 0.645 & \cfirst\textbf{0.40} & \cfirst\textbf{2.36} \\
        \midrule
        & & \multicolumn{4}{c}{KITTI-360, full scan} & \multicolumn{4}{c}{nuScenes, full scan} \\
        \cmidrule(lr){3-6}\cmidrule(l){7-10}
        Scene (resolution) & Method & CD $\downarrow$ & $F_{0.1}$ $\uparrow$ & dMAE $\downarrow$ & AbsRel $\downarrow$ & CD $\downarrow$ & $F_{0.1}$ $\uparrow$ & dMAE $\downarrow$ & AbsRel $\downarrow$ \\
        \midrule
        Occ., SSCBench 0.2\,m & Ray marching & 0.160 & 0.162 & 1.43 & 11.81 & -- & -- & -- & -- \\
        Occ., Occ3D 0.4\,m & Ray marching & -- & -- & -- & -- & 0.173 & 0.103 & 1.88 & 13.46 \\
        Occ., Occ3D 0.4\,m & UniScene~\cite{li2025uniscene} & -- & -- & -- & -- & 0.159 & 0.190 & 1.08 & 10.71 \\
        Mesh, NKSR 0.15\,m & Raycasting & \csecond 0.073 & \csecond 0.741 & 0.51 & 4.10 & 0.090 & 0.609 & 0.99 & 5.85 \\
        Mesh, NKSR 0.15\,m & LidarDM~\cite{zyrianov2025lidardm} & 0.103 & 0.561 & 0.75 & 5.38 & -- & -- & -- & -- \\
        Latent, OpenDWM~\cite{opendwm} & Render head & 0.081 & 0.704 & \csecond 0.33 & \csecond 3.11 & \csecond 0.075 & \csecond 0.706 & \csecond 0.36 & \csecond 2.95 \\
        Latent, \ours & Ray decoder & \cfirst\textbf{0.055} & \cfirst\textbf{0.817} & \cfirst\textbf{0.24} & \cfirst\textbf{1.96} & \cfirst\textbf{0.064} & \cfirst\textbf{0.762} & \cfirst\textbf{0.29} & \cfirst\textbf{2.29} \\
        \bottomrule
    \end{tabular}}
    \vspace{2.5mm}
    \includegraphics[width=\linewidth]{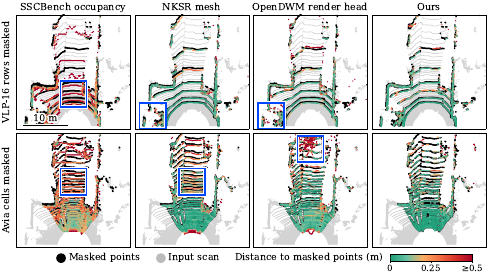}\par
    \vspace{-1.6mm}%
    \captionof{figure}{\textbf{Reading masked rays} on a KITTI-360 scan.
    OpenDWM and ours encode the gray input without the VLP-16 rows (top) or the Avia cells (bottom).
    Black points mark measured returns on masked rays. Colors show the distance to these points up to 0.5\,m. Blue boxes highlight typical errors.}
    \label{fig:occ_readout}
    \vspace{-6mm}
\end{figure}}%

\noindent\textbf{Unseen scan patterns.}
Because only the decoder receives the target scan pattern, one generated scene should be observable at the same mounting position by sensors that the model never saw in training.
In \figref{fig:transfer}, our model generates each scene latent once and decodes it along the ray directions of four HeLiPR~\cite{jung2024helipr} sensors, including the non-repetitive Livox Avia, and \figref{fig:teaser}d shows the same for a KITTI-360 latent.
T2LDM++, UniScene, and OpenDWM fail to preserve these patterns, while only LidarDM and our model preserve all four.
Our model generates a scene and all four scans in 1.3\,s, an order of magnitude faster than LidarDM at 16\,s, as detailed in Appendix~\ref{sec:app_gen}.
Further predicted sequences with the same four patterns can be found in Appendix~\ref{sec:app_qual}.
However, as the distribution distance between these generated scans and measured HeLiPR scans would also include differences in city, vehicle, and mounting, it could not isolate the effect of the scan pattern.
Since the sensor changes only how a scene is observed, we instead measure how accurately our model and the approaches of \figref{fig:teaser}a and b reproduce a measured scan, which serves as ground truth.

\afterpage{\newcommand{\reconKittiRows}{LiDM (VQ)~\cite{ran2024lidm} & 0.084 & 0.682 & 0.463 & 0.093 & 1.595 & 3.47 \\
    OpenDWM (VQ)~\cite{opendwm} & 0.122 & 0.387 & 1.204 & 0.280 & 3.262 & 11.61 \\
    OpenDWM (VAE, voxel head)~\cite{opendwm} & 0.118 & 0.406 & 1.040 & 0.252 & 2.951 & 10.10 \\
    OpenDWM (VAE, render head)~\cite{opendwm} & 0.083 & 0.688 & 0.421 & 0.087 & 1.769 & 4.17 \\
    R3DPA (VAE)~\cite{sereyjol2026r3dpa} & 0.095 & 0.628 & 0.456 & 0.115 & 1.581 & 3.72 \\
    GEM (VQ)~\cite{wu2026gem} & \csecond 0.063 & \cfirst\textbf{0.804} & \csecond 0.363 & \csecond 0.053 & \csecond 1.509 & \csecond 2.84 \\
    \ours & \cfirst\textbf{0.059} & \csecond 0.803 & \cfirst\textbf{0.303} & \cfirst\textbf{0.043} & \cfirst\textbf{1.492} & \cfirst\textbf{2.64} \\
}
\newcommand{\reconNuscenesRows}{OpenDWM (VQ)~\cite{opendwm} & 0.123 & 0.370 & 0.714 & 0.141 & 2.487 & 6.22 \\
    OpenDWM (VAE, voxel head)~\cite{opendwm} & 0.117 & 0.401 & 0.564 & 0.127 & 2.140 & 5.10 \\
    OpenDWM (VAE, render head)~\cite{opendwm} & \csecond 0.075 & \csecond 0.706 & \csecond 0.364 & \csecond 0.043 & \csecond 1.816 & \csecond 2.95 \\
    UniScene (occupancy VAE)~\cite{li2025uniscene} & 0.159 & 0.188 & 1.079 & 0.351 & 2.748 & 10.80 \\
    \ours & \cfirst\textbf{0.064} & \cfirst\textbf{0.762} & \cfirst\textbf{0.294} & \cfirst\textbf{0.033} & \cfirst\textbf{1.525} & \cfirst\textbf{2.29} \\
}
\begin{table}[t]
    \centering
    \caption{Reconstruction by the autoencoders of \ac{LiDAR} generators.}
    \label{tab:vae_fidelity}
    \scriptsize
    \setlength{\tabcolsep}{1.6pt}
    \renewcommand{\arraystretch}{0.94}
    \resizebox{\linewidth}{!}{\begin{tabular}{@{}lcccccc@{}}
        \toprule
        Method & CD $\downarrow$ & $F_{0.1}$ $\uparrow$ & dMAE $\downarrow$ & MedAE $\downarrow$ & RMSE $\downarrow$ & AbsRel $\downarrow$ \\
        \midrule
        \multicolumn{7}{@{}l}{\textbf{KITTI-360} (sequence 0018, 2,000 scans)} \\
        \reconKittiRows
        \midrule
        \multicolumn{7}{@{}l}{\textbf{nuScenes} (6,019 validation scans)} \\
        \reconNuscenesRows
        \bottomrule
    \end{tabular}}
    \vspace{-5mm}
\end{table}}%
\noindent\textbf{Returns from scene representations.}
As \figref{fig:teaser}b illustrates, rendering-based models store the scene as occupancy (UniScene), meshes (LidarDM), or latents (OpenDWM), and they render scans along the requested rays. Even with a perfectly generated scene, their scans are only as accurate as the returns read from this representation.
To compare only this observation step, we build each representation from measured data, using the occupancy labels of the SSCBench~\cite{li2024sscbench} and Occ3D~\cite{tian2023occ3d} benchmarks and NKSR~\cite{huang2023nksr} meshes from accumulated scans.
OpenDWM and our model encode only the input scan into latents.
To approximate an unseen sensor, the masked setting removes either the KITTI-360 rows closest to the VLP-16 beams or the cells hit by the Avia rays from the input of OpenDWM and our model.
Only the returns on these masked rays are scored.
The full setting masks nothing and tests how well each method reproduces the scan.

As \tabref{tab:occ_oracle} and \figref{fig:occ_readout} show, our model has the lowest depth errors in every setting and the lowest CD and highest $F_{0.1}$ except with the Avia cells masked.
There, the NKSR mesh, fused from scans that also observe the masked region, places more of its points close to the measured ones.
Even so, the lowest depth errors in every setting show that decoding each ray from the scene latent places returns more accurately than rendering them from occupancy, meshes, or a render head.
As ray marching stops at the first occupied voxel, quantization can place returns in front of the surface even with benchmark labels or the finer voxels checked in Appendix~\ref{sec:app_occ_var}.
Mesh raycasting returns a point on every ray that hits the mesh, even without a measured return, and the OpenDWM render head returns points off the surface.
Our depth-guided decoder instead learns from labeled hits and misses whether each ray returns and at which depth, using only the scene latent and the ray direction (Appendix~\ref{sec:app_rays}).
This test reflects the accuracy of the scans in \figref{fig:transfer}, which have no ground truth but come from the same decoder.

\afterpage{\begin{figure}[t]
    \centering
    \captionof{table}{Ablation study for autoencoder design, evaluated by reconstruction. Parentheses show the relative change (\%) from the full model in the last row.}
    \label{tab:ablation}
    \label{tab:ablation_paired}
    \scriptsize
    \setlength{\tabcolsep}{1.2pt}
    \renewcommand{\arraystretch}{0.94}
    \resizebox{\linewidth}{!}{\begin{tabular}{@{}cccccccccc@{}}
        \toprule
        \multicolumn{4}{c}{Components} & \multicolumn{3}{c}{KITTI-360} & \multicolumn{3}{c}{nuScenes} \\
        \cmidrule(r){1-4}\cmidrule(lr){5-7}\cmidrule(l){8-10}
        Grid & Desc. & Local & Cont. & CD $\downarrow$ & $F_{0.1}$ $\uparrow$ & dMAE $\downarrow$ & CD $\downarrow$ & $F_{0.1}$ $\uparrow$ & dMAE $\downarrow$ \\
        \midrule
        Cart. & $\times$ & $\checkmark$ & $\checkmark$ & 0.116\dd{+87} & 0.468\dd{\textminus 40} & 0.743\dd{+141} & 0.124\dd{+95} & 0.405\dd{\textminus 47} & 0.753\dd{+156} \\
        HCS & $\times$ & $\checkmark$ & $\checkmark$ & 0.078\dd{+26} & 0.697\dd{\textminus 11} & \csecond 0.375\dd{+21} & 0.095\dd{+49} & 0.588\dd{\textminus 23} & \csecond 0.390\dd{+33} \\
        HCS & $\checkmark$ & $\times$ & -- & \csecond 0.073\dd{+18} & \csecond 0.722\dd{\textminus 8.0} & 0.396\dd{+28} & 0.077\dd{+20} & 0.693\dd{\textminus 9.0} & 0.407\dd{+39} \\
        HCS & $\checkmark$ & $\checkmark$ & $\times$ & 0.074\dd{+19} & \csecond 0.722\dd{\textminus 8.0} & 0.405\dd{+31} & \csecond 0.076\dd{+19} & \csecond 0.701\dd{\textminus 7.9} & 0.405\dd{+38} \\
        HCS & $\checkmark$ & $\checkmark$ & $\checkmark$ & \cfirst\textbf{0.062} & \cfirst\textbf{0.785} & \cfirst\textbf{0.308} & \cfirst\textbf{0.064} & \cfirst\textbf{0.762} & \cfirst\textbf{0.294} \\
        \bottomrule
    \end{tabular}}
    \par\vspace{2pt}\parbox{\linewidth}{\scriptsize Grid: Cartesian or HCS with equal cell count and latent size. Desc.: point count and radial offset added to occupancy. Local: local features for depth refinement. Cont.: lookup at the continuous preliminary depth ($\checkmark$) or bin center ($\times$). --: not applicable.}
    \vspace{-3mm}
\end{figure}\begin{figure}[t]
    \centering
    \captionof{table}{World model ablation on KITTI-360. Parentheses show the relative change (\%) from the full model.}
    \label{tab:wm_ablation}
    \scriptsize
    \setlength{\tabcolsep}{1.0pt}
    \renewcommand{\arraystretch}{0.94}
    \resizebox{\linewidth}{!}{\begin{tabular}{@{}cccccccc@{}}
        \toprule
        \multicolumn{2}{c}{Components} & \multicolumn{3}{c}{Prediction accuracy} & \multicolumn{3}{c}{Temporal consistency} \\
        \cmidrule(r){1-2}\cmidrule(lr){3-5}\cmidrule(l){6-8}
        $\mathcal{L}_{\mathrm{ray}}$ & Warp & CD $\downarrow$ & $F_{0.1}$ $\uparrow$ & MedAE $\downarrow$ & TTCE$_3$ $\downarrow$ & TTCE$_4$ $\downarrow$ & CTC$_4$ $\downarrow$ \\
        \midrule
        $\checkmark$ & $\times$ & 0.091\dd{+4.9} & 0.623\dd{\textminus 4.1} & 0.097\dd{+11} & 0.049\dd{+142} & 0.071\dd{+189} & 0.126\dd{+4.4} \\
        $\times$ & $\checkmark$ & \csecond 0.088\dd{+1.6} & \csecond 0.640\dd{\textminus 1.4} & \csecond 0.090\dd{+2.8} & \csecond 0.023\dd{+16} & \csecond 0.040\dd{+62} & \csecond 0.122\dd{+1.3} \\
        $\checkmark$ & $\checkmark$ & \cfirst\textbf{0.087} & \cfirst\textbf{0.649} & \cfirst\textbf{0.087} & \cfirst\textbf{0.020} & \cfirst\textbf{0.025} & \cfirst\textbf{0.120} \\
        \bottomrule
    \end{tabular}}
    \vspace{-4.5mm}
\end{figure}}%
\noindent\textbf{Reconstruction by autoencoders.}
For generators with autoencoders in \figref{fig:teaser}a and b, generated scans are also limited by the accuracy of their autoencoders.
\tabref{tab:vae_fidelity} compares these vector-quantized (VQ) and variational (VAE) autoencoders on full measured scans.
On both datasets, our model has the lowest CD and depth errors.
On KITTI-360, GEM comes closest, with a slightly higher $F_{0.1}$ but higher CD and depth errors.
Overall, our model preserves measured scans most closely.
Since range image tokenizers decode a fixed uniform raster whose rows do not align with the uneven beam elevations, their reconstructed points can be offset in elevation.
Our decoder instead queries the rays that the sensor fired, as for any other scan pattern.
OpenDWM's render head is more accurate than its voxel head, and UniScene's autoencoder, which reconstructs accumulated occupancy instead of the scan, has the largest errors on nuScenes.
Appendix~\ref{sec:app_rays} shows that our autoencoder also leads most metrics on masked inputs.

\subsection{Ablation Study}
\label{sec:ablation}

\noindent\textbf{Autoencoder.}
\tabref{tab:ablation} ablates the grid, the cell descriptors, the local features, and their retrieval depth.
The full model is best in every metric.
With occupancy alone, a Cartesian grid of the same cell count and latent size is worse than the \ac{HCS} grid in every metric and roughly doubles the dMAE.
Its uniform cells span many beams near the sensor, where returns are densest.
Without the point count and radial offset, a cell records only its occupancy, not where its points lie in range, and the CD rises on both datasets.
Without the local features, the depth rests on global scene context that lacks the fine geometry around the hit, and the dMAE rises by about a third.
Retrieving them at the bin center, without the offset, raises the dMAE about as much as omitting them.
Appendix~\ref{sec:app_rays} shows that, in the full decoder, refinement more than halves the median depth error.
It also shows that the scan pattern augmentation, drawn from aggregated scans, improves the autoencoder on sparse and irregular inputs at a small cost on complete scans.

\afterpage{\begin{figure}[t]
    \centering
    \includegraphics[width=\linewidth]{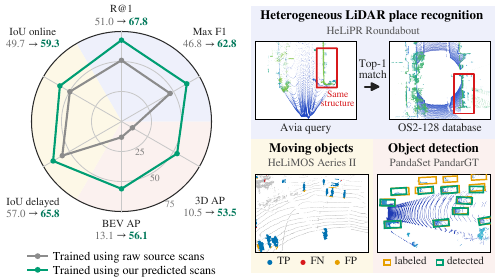}
    \caption{\textbf{Downstream tasks on unseen LiDARs.}
    Left: mean Recall@1 (R@1) and maximum F1 score of place recognition on Roundabout in HeLiPR, mean online and delayed moving IoU of moving object segmentation, and zero-shot 3D and BEV car AP (IoU 0.5) of object detection in \%.
    Right: one example per task from the models trained on our predicted scans.
    Background colors link each task on the left to its example on the right.}
    \label{fig:downstream}
    \vspace{-5mm}
\end{figure}}%
\noindent\textbf{World model.}
In \tabref{tab:wm_ablation}, each variant removes one component from the full model and is evaluated against measured future scans in the KITTI-360 windows of \tabref{tab:sequence_generation}.
Removing the warp degrades prediction accuracy and more than doubles the registration error between predicted frames.
Without it, the network must infer the motion from unaligned references.
Removing the ray loss degrades every metric, as the latent loss weighs all latent errors equally, although a small latent error can still change whether a ray returns or at which depth.
Appendix~\ref{sec:app_pred} analyzes both components further.

\subsection{Downstream Tasks across LiDARs}
\label{sec:hetero_lpr}
In our method, a predicted scene can be decoded with the scan pattern of any LiDAR at the same mounting position.
We test whether such scans help three downstream tasks on sensors absent from training, without training on any point or label of these sensors.
Our predictions decoded with HeLiPR patterns train HeLiOS~\cite{jung2025helios} for place recognition and fine-tune 4DMOS~\cite{mersch2022receding} for moving object segmentation on HeLiMOS~\cite{lim2024helimos}.
Those decoded with the PandarGT pattern train Voxel R-CNN~\cite{deng2021voxelrcnn} for car detection on PandaSet~\cite{xiao2021pandaset}.
As \figref{fig:downstream} shows, these scans raise every mean score over training on the raw source scans, most for detection.
As our decoder follows the scan pattern of each test sensor, our world model can supply training data for a LiDAR by using its ray directions alone.
Appendix~\ref{sec:app_downstream} gives the setups and detailed results of these downstream tasks and shows that our predicted scans come close to our autoencoded source scans.
\section{Conclusion}
\label{sec:conclusion}

We presented a \ac{LiDAR} world model that separates the modeled scene from the sensor that observes it.
Scans of different \acp{LiDAR} are encoded into a shared spatial latent, in which the model generates realistic scenes with or without layouts and predicts temporally consistent sequences under given ego and object motion.
Warping the observed latents by this motion more than halves the registration error between predicted frames on KITTI-360.
A depth-guided decoder can query this latent along any ray from the sensor origin without an intermediate 3D surface or field.
In our evaluation, it yields lower depth errors than rendering from occupancy, meshes, or a render head.
The same generated scene can be decoded with scan patterns not used in training, and such scans help three downstream tasks.

{
    \small
    \bibliographystyle{IEEEtranN}
    \bibliography{references}
}

\clearpage
\maketitlesupp
\setcounter{section}{0}
\renewcommand{\thesection}{\Alph{section}}
\renewcommand{\thesectiondis}{\thesection.}
\renewcommand{\thesubsection}{\thesection.\arabic{subsection}}
\renewcommand{\thesubsectiondis}{\thesubsection.}
\renewcommand{\theHsection}{app.\Alph{section}}
\renewcommand{\theHsubsection}{app.\Alph{section}.\arabic{subsection}}
\raggedbottom
\makeatletter
\let\app@citep\citep
\DeclareRobustCommand\cite[2][]{\begingroup\@fileswfalse
  \edef\app@keys{\zap@space#2 \@empty}%
  \def\app@out{}%
  \@for\app@x:=\app@keys\do{\edef\app@out{\ifx\app@out\@empty\else\app@out,\fi app:\app@x}}%
  \edef\app@cmd{\noexpand\app@citep\ifx\relax#1\relax\else[\unexpanded{#1}]\fi{\app@out}}%
  \app@cmd\endgroup}%
\makeatother
\begingroup
\makeatletter
\let\texorpdfstring\@secondoftwo
\newcommand\appsec[1]{\par\addvspace{0.35em}\noindent\hyperref[#1]{\makebox[1.5em][l]{\bfseries\ref*{#1}.}\textbf{\nameref*{#1}}}\hfill\hyperref[#1]{\textbf{\pageref*{#1}}}\par}
\newcommand\appsub[1]{\par\noindent\hspace*{1.5em}\hyperref[#1]{\makebox[2.3em][l]{\ref*{#1}.}\nameref*{#1}}\ \dotfill\ \hyperref[#1]{\pageref*{#1}}\par}
\section*{Contents}
\appsec{sec:app_setup}
\appsub{sec:app_data}
\appsub{sec:app_train}
\appsub{sec:app_evaldata}
\appsub{sec:app_metrics}
\appsub{sec:app_baselines}
\appsec{sec:app_model}
\appsec{sec:app_cond}
\appsec{sec:app_occ}
\appsub{sec:app_occ_proto}
\appsub{sec:app_occ_var}
\appsub{sec:app_occ_depth}
\appsub{sec:app_occ_neural}
\appsec{sec:app_gen}
\appsub{sec:app_gen_transfer}
\appsub{sec:app_gen_scan}
\appsec{sec:app_rays}
\appsub{sec:app_rays_pred}
\appsub{sec:app_rays_err}
\appsub{sec:app_rays_consist}
\appsub{sec:app_rays_unseen}
\appsub{sec:app_rays_aug}
\appsec{sec:app_pred} \appsub{sec:app_pred_warp} \appsub{sec:app_pred_gem} \appsub{sec:app_pred_inputs} \appsub{sec:app_pred_ablation} \appsub{sec:app_pred_edit}
\appsec{sec:app_downstream} \appsub{sec:app_down_lpr} \appsub{sec:app_down_mos} \appsub{sec:app_down_det}
\appsec{sec:app_qual}
\appsec{sec:app_limits}
\par\addvspace{0.6em}
\makeatother
\endgroup
\section{Experimental Setup Details}
\label{sec:app_setup}
This section details the training data and schedule, evaluation data and metrics, and methods compared in Sec.~\ref{sec:evaluation}.

\subsection{Training Data Preparation}
\label{sec:app_data}
We train on single and aggregated scans from KITTI-360~\cite{liao2022kitti360} sequences 0003 to 0007, 0009, and 0010 and from the 700 nuScenes~\cite{caesar2020nuscenes} training scenes.
Each aggregated scan fuses many scans into a denser scan in the raw sensor frame of a target scan.
We build aggregated scans for 12,088 KITTI-360 frames and all 28,130 nuScenes training keyframes. No evaluation scan contributes to an aggregated scan used for training.
Aggregated scans train both the autoencoder and the world model, whereas every evaluation reference is a single measured scan.

\noindent\textbf{Static scene fusion.}
We transform scans into the world frame using measured poses and official \ac{LiDAR} extrinsics.
As each KITTI-360 scan spans one rotation during vehicle motion, we first undistort it using poses interpolated with splines at each point's acquisition time.
On KITTI-360, we fuse all scans of the sequence into a map and retain the map within 72\,m of the target.
Because nuScenes annotates moving objects only on keyframes, we fuse the target with every keyframe of the same scene within 20\,m and 10\,s.
To prevent trails of moving objects, we remove points inside annotated dynamic boxes enlarged by 0.2\,m at each scan's timestamp.
A nuScenes object is dynamic if its attribute marks it as moving, or if it belongs to a mobile category and moves faster than 0.5\,m/s.
On both datasets, the target scan itself keeps its dynamic points.

\noindent\textbf{Hit and miss labels.}
Because our decoder learns hits and misses, every supervised direction needs a label.
We project fused points from the sensor origin of the target scan onto a $0.05^\circ$ direction grid, finer than the beam spacing of both training sensors.
The grid spans $-30^\circ$ to $40^\circ$ in elevation to support scan patterns that reach above the fields of view of the training sensors, which end at $2^\circ$ on KITTI-360 and $10.7^\circ$ on nuScenes.
Each cell's nearest point is a hit if its depth falls within the decoder range of 2.4 to 72\,m.
A miss denotes no return within this range.
For single scans, sensor rays without measured returns are misses, as in the ray drop labels of \ac{LiDAR} simulators~\cite{huang2023nfl,tao2024lidarnerf,zyrianov2025lidardm}.
For aggregated samples, the two datasets differ in which cells without a hit become misses.

\noindent\textbf{Miss labels on KITTI-360.}
KITTI-360 aggregation fuses a median of 6,743 scans from the whole sequence.
Within the sensor's field of view, this map densely covers the static surfaces, so we label every cell without a hit as a miss.
SurroundOcc similarly treats voxels outside its reconstructed mesh as empty~\cite{wei2023surroundocc}.
Above the field of view, this rule also labels the upper parts of nearby structures as misses.
We keep the rule there because one decoder is trained on both datasets, and nuScenes labels almost no direction above its field of view ($10.7^\circ$).
Without these misses, the decoder would receive almost no supervision above $10.7^\circ$.
Between $2^\circ$ and $10.7^\circ$, this rule differs from nuScenes, whose sensor observes this band.
Since every cell with a fused point remains a hit, the rule affects only directions without points, and the decoder still returns where the encoded scan has points.
To check whether the map captures surfaces observed by the target sensor, we rebuild labels for 200 scans sampled from sequences 0000 and 0002, excluding each target scan.
Among measured returns within the decoder range and outside moving objects, 2.2\% lack a hit in the surrounding $3\times3$ cells and would be labeled misses.
A stricter test that also excludes all scans within 0.5\,s of the target raises this fraction to 7.5\%.
Thus, the map covers nearly every direction in which the target sensor measures a static surface.
Near the sensor, cells between the fused points on these surfaces also receive miss labels, yet the decoder predicts returns in most of them when the aggregated scan is encoded.

\noindent\textbf{Miss labels on nuScenes.}
nuScenes aggregation fuses only nearby keyframes, which view the scene from a short segment of the trajectory.
Its fused points cover nearly an order of magnitude fewer cells than the KITTI-360 maps.
As a result, many empty cells are unobserved rather than free.
A cell without a hit in range is a miss only if its nearest projected point lies between 72 and 150\,m or a ray of the target single scan in the cell has no return.
Other cells remain unlabeled because the fused scans provide no evidence of whether their rays return.
Occ3D and UnO likewise label as free only the space traversed by measured rays~\cite{tian2023occ3d,agro2024uno}.

\subsection{Training Schedule}
\label{sec:app_train}
The autoencoder and the world model both use AdamW at a learning rate of $10^{-4}$, with 2,000 steps of linear warmup and cosine decay toward $10^{-6}$.

\noindent\textbf{Autoencoder.}
The autoencoder trains for 160k steps with equal shares of single and aggregated scans from each dataset.
Samples of single scans encode the measured scan and supervise the decoder along the directions of a fixed beam grid, with $64\times1024$ rays on KITTI-360 and $32\times1024$ on nuScenes.
Aggregated samples supervise up to 65,536 labeled cells.
When more cells are labeled, we subsample them with a target ratio of three hits per miss.
For half of the aggregated samples, the encoder receives all aggregated hits.
The other half are observed with a random scan pattern, as described in Sec.~\ref{sec:training}.
Each pattern is equally likely to be a raster of rows and columns as in a rotating \ac{LiDAR}, a sweep along continuous curves, or irregular directions.
Vertical and horizontal fields of view span $12^\circ$ to $70^\circ$ and $60^\circ$ to $360^\circ$, respectively, with half covering the full azimuth.
Rasters have 8 to 128 rows and 256 to 2,048 columns.
Their rows are spread over the vertical field of view with random jitter rather than copied from a real sensor.
Sweeps and irregular patterns have 16,384 to 131,072 rays.
We generate patterns in continuous angles and move each ray to the nearest cell center of the $0.05^\circ$ label grid.
Training uses $\beta=(0.9,0.95)$, weight decay 0.01, a global batch of 24 with six samples per dataset and sample type, and gradient clipping at norm 1.0.
Loss weights are $\lambda_h=30$, $\lambda_b=4$, and $\lambda_{\mathrm{KL}}=5\times10^{-7}$.
Within $\lambda_d\mathcal{L}_{\mathrm{depth}}$, the terms for the metric depth and for the ground-truth bin have weights 3 and 2.5.
The focal loss uses a focusing parameter of 2 without class weighting. Both Smooth-$\ell_1$ penalties use a threshold of 0.02, in meters for the metric depth and in bin widths for the term at the ground-truth bin.

\begin{table*}[t]
    \centering
    \caption{Metric definitions and evaluators. FD, $U_k$, and $V_k$ are defined in Appendix~\ref{sec:app_metrics}. CD in Appendix~\ref{sec:app_pred_gem} uses point sets.}
    \label{tab:metric_summary}
    \scriptsize
    \setlength{\tabcolsep}{4pt}
    \renewcommand{\arraystretch}{1.12}
    \begin{tabularx}{\textwidth}{@{}>{\raggedright\arraybackslash}p{6.1cm}>{\raggedright\arraybackslash}X>{\raggedright\arraybackslash}p{3.3cm}@{}}
        \toprule
        Metric & Definition & Features or evaluator \\
        \midrule
        \multicolumn{3}{@{}l}{\textit{Scene distribution in Table~\ref{tab:uncond_generation}}} \\
        FSVD$\downarrow$, Fr\'echet sparse volume distance~\cite{ran2024lidm} & FD of scan features & MinkowskiNet~\cite{choy2019minkowski} (R2Flow, LiDM) \\
        FPVD$\downarrow$, Fr\'echet point-based volume distance~\cite{ran2024lidm} & FD of scan features & SPVCNN~\cite{tang2020spvnas} of LiDARCrafter \\
        MMD$\downarrow$, maximum mean discrepancy & $U_k$ with $k(x,y)=(x^\top y/d+1)^3$ for feature dimension $d$, in $10^{-2}$ & SPVCNN features of FPVD \\
        JSD (\ac{BEV})$\downarrow$, Jensen--Shannon distance & $\sqrt{\mathrm{JS}(p\,\|\,q)}$ of the pooled histograms, in $10^{-2}$ & $100\times100$ grid of 1.6\,m cells \\
        MMD (\ac{BEV})$\downarrow$, maximum mean discrepancy & $V_k$ with $k(p,q)=\exp(-\|p-q\|_2^2/2\sigma_m^2)$, $\sigma_m=0.5$, in $10^{-3}$ & Histograms of each scan on the same grid \\
        \midrule
        \multicolumn{3}{@{}l}{\textit{Objects in Table~\ref{tab:object_generation}}} \\
        FPD$\downarrow$, Fr\'echet point cloud distance~\cite{shu2019treegan}, and MMD$\downarrow$ of objects & FD and $U_k$ (MMD in $10^{-3}$) between detected and annotated car crops & PointMLP~\cite{ma2022pointmlp} features \\
        FDC$\uparrow$, foreground detection confidence & Mean confidence of car detections with at least 50 points & Voxel R-CNN \\
        CDA$_{\mathrm{3D}}\uparrow$, conditioned detection accuracy & Car 3D AP$_{\mathrm{R40}}$ at an IoU of 0.7 against the conditioning boxes & Voxel R-CNN \\
        CFCA$\uparrow$, conditioned foreground classification accuracy & Classification accuracy of objects with at least 50 points & PointMLP \\
        CFSC$\uparrow$, conditioned foreground spatial consistency & Mean IoU of estimated and conditioning boxes with more than 300 points & GLENet~\cite{zhang2023glenet} \\
        \midrule
        \multicolumn{3}{@{}l}{\textit{Sequences in Table~\ref{tab:sequence_generation}}} \\
        TTCE$_k\downarrow$, temporal transformation consistency error & $\operatorname{mean}\|\Delta\hat{\mathbf{t}}-\Delta\mathbf{t}\|_1/3$, the mean absolute error per axis of registered and measured relative translations & KISS-Matcher and generalized ICP \\
        CTC$_k\downarrow$, Chamfer temporal consistency & Capped CD of frames $k$ apart aligned by measured poses, $k\leq4$ & Point sets \\
        \midrule
        \multicolumn{3}{@{}l}{\textit{Reconstruction, rendering, and prediction in Tables~\ref{tab:occ_oracle} to~\ref{tab:wm_ablation} and Appendices~\ref{sec:app_occ} to~\ref{sec:app_pred}}} \\
        Acc, Comp, accuracy and completeness & Mean nearest-neighbor distance, capped at 0.2\,m, to and from the measured scan & Range images \\
        CD$\downarrow$, Chamfer distance & $(\mathrm{Acc}+\mathrm{Comp})/2$ & Range images \\
        $F_{0.1}\uparrow$, F-score at 0.1\,m & $2\,\mathrm{Pr}\,\mathrm{Re}/(\mathrm{Pr}+\mathrm{Re})$, with $\mathrm{Pr}$ and $\mathrm{Re}$ the fractions of output and measured points within 0.1\,m of the other set & Range images \\
        dMAE$\downarrow$, depth mean absolute error & $\operatorname{mean}_{i\in\mathcal{H}}|\hat d_i-d^\star_i|$ over cells $\mathcal{H}$ with a return in both range images & Range images \\
        MedAE$\downarrow$, median absolute depth error & $\operatorname{median}_{i\in\mathcal{H}}|\hat d_i-d^\star_i|$ & Range images \\
        RMSE$\downarrow$, root-mean-square depth error & $\bigl(\operatorname{mean}_{i\in\mathcal{H}}(\hat d_i-d^\star_i)^2\bigr)^{1/2}$ & Range images \\
        AbsRel$\downarrow$, absolute relative error & $\operatorname{mean}_{i\in\mathcal{H}}|\hat d_i-d^\star_i|/d^\star_i$, in \% & Range images \\
        H-F1$\uparrow$, hit F1 score & $2\mathrm{TP}/(2\mathrm{TP}+\mathrm{FP}+\mathrm{FN})$ of the hit decisions per cell & Range images \\
        \bottomrule
    \end{tabularx}
    \vspace{-4mm}
\end{table*}
\noindent\textbf{World model.}
One world model learns scene generation and prediction, as described in Sec.~\ref{sec:worldmodel}.
It trains jointly on both datasets for 120k steps, then fine-tunes on each for 30k steps, with $\beta=(0.9,0.99)$, weight decay $10^{-4}$, and a global batch of 58.
Its target latents encode full scans without the autoencoder's scan pattern augmentation.
Each training window contains eight frames of either single scans or aggregated scans.
The windows use several frame spacings such as 0.1 to 0.6\,s on KITTI-360 and 0.08 to 0.5\,s on nuScenes.
Between nuScenes keyframes, we use boxes interpolated from the keyframe annotations.
Boxes and maps are each dropped independently with probability 0.2, so the model can also be conditioned on either one alone, and dropping the boxes also drops their warp.
To support generation without layouts, generation samples also drop the whole layout with probability 1/3 for single scans and 1/2 for aggregated scans.
Generation samples receive no ego pose, as they have no references to warp.
Layout-conditioned generation uses classifier-free guidance~\cite{ho2022classifier} with a scale of 3.
Its unconditional branch drops the layout and keeps all other inputs. Prediction uses no guidance.
Each sample's noise level follows a logit-normal distribution with a time shift of 3.0~\cite{esser2024scaling}, and the 10 Euler steps of inference use the same shift.
Evaluation uses an exponential moving average of the world model weights with decay 0.9998.
On each dataset, one fine-tuned checkpoint produces the generation results, with and without layouts, and the prediction results in Tables~\ref{tab:uncond_generation} to~\ref{tab:sequence_generation}.
The shared autoencoder decodes the latents along unseen scan patterns without further training.

\noindent\textbf{Ray loss.}
For samples with $\sigma\leq\sigma_r=0.6$, the ray loss $\mathcal{L}_{\mathrm{ray}}$ decodes 2,048 rays per frame, along the fixed beam grid of the autoencoder for single scans and a $0.05^\circ$ grid for aggregated scans.
The hit term is the binary cross-entropy between the hit probabilities decoded from $\hat{\mathbf{z}}_0$ and $\mathbf{z}$.
The bin term is a soft cross-entropy over depth bins, normalized by $\log B$ and weighted by the hit probability from $\mathbf{z}$.
It constrains depth indirectly, through the bin that sets the preliminary depth, but does not compare offsets or refined depths.
Within $\lambda_r\mathcal{L}_{\mathrm{ray}}$, the hit and bin terms have weights 0.03 and 0.02.

\subsection{Evaluation Data}
\label{sec:app_evaldata}
KITTI-360 scans are captured with a Velodyne HDL-64E and nuScenes scans with a Velodyne HDL-32E.

\noindent\textbf{KITTI-360.}
Generation in Table~\ref{tab:uncond_generation} and reconstruction in Table~\ref{tab:vae_fidelity} use 2,000 scans sampled uniformly along sequence 0018's trajectory.
No compared method trains on this sequence, whereas several baselines train on sequence 0000 or 0002.
Object, occupancy, and sequence experiments require 3D boxes or occupancy labels, which sequence 0018 lacks.
These experiments use sequences 0000 and 0002 instead.
Object experiments in Table~\ref{tab:object_generation} use 2,000 scans sampled the same way from these sequences.
SSCBench occupancy labels~\cite{li2024sscbench} cover only 362 of these 2,000 scans.
The KITTI-360 blocks of Table~\ref{tab:occ_oracle} score every method on these 362 scans.
The autoencoder ablation in Table~\ref{tab:ablation} scores reconstruction as in Table~\ref{tab:vae_fidelity}, but on sequences 0000 and 0002.

\noindent\textbf{nuScenes.}
Tables~\ref{tab:uncond_generation}, \ref{tab:occ_oracle}, \ref{tab:vae_fidelity}, and~\ref{tab:ablation} use all 6,019 validation keyframes.
The released nuScenes checkpoints of T2LDM and T2LDM++ also train on these keyframes (\tabref{tab:method_inventory}), whereas the other compared nuScenes models never train on the validation split.
Object experiments in Table~\ref{tab:object_generation} use 2,400 keyframes, taken from two non-overlapping windows of eight keyframes in each of the 150 validation scenes.

\noindent\textbf{HeLiPR.}
The four unseen scan patterns come from the HeLiPR~\cite{jung2024helipr} sensors, the rotating Ouster OS2-128 and Velodyne VLP-16, the Aeva Aeries~II, and the Livox Avia with a non-repetitive pattern.
We refer to them as OS2-128, VLP-16, Aeries~II, and Avia.

\noindent\textbf{Sequences.}
Sequence experiments in Table~\ref{tab:sequence_generation} use 250 KITTI-360 windows and 326 nuScenes windows.
The KITTI-360 windows start at evenly spaced positions along each sequence and do not overlap.
On nuScenes, we split each validation scene after its first five keyframes into windows of five generated keyframes.
We drop the windows that overlap the last window generated by UniScene and keep every third of the rest, which leaves 326 windows, one to three per scene.
Each window contains three observed and five generated frames, 0.1\,s apart on KITTI-360 and 0.5\,s apart on nuScenes.
The world model ablation in Table~\ref{tab:wm_ablation} uses the same KITTI-360 windows.

\subsection{Metrics and Evaluators}
\label{sec:app_metrics}
Table~\ref{tab:metric_summary} lists each metric's full name, formula, and evaluator.

\noindent\textbf{Common protocol.}
All methods are evaluated in their dataset's sensor coordinate convention within a common crop defined by range $r=\|\mathbf{x}\|_2$ and elevation $\phi$.
Following the range image setting of LiDM~\cite{ran2024lidm} and R3DPA~\cite{sereyjol2026r3dpa}, we use $3<r<56$\,m and $-25^\circ\leq\phi\leq3^\circ$ on KITTI-360 and $3<r<45$\,m and $-30^\circ\leq\phi\leq10^\circ$ on nuScenes.
The KITTI-360 blocks of Table~\ref{tab:occ_oracle} further restrict evaluation to the $51.2\times51.2\times6.4$\,m volume in front of the vehicle covered by SSCBench labels.
Generation metrics score each method's output points without normalizing scan density.
Our generated scans are decoded along the fixed beam grid of Appendix~\ref{sec:app_train}.
All reported CD values cap nearest-neighbor distances at 0.2\,m to prevent a few distant points from dominating.

\noindent\textbf{Distribution distances.}
$\mathcal{G}$ and $\mathcal{M}$ denote generated and measured feature distributions, with means $\mu$ and covariances $\Sigma$.
The Fr\'echet distance is $\mathrm{FD}=\|\mu_{\mathcal{G}}-\mu_{\mathcal{M}}\|_2^2+\operatorname{tr}\bigl(\Sigma_{\mathcal{G}}+\Sigma_{\mathcal{M}}-2(\Sigma_{\mathcal{G}}\Sigma_{\mathcal{M}})^{1/2}\bigr)$.
Feature MMD draws $m$ generated features $x_i$ and $m$ measured features $y_j$ without replacement, with $m$ the smaller feature count capped at 1,000, and computes
\begin{equation*}
    U_k=\frac{\sum_{i\neq j}\bigl[k(x_i,x_j)+k(y_i,y_j)\bigr]}{m(m-1)}-\frac{2}{m^2}\sum_{i,j}k(x_i,y_j).
\end{equation*}
It averages $U_k$ over 100 such draws.
\ac{BEV} MMD uses the normalized histograms $p_i$ and $q_j$ of all $n_{\mathcal{G}}$ generated and $n_{\mathcal{M}}$ measured scans and includes the diagonal terms within each set,
\begin{equation*}
    V_k=\frac{\sum_{i,j}k(p_i,p_j)}{n_{\mathcal{G}}^2}+\frac{\sum_{i,j}k(q_i,q_j)}{n_{\mathcal{M}}^2}-\frac{2\sum_{i,j}k(p_i,q_j)}{n_{\mathcal{G}}n_{\mathcal{M}}}.
\end{equation*}
Neither MMD takes a square root.

\begin{table}[t]
    \centering
    \def\Ta{\ref{tab:uncond_generation}}\def\Tb{\ref{tab:object_generation}}\def\Tc{\ref{tab:vae_fidelity}}
    \def\Td{\ref{tab:occ_oracle}}\def\Te{\ref{tab:sequence_generation}}
    \def\Y{$\checkmark$}\def\N{$\times$}
    \caption{Compared methods. A checkmark denotes a checkpoint that generates scenes of the dataset. Training data are those of these checkpoints. Odom.\ marks KITTI Odometry checkpoints and sequences, and other sequence numbers refer to KITTI-360.}
    \label{tab:method_inventory}
    \scriptsize
    \setlength{\tabcolsep}{2.2pt}
    \renewcommand{\arraystretch}{1.05}
    \resizebox{\linewidth}{!}{\begin{tabular}{@{}lcccccc@{}}
        \toprule
        & \multicolumn{2}{c}{Checkpoint} & \multicolumn{2}{c}{Training data} & \multicolumn{2}{c}{Tables in Sec.~\ref{sec:evaluation}} \\
        \cmidrule(lr){2-3}\cmidrule(lr){4-5}\cmidrule(lr){6-7}
        Method & KITTI-360 & nuScenes & KITTI-360 & nuScenes & KITTI-360 & nuScenes \\
        \midrule
        \ours & \Y & \Y & 3--7, 9, 10 & train & \ref{tab:uncond_generation}--\ref{tab:wm_ablation} & \ref{tab:uncond_generation}--\ref{tab:ablation} \\
        \midrule
        \multicolumn{7}{@{}l}{\textit{Range image models}} \\
        LiDARGen~\cite{zyrianov2022lidargen} & \Y & \N & 3--7, 9, 10 & -- & \Ta & -- \\
        R2DM~\cite{nakashima2024lidar} & \Y & \N & 3--7, 9, 10 & -- & \Ta, \Tb & -- \\
        R2Flow~\cite{nakashima2025r2flow} & \Y & \N & 3--7, 9, 10 & -- & \Ta, \Tb & -- \\
        LiDM~\cite{ran2024lidm} & \Y & \N & 0, 2, 4--10 & -- & \Ta, \Tc & -- \\
        R3DPA~\cite{sereyjol2026r3dpa} & \Y & \N & 3--7, 9, 10 & -- & \Ta, \Tb, \Tc & -- \\
        T2LDM~\cite{qu2026t2ldm} & \Y & \Y & 0, 2--7, 9, 10 & train, val & \Ta & \Ta \\
        T2LDM++~\cite{qu2026t2ldmpp} & \Y & \Y & 0, 2--7, 9, 10 & train, val & \Ta & \Ta \\
        GEM~\cite{wu2026gem} & Odom. & \N & Odom.\ 00--07 & -- & \Tc & -- \\
        ST3DCNN~\cite{mersch2021st3dcnn} & Odom. & \N & Odom.\ 00--05 & -- & -- & -- \\
        \midrule
        \multicolumn{7}{@{}l}{\textit{Rendering-based models}} \\
        LidarDM~\cite{zyrianov2025lidardm} & \Y & \N & 2--7, 9, 10 & -- & \Ta, \Td & -- \\
        OpenDWM-DiT~\cite{opendwm} & \Y & \Y & 3--7, 9, 10 & train & \Ta, \Tb, \Te & \Ta, \Tb, \Te \\
        OpenDWM-MaskGIT~\cite{opendwm} & \Y & \Y & 3--7, 9, 10 & train & \Tb, \Te & \Tb, \Te \\
        OpenDWM (VQ, VAE)~\cite{opendwm} & \Y & \Y & 3--7, 9, 10 & train & \Td, \Tc & \Td, \Tc \\
        UniScene~\cite{li2025uniscene} & \N & \Y & -- & train & -- & \Ta, \Tb, \Te, \Td, \Tc \\
        \midrule
        \multicolumn{7}{@{}l}{\textit{Per-scene representations (Appendix~\ref{sec:app_occ_neural})}} \\
        GS-LiDAR~\cite{jiang2025gslidar} & -- & -- & per scene & -- & -- & -- \\
        LiDAR4D~\cite{zheng2024lidar4d} & -- & -- & per scene & -- & -- & -- \\
        \midrule
        \multicolumn{7}{@{}l}{\textit{Not compared, no released checkpoint for scenes}} \\
        LiSTAR~\cite{liu2025listar} & \N & \N & -- & -- & -- & -- \\
        LiDARCrafter~\cite{liang2025lidarcrafter} & \N & \N & -- & -- & -- & -- \\
        Copilot4D~\cite{zhang2024copilot4d} & \N & \N & -- & -- & -- & -- \\
        U4D~\cite{xu2026u4d} & \N & \N & -- & -- & -- & -- \\
        \bottomrule
    \end{tabular}}
    \vspace{-6mm}
\end{table}
\noindent\textbf{Object evaluators.}
nuScenes object metrics use LiDARCrafter's released Voxel R-CNN detector~\cite{deng2021voxelrcnn}, PointMLP classifier~\cite{ma2022pointmlp}, and GLENet box estimator~\cite{zhang2023glenet,liang2025lidarcrafter}.
Because LiDARCrafter releases no KITTI-360 evaluators, we train the same networks on annotated scans from training sequences 0003 to 0007, 0009, and 0010.
We compare object scores only within each dataset.

\noindent\textbf{Sequence metrics.}
TTCE follows LiDARCrafter's definition, frame pairs $(t,t+k)$, and measured poses, changing only registration.
Since TTCE averages over all frame pairs, it also includes failed registrations, which add large errors whether they come from inconsistent scans or from the registration itself.
LiDARCrafter uses \ac{ICP}, but ego motion spans about 2 to 4\,m on KITTI-360 and 8 to 11\,m on nuScenes over these gaps.
In this setting, ICP without an initial pose fails even on measured frames and yields a TTCE$_3$ of 0.559\,m on KITTI-360.
To minimize failures of the registration itself, we instead use KISS-Matcher~\cite{lim2025kissmatcher} registration for an initial estimate without ground truth, then refine it with coarse-to-fine generalized ICP~\cite{segal2009gicp}.
We tune this registration on measured scans only and apply it unchanged to every method.
On the measured scans of all evaluation windows, it lowers the fraction of pairs with translation errors above 1\,m from 29\% with ICP to 0.1\% on KITTI-360 and from 76\% to 1.7\% on nuScenes.
CTC replaces LiDARCrafter's squared nearest-neighbor distances with capped CD.

\noindent\textbf{Ray metrics.}
We compare depths and returns in range images, which are grids of measured beam directions with $64\times1024$ cells on KITTI-360 and $32\times1024$ on nuScenes, one row per beam.
Each return is placed in the cell nearest to its direction, and each cell keeps the return closest to the sensor, both for the measured scan and for every method.
Methods that accept arbitrary rays are queried along the rays that the sensor fired.
A ray with a return points to that return, and a ray without one uses the elevation of its beam at the firing azimuth.
nuScenes scans store the beam of each point, and on KITTI-360 we recover it from the order of the points in each scan~\cite{triess2020scan}.
Because queried methods are read only along these fired rays, they place no return in grid cells that no fired ray reaches.
Methods that output their own range images or point clouds instead place their points in the grid by the same rule, including in such cells.
CD and $F_{0.1}$ are computed from these range images~\cite{khurana2023point,zhang2024copilot4d}, with each kept return placed along its cell direction at its depth.
CTC and the CD of Appendix~\ref{sec:app_pred_gem} instead use raw point sets.
CD is averaged over scans, or over predicted frames in prediction.
$F_{0.1}$ and H-F1 are computed once from the matched points and hit decisions pooled over all scans.
dMAE, MedAE, RMSE, and AbsRel likewise pool the cells where both range images return over all scans.

\subsection{Compared Methods}
\label{sec:app_baselines}
Table~\ref{tab:method_inventory} summarizes the compared methods.
Appendix~\ref{sec:app_occ_neural} also compares the per-scene representations GS-LiDAR and LiDAR4D, which we optimize on 51 measured scans of each scene.
For prediction, Appendix~\ref{sec:app_pred_gem} also compares GEM and ST3DCNN, using their released models trained on KITTI Odometry~\cite{geiger2012kitti}.
Range image baselines convert their range images to points with their released projections.
Table~\ref{tab:object_generation} excludes unconditional baselines trained on its evaluation frames.
LidarDM requires trajectory data to insert moving agents.
As its released KITTI-360 model lacks these data, its generated scenes contain no moving agents.
OpenDWM-MaskGIT generates tokens of its VQ autoencoder, and OpenDWM-DiT generates latents of its VAE.
The voxel heads of both decoders place a point in each occupied voxel.
The VAE also has a render head that decodes a separate density field and reads the depth along each ray by volume rendering.

\section{Model Architecture Details}
\label{sec:app_model}
\begin{figure}[t]
    \centering
    \includegraphics[width=\linewidth]{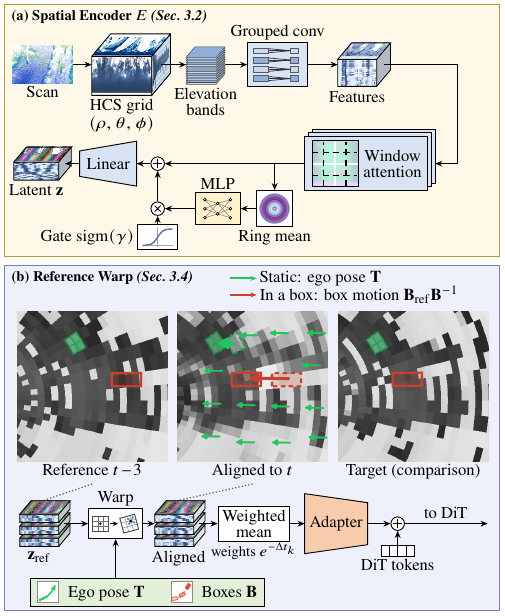}
    \caption{\textbf{Encoder and reference warp.}
    (a) The encoder maps a KITTI-360 point cloud to the latent $\mathbf{z}$.
    (b) Around a moving car, a reference latent is warped to the target frame.
    Arrows show where the content of each cell moves, and darker cells deviate more from the mean latent at their radius.}
    \label{fig:app_model}
    \vspace{-3mm}
\end{figure}

\figref{fig:app_model} illustrates the encoder and the reference warp.

\noindent\textbf{Encoder.}
The \ac{HCS} grid divides the radius up to 80\,m, the azimuth, and the elevation uniformly into 80, 640, and 160 bins.
The count channel is $\log(1+\min(N,128))/\log 129$ for $N$ points, and empty cells contain zeros in all three channels.
Elevation spans $[-0.6,0.7]$\,rad in 16 bands of 10 bins.
Within each band, a grouped $1\times8$ convolution with stride $(1,8)$ reduces azimuth to 80 bins.
The backbone has three stages of $8\times8$ window attention~\cite{liu2021swin} with depths 2, 2, and 4 and widths 128, 256, and 512.
All stages keep the $80\times80$ grid, and $1\times1$ convolutions change the width between stages.
Windows shift cyclically along radius and azimuth without attention masks, and the rotary encodings~\cite{su2024roformer} of radius and azimuth in all stages shift with the tokens.
For the ring context of Sec.~\ref{sec:encoder}, each of the 16 rings spans five radial rows.
The azimuthal average of each ring passes through RMS normalization~\cite{zhang2019rmsnorm} and a bottleneck \ac{MLP}.
A learnable scalar $\gamma$, initialized to $-3$, scales the output by $\operatorname{sigm}(\gamma)\approx0.05$ at the start of training.
Two linear layers map the 512 features of each cell to the mean and log variance of the 32 latent channels.

\noindent\textbf{Decoder.}
The decoder uses $B=64$ depth bins over $[2.4, 72]$\,m.
An \ac{MLP} embeds each ray direction together with the sines and cosines of its azimuth and elevation.
A linear layer maps each of the $80\times80$ latent cells to a 384-dimensional token.
The decoder adds no positional encoding to these tokens, so the global attention relies on the location cues that the encoder leaves in their content.
The local refinement instead uses position explicitly, as $\pi(\mathbf{x}_i)$ sets where $\mathbf{Z}$ is sampled and $\Delta$ receives $\mathbf{x}_i$ itself.
Six cross-attention blocks with six heads query these tokens with the ray embedding.
The local feature map $\mathbf{Z}$ has $C_f=128$ channels.
$\Delta$ sums four 128-channel terms.
They are $\mathbf{f}_i$, a learned embedding of $b_i$, a linear projection of $\mathbf{t}_i$, and a linear projection of the concatenation of $\mathbf{x}_i/72$ and the normalized bin center $(b_i+1/2)/B$.
A LayerNorm, a SiLU, a 128-channel linear layer, another SiLU, and an output linear layer map this sum to $\delta_i$.
The output layer is initialized to zero, so refinement starts from the preliminary depth.
For the term of $\mathcal{L}_{\mathrm{depth}}$ at the ground-truth bin in Sec.~\ref{sec:training}, the decoder recomputes $a_i$, $\mathbf{x}_i$, $\mathbf{f}_i$, $\delta_i$, and $\hat{d}_i$ at the ground-truth bin instead of the predicted bin.
The offset $o_i$ is one scalar per ray and is shared by both bins. No loss supervises $d^0_i$ directly, so $o_i$ learns from both depth terms and through the lookup position $\mathbf{x}_i$.

\noindent\textbf{World model.}
The \ac{DiT} has 12 blocks of width 1280 with 20 heads, and its attention layers normalize queries and keys with RMS normalization.
With a patch size of 2, each $80\times80$ latent frame becomes $40\times40$ tokens.
Its temporal attention connects each token position across the eight frames of a window.
Each frame receives only an index embedding, so the frame spacing is given only through the layouts of the frames and the ego poses and boxes used in the warp, when they are given.

\noindent\textbf{Layout encoder.}
The box and map rasters have three color channels each and are drawn at $640\times640$ pixels with 6.4\,px/m.
The encoder resamples them onto a $640\times640$ $(\rho,\theta)$ grid and folds each $8\times8$ patch into channels, which gives 384 channels on the $80\times80$ latent grid.
A $3\times3$ convolution maps these channels to 512.
A third raster, computed from the 3D boxes alone, stores ten box attributes per cell directly on a $320\times320$ $(\rho,\theta)$ grid.
For each cell inside a box, six channels encode object class as car, truck or trailer, bus, construction or emergency vehicle, two-wheeler, or pedestrian, and four store box bottom height, height, and yaw cosine and sine.
Cells inside several boxes take the box with the smallest footprint.
This raster is folded onto the same $80\times80$ grid and added to the first features through its own $3\times3$ convolution, initialized to zero.
A stride-2 convolution then reduces the features to the $40\times40$ token grid, and three residual blocks follow.
Twelve $1\times1$ convolutions initialized to zero, one per \ac{DiT} block, project the result to 1280 channels and add it to the input tokens of their block.
All convolutions pad circularly in azimuth, and the outputs are zero when the layout is dropped.

\noindent\textbf{Reference warp.}
For reference frame $k$ at a gap of $\Delta t_k$ frames, $\tilde{\mathbf{z}}_k$ denotes its warp by Eq.~\eqref{eq:warp}, and $m_k(\rho,\theta)\in\{0,1\}$ marks the cells where this warp is used.
A warp is used only if its source lies within the grid's radius range.
An ego warp is also dropped inside target boxes matched to a reference box, where the box warp replaces it, and wherever its source lies inside a reference box.
The box motion acts on the horizontal coordinates of $\mathbf{x}_{\rho\theta}$, which lies at the height of the sensor origin, and scales them by the ratio of the box sizes.
Box regions span all heights of their cells, and where target boxes overlap, their valid box warps are averaged with equal weight.
Since elevation is folded into the channels, the warp keeps the elevation of each cell's content when its radius changes, which is exact only at the height of the sensor origin.
This approximation affects only the condition, since the world model predicts the target latent itself.
The warped latents are averaged per cell as
\begin{equation*}
    \bar{\mathbf{z}}=\frac{\sum_k w_k\,\tilde{\mathbf{z}}_k}{\sum_k w_k},\qquad w_k=e^{-\Delta t_k}\,m_k,
\end{equation*}
and cells that no warp reaches are set to zero.
Along with $\bar{\mathbf{z}}$, the adapter of Sec.~\ref{sec:worldmodel} receives a binary map of the cells that at least one warp reaches, its complement, and a map counting the warps that reach each cell.
These three maps allow the adapter to distinguish unreached cells, whose average is zero, from cells with warped content.
The adapter stacks the 32 latent channels of $\bar{\mathbf{z}}$ and these three maps into 35 channels and averages them over $2\times2$ cells to reach the token grid.
It then applies two $1\times1$ convolutions with SiLU in between, the second initialized to zero.
In every \ac{DiT} block, its output is added to the tokens after spatial attention and the feed-forward network, before temporal attention.

\section{Box Agreement on nuScenes}
\label{sec:app_cond}
We quantify box agreement over all 2,400 nuScenes keyframes of \tabref{tab:object_generation}, extending the single frame of \figref{fig:det_cmp}.
In each frame, the detector of \tabref{tab:object_generation} runs on the measured scan and on the scan generated by each method.
Conditioning cars within 20\,m are matched one-to-one to detections scoring above 0.3 so that the total \ac{BEV} IoU is maximal.
\begin{table}[t]
    \centering
    \caption{Box agreement on nuScenes. Share of conditioning cars within 20\,m whose matched detection reaches each IoU, in \%.}
    \label{tab:app_box_agreement}
    \scriptsize
    \renewcommand{\arraystretch}{0.94}
    \begin{tabularx}{\linewidth}{@{}l*{2}{>{\centering\arraybackslash}X}@{}}
        \toprule
        Method & 3D IoU $\geq 0.7$ $\uparrow$ & \ac{BEV} IoU $\geq 0.5$ $\uparrow$ \\
        \midrule
        OpenDWM-MaskGIT & 22.7 & 63.2 \\
        OpenDWM-DiT & \csecond 27.8 & \csecond 79.0 \\
        UniScene & 4.3 & 40.5 \\
        \ours & \cfirst\textbf{56.8} & \cfirst\textbf{82.2} \\
        \textit{Measured scans (Ref.)} & 80.2 & 91.0 \\
        \bottomrule
    \end{tabularx}
    \vspace{-3mm}
\end{table}
In \tabref{tab:app_box_agreement}, our scans achieve the highest match rates among generated scans at both IoU thresholds, and matches at a 3D IoU of at least 0.7 are about twice as frequent as on OpenDWM-DiT scans.

\section{Returns from Scene Representations}
\label{sec:app_occ}
In \tabref{tab:occ_oracle}, returns from scene representations are less accurate than our decoder's.
This section describes how these returns are read, tests whether variants of occupancy and meshes close this gap, compares depth on the rays where each method returns, and compares our decoder with per-scene neural representations.

\subsection{Reconstruction and Rendering}
\label{sec:app_occ_proto}
\noindent\textbf{Occupancy grids.}
The occupancy rows of \tabref{tab:occ_oracle} use the released Occ3D-nuScenes labels~\cite{tian2023occ3d} with 0.4\,m voxels and SSCBench-KITTI-360 labels~\cite{li2024sscbench} with 0.2\,m voxels.
Voxels not labeled free or unknown are occupied.
We march each fired ray from the sensor origin through every voxel it traverses.
The return depth is the projection of the first occupied voxel's center onto the ray.
Using the point where the ray enters this voxel instead gives a larger depth error.
We also test twice the label voxel size, where a voxel is occupied if any of its $2\times2\times2$ label voxels is occupied.
UniScene renders the same Occ3D labels with its own renderer.
In addition, we build accumulated occupancy on KITTI-360 from the static map of all scans in the sequence, as for our training labels in Appendix~\ref{sec:app_data}.
From this map and the target scan, we keep only the nearest point in each $0.05^\circ$ direction cell viewed from the target sensor.
This removes points behind visible surfaces.
We voxelize the remaining points at several voxel sizes, and each grid keeps the voxels that contain at least $k$ points, written $\geq k$.
Rays are marched through these grids as through the released labels.

\noindent\textbf{Meshes.}
The mesh rows of \tabref{tab:occ_oracle} reconstruct NKSR meshes~\cite{huang2023nksr} with 0.15\,m voxels using the released LidarDM pipeline.
Each mesh fuses the scans within 0.75\,s of the target.
On KITTI-360, these are the 15 scans centered on the target, as in LidarDM.
On nuScenes, a median of 30 scans falls within this time span.
Following LidarDM, we place meshes of moving objects from its asset bank at their annotated boxes.
We then raycast these meshes along the fired rays.
On KITTI-360, LidarDM itself instead raycasts the mesh on its range image raster, and its learned ray drop model removes the returns it predicts to be dropped.
In addition, we build VDB meshes on KITTI-360 by fusing up to 300 nearby scans into a truncated signed distance field (TSDF) with VDBFusion~\cite{vizzo2022vdbfusion}.
These scans come from sensor poses within 60\,m of the target pose, including other passes through the same place.
Each scan, including the target scan, is undistorted, cleared of dynamic objects, and integrated from its own sensor origin.
Moving objects are not reinserted, but they cover only 0.4\% of the scored returns.
Mesh extraction uses every voxel with nonzero fusion weight by default and only voxels with weight at least 5 in the rows marked $w\geq5$.
The VDB variant with 15 scans instead fuses the 15 consecutive scans centered on the target with 0.15\,m voxels, as in LidarDM.
All VDB meshes are raycast in the same way as the NKSR meshes.

\noindent\textbf{Latents.}
OpenDWM and our model encode the measured scan into latents and read returns from them.
The OpenDWM render head returns a point on a ray when the accumulated opacity along the ray exceeds 0.5.
Our decoder returns a point when its hit probability exceeds 0.5, as analyzed in Appendix~\ref{sec:app_rays_pred}.

\subsection{Occupancy and Mesh Variants}
\label{sec:app_occ_var}
\begin{table}[t]
    \centering
    \caption{Occupancy and mesh variants on KITTI-360. The first block follows the protocol of \tabref{tab:occ_oracle}, and the second that of \tabref{tab:ablation}. Size counts grid cells, triangles, points, or latent values per scene.}
    \label{tab:app_occ_kitti}
    \scriptsize
    \setlength{\tabcolsep}{2.4pt}
    \renewcommand{\arraystretch}{0.94}
    \resizebox{\linewidth}{!}{\begin{tabular}{@{}llccccc@{}}
        \toprule
        Scene (resolution) & Method & Size & CD $\downarrow$ & $F_{0.1}$ $\uparrow$ & dMAE $\downarrow$ & AbsRel $\downarrow$ \\
        \midrule
        \multicolumn{7}{@{}l}{\textbf{SSCBench volume} (362 scans)} \\
        Occ., SSCBench 0.4\,m & Ray marching & 0.26M & 0.185 & 0.059 & 2.09 & 17.8 \\
        Occ., SSCBench 0.2\,m & Ray marching & 2.10M & 0.160 & 0.162 & 1.43 & 11.8 \\
        Occ., accumulated 0.2\,m, $\geq$1 & Ray marching & 2.10M & 0.166 & 0.141 & 2.06 & 16.0 \\
        Occ., accumulated 0.2\,m, $\geq$5 & Ray marching & 2.10M & 0.151 & 0.212 & 1.36 & 11.2 \\
        Occ., accumulated 0.1\,m, $\geq$3 & Ray marching & 16.8M & 0.123 & 0.410 & 0.96 & 7.8 \\
        Occ., accumulated 0.05\,m, $\geq$1 & Ray marching & 134M & 0.120 & 0.456 & 1.11 & 8.4 \\
        Occ., accumulated 0.05\,m, $\geq$2 & Ray marching & 134M & 0.106 & 0.546 & 0.72 & 5.9 \\
        Occ., accumulated 0.05\,m, $\geq$3 & Ray marching & 134M & 0.099 & 0.595 & 0.58 & 5.1 \\
        Mesh, NKSR 0.15\,m & Raycasting & 0.50M & \csecond 0.073 & \csecond 0.741 & 0.51 & 4.1 \\
        Mesh, VDB 0.1\,m & Raycasting & 6.29M & 0.090 & 0.645 & 0.95 & 7.1 \\
        Mesh, VDB 0.1\,m, $w\geq5$ & Raycasting & 1.77M & 0.081 & 0.703 & 0.52 & 4.3 \\
        Mesh, VDB 0.05\,m & Raycasting & 24.8M & 0.084 & 0.676 & 0.75 & 5.8 \\
        Mesh, VDB 0.05\,m, $w\geq5$ & Raycasting & 5.11M & 0.075 & 0.737 & 0.43 & 3.8 \\
        Mesh, VDB 0.15\,m, 15 scans & Raycasting & 0.51M & 0.075 & 0.732 & 0.59 & 4.6 \\
        Latent, OpenDWM & Render head & 0.20M & 0.081 & 0.704 & \csecond 0.33 & \csecond 3.1 \\
        Latent, \ours & Ray decoder & 0.20M & \cfirst\textbf{0.055} & \cfirst\textbf{0.817} & \cfirst\textbf{0.24} & \cfirst\textbf{2.0} \\
        \midrule
        \multicolumn{7}{@{}l}{\textbf{All scans} (2,000 scans of sequences 0000 and 0002, $112\times112\times8$\,m grid)} \\
        Occ., accumulated 0.4\,m, $\geq$1 & Ray marching & 1.57M & 0.183 & 0.074 & 3.30 & 23.9 \\
        Occ., accumulated 0.2\,m, $\geq$1 & Ray marching & 12.5M & 0.169 & 0.133 & 2.35 & 16.8 \\
        Occ., accumulated 0.1\,m, $\geq$1 & Ray marching & 100M & 0.146 & 0.267 & 1.75 & 12.4 \\
        Occ., accumulated 0.1\,m, $\geq$3 & Ray marching & 100M & 0.132 & 0.355 & 1.16 & 8.8 \\
        Occ., accumulated 0.1\,m, $\geq$8 & Ray marching & 100M & 0.120 & 0.436 & \csecond 0.86 & \csecond 7.3 \\
        Points, accumulated & Nearest point & 2.79M & \csecond 0.099 & \csecond 0.583 & 0.99 & 8.5 \\
        Latent, \ours & Ray decoder & 0.20M & \cfirst\textbf{0.062} & \cfirst\textbf{0.785} & \cfirst\textbf{0.31} & \cfirst\textbf{2.3} \\
        \bottomrule
    \end{tabular}}
    \vspace{-3mm}
\end{table}
\begin{table}[t]
    \centering
    \caption{Depth on range image cells where both the method and the measured scan return, on the full scans of \tabref{tab:occ_oracle}. The last two columns give the percentages of rays within 0.1\,m and of rays stopping more than 1\,m early.}
    \label{tab:app_occ_gtdir}
    \scriptsize
    \setlength{\tabcolsep}{3pt}
    \renewcommand{\arraystretch}{0.94}
    \resizebox{\linewidth}{!}{\begin{tabular}{@{}llcccc@{}}
        \toprule
        Scene (resolution) & Method & dMAE $\downarrow$ & MedAE $\downarrow$ & $\leq$0.1\,m $\uparrow$ & Early $\downarrow$ \\
        \midrule
        \multicolumn{6}{@{}l}{\textbf{KITTI-360} (SSCBench volume, 362 scans)} \\
        Occ., SSCBench 0.2\,m & Ray marching & 1.43 & 0.555 & 6.2 & 30.5 \\
        Occ., SSCBench 0.4\,m & Ray marching & 2.09 & 1.057 & 3.2 & 52.7 \\
        Occ., accumulated 0.1\,m, $\geq$3 & Ray marching & 0.96 & 0.278 & 16.4 & 16.5 \\
        Mesh, NKSR 0.15\,m & Raycasting & 0.51 & \csecond 0.057 & \csecond 66.6 & 6.7 \\
        Mesh, VDB 0.05\,m, $w\geq5$ & Raycasting & 0.43 & 0.064 & 65.1 & 5.1 \\
        Latent, OpenDWM & Render head & \csecond 0.33 & 0.085 & 56.4 & \csecond 3.3 \\
        Latent, \ours & Ray decoder & \cfirst\textbf{0.24} & \cfirst\textbf{0.035} & \cfirst\textbf{74.1} & \cfirst\textbf{2.1} \\
        \midrule
        \multicolumn{6}{@{}l}{\textbf{nuScenes} (6,019 scans)} \\
        Occ., Occ3D 0.4\,m & Ray marching & 1.88 & 0.557 & 9.2 & 27.7 \\
        Occ., Occ3D 0.8\,m & Ray marching & 3.25 & 1.668 & 3.1 & 68.3 \\
        Occ., Occ3D 0.4\,m & UniScene & 1.08 & 0.350 & 18.1 & 8.3 \\
        Mesh, NKSR 0.15\,m & Raycasting & 0.99 & 0.045 & 66.3 & 11.8 \\
        Latent, OpenDWM & Render head & \csecond 0.36 & \csecond 0.043 & \csecond 74.8 & \csecond 3.3 \\
        Latent, \ours & Ray decoder & \cfirst\textbf{0.29} & \cfirst\textbf{0.033} & \cfirst\textbf{75.8} & \cfirst\textbf{2.3} \\
        \bottomrule
    \end{tabular}}
    \vspace{-3mm}
\end{table}
To test whether better occupancy or meshes close the gap, \tabref{tab:app_occ_kitti} adds the accumulated occupancy and VDB meshes of Appendix~\ref{sec:app_occ_proto} on KITTI-360.

\noindent\textbf{Occupancy grids.}
The first block of \tabref{tab:app_occ_kitti} varies the point threshold and the voxel size of the accumulated occupancy.
At the 0.2\,m resolution of the released labels, we make the accumulated grid either stricter or looser.
The strict grid keeps only voxels with at least five points and is only slightly more accurate than the released labels.
It still falls far short of our decoder.
The loose grid keeps every voxel with at least one point and is worse than the released labels on every metric.
Most of its early stops occur at the extra voxels it keeps.
In contrast, shrinking the accumulated voxels to 0.05\,m lowers CD and raises $F_{0.1}$ over the released labels.
Even so, these finer grids still fall short of our decoder.
At this resolution, a stricter point threshold lowers dMAE, as it removes sparse voxels that end rays early.
Moreover, such a grid contains about 655 times as many values as our latent.
To reach even this accuracy with such a grid, a world model would have to generate or decode far more values than ours.
Occupancy voxels quantize surfaces, and only much finer and larger grids reduce this error.
However, any occupied voxel ahead of a surface ends a ray, so each added voxel can stop a ray early.
Therefore, finer resolution and stricter thresholds narrow the gap, but no tested occupancy grid closes it.

\noindent\textbf{Fused meshes.}
We also evaluate the VDB meshes alongside NKSR, as raycasting a mesh does not stop at voxels.
The fused meshes can retain surfaces ahead of those visible from the target pose.
Typical examples are ground and curbs seen at grazing angles from other poses, vegetation, and pole edges.
Consequently, these meshes still stop early on some rays where both they and the measured scan return (\tabref{tab:app_occ_gtdir}).
Pruning voxels with low fusion weight removes some of these surfaces and lowers dMAE in \tabref{tab:app_occ_kitti}.
Still, no mesh matches the CD or $F_{0.1}$ of our decoder on these full scans.
A mesh returns on every ray that meets a surface, whereas our decoder learns from hit and miss labels whether each ray returns.
The 0.05\,m VDB meshes also have 25 to 121 times as many triangles as our latent has values.
Across all variants, our decoder remains the most accurate on every reported metric.

\noindent\textbf{All KITTI-360 scans.}
As SSCBench labels cover only part of the evaluated KITTI-360 scans, the second block of \tabref{tab:app_occ_kitti} covers all scans with grids built from the accumulated sequence.
At the finest resolution, raising the point threshold lowers CD and dMAE.
Reading the nearest accumulated point along each ray without a grid gives the lowest CD among the accumulated settings, but its depth errors remain far above ours.
Our decoder is again more accurate than every tested setting on every reported metric.

\subsection{Depth on Co-Returned Rays}
\label{sec:app_occ_depth}
\begin{figure}[t]
    \centering
    \includegraphics[width=\linewidth]{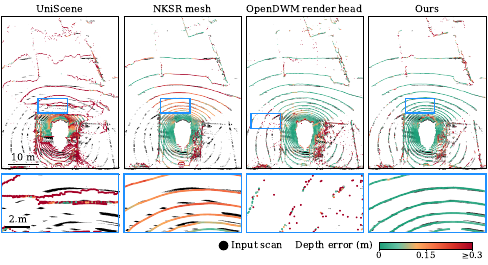}
    \caption{\textbf{Reading returns from scene representations} on a nuScenes scan in \ac{BEV}, with highlighted regions enlarged below.}
    \label{fig:app_readout_nus}
    \vspace{-3mm}
\end{figure}
The dMAE of \tabref{tab:occ_oracle} uses only the cells where both the method and the measured scan return.
On the same cells, \tabref{tab:app_occ_gtdir} adds the median error and the fractions of rays within 0.1\,m and stopping more than 1\,m early.
Our decoder has the lowest mean and median depth errors on both datasets.
As each method returns on a different set of rays, a method could lower these errors by skipping difficult rays.
However, our decoder also has the lowest CD and the highest $F_{0.1}$ on these full scans in Tables~\ref{tab:occ_oracle} and~\ref{tab:app_occ_kitti}, which also penalize missing returns.
Ray marching through occupancy suffers most from early stops.
Through the released labels, it stops more than 1\,m early on more than a quarter of these rays and on more than half at twice the released voxel size.
As in Appendix~\ref{sec:app_occ_var}, these early stops occur because ray marching ends a ray at the first occupied voxel, even one ahead of the visible surface.
Meshes, the UniScene renderer, and the OpenDWM render head stop early less often but have larger depth errors than ours.
Our decoder stops early least often of all methods.
Therefore, our advantage in \tabref{tab:occ_oracle} comes not only from deciding which rays return but also from placing each return close to the measured surface.

\begin{figure}[t]
    \centering
    \includegraphics[width=\linewidth]{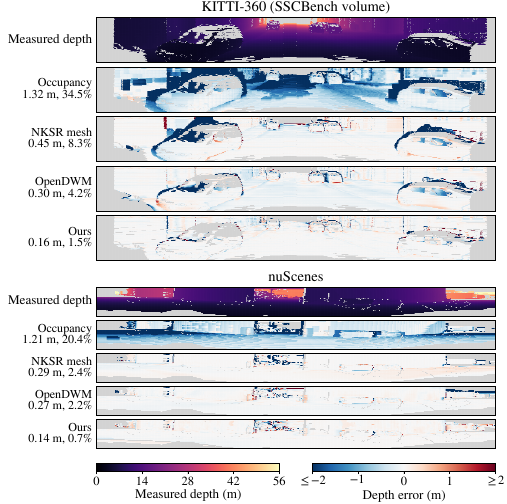}
    \caption{\textbf{Where returns are read too early or too late.}
    Signed depth error of each representation on the front $180^\circ$ of one scan per dataset, with one row per beam.
    Occupancy uses the released labels, and OpenDWM uses its render head.
    Blue rays stop in front of the measured return, red rays stop behind it, and gray rays have no return from the representation or the measured scan.
    Values give the mean absolute error and the fraction of rays stopping more than 1\,m early.}
    \label{fig:app_earlystop}
    \vspace{-2mm}
\end{figure}

\figref{fig:app_readout_nus} shows the returns of these representations along the fired rays of a nuScenes scan in \ac{BEV}.
In its close-ups, the UniScene returns run between the measured scan lines, the mesh returns run slightly beside them, and the OpenDWM render head places returns between the scan lines beside the road.
Our returns instead follow the measured scan lines.
\figref{fig:app_earlystop} further shows where the early stops occur.
Occupancy stops early across much of the ground and on distant surfaces, whereas the larger errors of the other representations lie mainly at object boundaries.

\subsection{Per-Scene Neural Representations}
\label{sec:app_occ_neural}
\begin{table}[t]
    \centering
    \caption{Per-scene neural representations and our decoder on six KITTI-360 scans of \tabref{tab:occ_oracle}, under its protocol.}
    \label{tab:app_neural}
    \scriptsize
    \setlength{\tabcolsep}{2.4pt}
    \renewcommand{\arraystretch}{0.92}
    \resizebox{\linewidth}{!}{\begin{tabular}{@{}llcccc@{}}
        \toprule
        Method & Input & CD $\downarrow$ & $F_{0.1}$ $\uparrow$ & dMAE $\downarrow$ & MedAE $\downarrow$ \\
        \midrule
        \multicolumn{6}{@{}l}{\textit{VLP-16 rows masked (6 scans)}} \\
        GS-LiDAR~\cite{jiang2025gslidar} & 51 scans & 0.093 & 0.620 & \csecond 0.343 & 0.085 \\
        LiDAR4D~\cite{zheng2024lidar4d} & 51 scans & \cfirst\textbf{0.087} & \cfirst\textbf{0.658} & \cfirst\textbf{0.297} & \cfirst\textbf{0.075} \\
        NKSR mesh, raycasting & 15 scans & \csecond 0.092 & 0.627 & 0.589 & 0.079 \\
        NKSR mesh, LidarDM~\cite{zyrianov2025lidardm} & 15 scans & 0.140 & 0.333 & 0.956 & 0.228 \\
        OpenDWM~\cite{opendwm}, render head & 1 scan & 0.104 & 0.539 & 0.519 & 0.108 \\
        \ours & 1 scan & \cfirst\textbf{0.087} & \csecond 0.644 & 0.385 & \csecond 0.076 \\
        \midrule
        \multicolumn{6}{@{}l}{\textit{Avia cells masked (6 scans)}} \\
        GS-LiDAR~\cite{jiang2025gslidar} & 51 scans & 0.093 & 0.595 & 0.396 & 0.081 \\
        LiDAR4D~\cite{zheng2024lidar4d} & 51 scans & 0.089 & 0.632 & \cfirst\textbf{0.324} & 0.075 \\
        NKSR mesh, raycasting & 15 scans & \csecond 0.082 & \cfirst\textbf{0.680} & 0.570 & \csecond 0.057 \\
        NKSR mesh, LidarDM~\cite{zyrianov2025lidardm} & 15 scans & 0.134 & 0.357 & 1.066 & 0.317 \\
        OpenDWM~\cite{opendwm}, render head & 1 scan & 0.099 & 0.582 & 0.676 & 0.093 \\
        \ours & 1 scan & \cfirst\textbf{0.080} & \csecond 0.669 & \csecond 0.364 & \cfirst\textbf{0.048} \\
        \midrule
        \multicolumn{6}{@{}l}{\textit{Full scan (6 scans)}} \\
        GS-LiDAR~\cite{jiang2025gslidar} & 51 scans & \cfirst\textbf{0.056} & \cfirst\textbf{0.810} & \csecond 0.234 & \cfirst\textbf{0.032} \\
        LiDAR4D~\cite{zheng2024lidar4d} & 51 scans & 0.064 & 0.788 & \cfirst\textbf{0.225} & 0.052 \\
        NKSR mesh, raycasting & 15 scans & 0.077 & 0.716 & 0.520 & 0.065 \\
        NKSR mesh, LidarDM~\cite{zyrianov2025lidardm} & 15 scans & 0.109 & 0.534 & 0.792 & 0.182 \\
        OpenDWM~\cite{opendwm}, render head & 1 scan & 0.081 & 0.704 & 0.313 & 0.089 \\
        \ours & 1 scan & \csecond 0.057 & \csecond 0.808 & 0.253 & \csecond 0.039 \\
        \bottomrule
    \end{tabular}}
    \vspace{-4mm}
    \vspace{-3mm}
\end{table}
\begin{figure}[t]
    \centering
    \includegraphics[width=\linewidth]{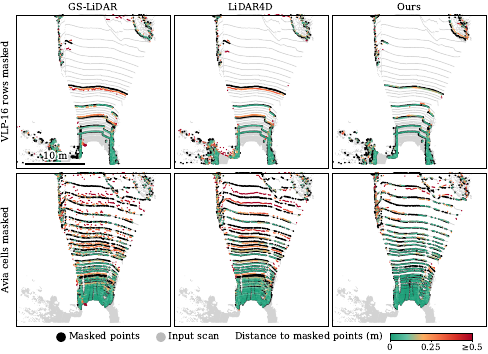}
    \caption{\textbf{Masked rays with per-scene representations} on a KITTI-360 scan of \tabref{tab:app_neural}.
    GS-LiDAR and LiDAR4D are optimized without the masked rays of this scan, and ours encodes the gray input.}
    \label{fig:app_neural}
    \vspace{-3mm}
\end{figure}
\begin{figure}[t]
    \centering
    \includegraphics[width=\linewidth]{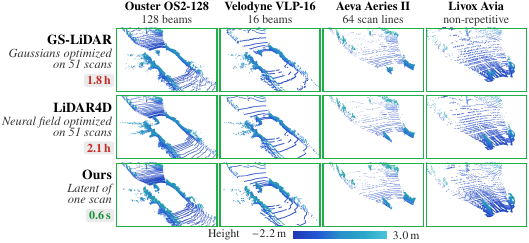}
    \caption{\textbf{Four HeLiPR scan patterns from per-scene representations and our latent} on the KITTI-360 scan of \figref{fig:app_neural}.
    Times give the optimization time of each representation and our encoding and decoding time for the four scans.}
    \label{fig:app_neural4}
    \vspace{-3mm}
\end{figure}
In our method, a scene latent is encoded from a single scan and decoded along any requested rays.
Neural scene representations such as 3D Gaussians and neural fields can also render returns along new rays once they are optimized on many scans of one scene.
We compare our decoder with GS-LiDAR~\cite{jiang2025gslidar} and LiDAR4D~\cite{zheng2024lidar4d}, each optimized on 51 measured scans per scene as in its official KITTI-360 setting.
In the masked settings, the masked rays of the target scan are removed from their supervision, as they are from our input, while the other 50 scans keep all their rays.
As optimization takes about two hours per scene (Appendix~\ref{sec:app_gen_transfer}), we use six of the KITTI-360 scans of \tabref{tab:occ_oracle} and its protocol.
\tabref{tab:app_neural} also lists the NKSR mesh and the OpenDWM render head of \tabref{tab:occ_oracle} on the same scans.
As in \tabref{tab:occ_oracle}, the mesh keeps the masked rays of the target scan.

On the masked rays, which the input lacks, our decoder reads returns from a single scan about as accurately as the per-scene representations.
It has the lowest CD with both masks, shared with LiDAR4D when the VLP-16 rows are masked, and its MedAE is the lowest with the Avia cells masked and within 0.001\,m of LiDAR4D with the VLP-16 rows masked.
LiDAR4D has the lowest dMAE in every setting and a higher $F_{0.1}$ with the VLP-16 rows masked.
Our dMAE stays higher because a few large errors at depth discontinuities raise the mean, as analyzed in Appendices~\ref{sec:app_rays_pred} and~\ref{sec:app_rays_err}.
The NKSR mesh and the OpenDWM render head have a higher dMAE than both per-scene representations, although the mesh achieves the highest $F_{0.1}$ with the Avia cells masked.
On the full scans, the per-scene representations are supervised with the evaluated scan itself.
Even so, our model, which encodes only this scan, is second to GS-LiDAR in CD, $F_{0.1}$, and MedAE, with a CD only 0.001\,m higher.

\figref{fig:app_neural} shows the masked rays of one of these scans.
With the Avia cells masked, GS-LiDAR scatters returns between the measured scan lines, and LiDAR4D places the ground scan lines slightly off the measured ones.
Our returns on this scan stay on the measured scan lines, and about 80\% of them lie within 0.1\,m of a measured return, against 59 to 73\% for the per-scene representations.
However, our decoder leaves some distant rays of the masked Avia cells without a return.
Without per-scene optimization, our model reads the masked rays from a single scan with a CD at least as low as those of the per-scene representations in \tabref{tab:app_neural}, which are optimized on 51 scans for about two hours. It nearly matches them on the full scans.
For a comparison on the same measured scene, our model decodes the four HeLiPR scan patterns from the latent of one encoded scan instead of a generated one.
\figref{fig:app_neural4} shows that it keeps these patterns, as both per-scene representations do after this long optimization.

\section{Computational Time}
\label{sec:app_gen}
This section reports the time to generate a scene together with its scans in the four HeLiPR patterns and the time to generate a single scan.

\subsection{Scenes and Scans with Four Patterns}
\label{sec:app_gen_transfer}
\begin{table}[t]
    \centering
    \caption{Time to generate one scene and decode its scans with the four HeLiPR scan patterns of \figref{fig:transfer} (333k rays) on one RTX 3090. Kept marks whether the decoded scans keep the four patterns.}
    \label{tab:app_transfer_time}
    \scriptsize
    \setlength{\tabcolsep}{2.4pt}
    \renewcommand{\arraystretch}{0.94}
    \resizebox{0.9\linewidth}{!}{\begin{tabular}{@{}llcccc@{}}
        \toprule
        Method & Representation & Kept & Gen. (s) $\downarrow$ & Decode (s) $\downarrow$ & Total (s) $\downarrow$ \\
        \midrule
        \multicolumn{6}{@{}l}{\textbf{Generators}} \\
        UniScene & Occupancy & $\times$ & 1.29 & 0.86 & 2.15 \\
        OpenDWM & Latent & $\times$ & \cfirst\textbf{0.55} & 0.47 & \cfirst\textbf{1.02} \\
        LidarDM & Mesh & $\checkmark$ & 16.35 & \cfirst\textbf{0.02} & 16.37 \\
        \ours & Latent & $\checkmark$ & \csecond 0.75 & 0.59 & \csecond 1.34 \\
        \midrule
        \multicolumn{6}{@{}l}{\textbf{Per-scene representations, optimized on 51 measured scans}} \\
        GS-LiDAR~\cite{jiang2025gslidar} & Gaussians & $\checkmark$ & 6,369 & \csecond 0.18 & 6,369 \\
        LiDAR4D~\cite{zheng2024lidar4d} & Neural field & $\checkmark$ & 7,570 & 2.65 & 7,573 \\
        \bottomrule
    \end{tabular}}
    \vspace{-3mm}
\end{table}
\begin{figure}[t]
    \centering
    \includegraphics[width=0.9325\linewidth]{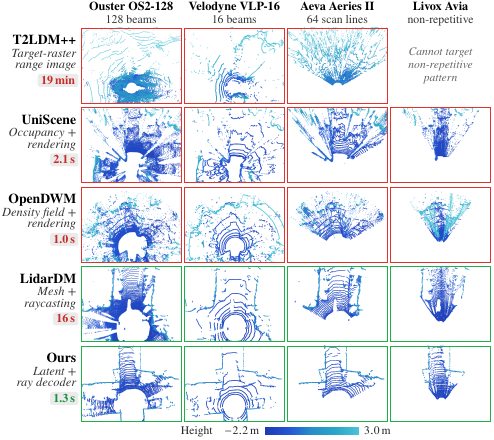}%
    \caption{\textbf{Generated scenes with the four HeLiPR scan patterns, seen from above.}
    The sensor faces up. Red frames mark broken scan patterns, and times are those of \figref{fig:transfer}.}
    \label{fig:app_transfer_bev}
    \vspace{-3mm}
\end{figure}

\tabref{tab:app_transfer_time} splits the time of the methods in \figref{fig:transfer} into generating a scene and decoding its scans with the four HeLiPR patterns.
\figref{fig:app_transfer_bev} shows such scans from above, where broken scan patterns are more visible.
Each stage is timed per scene on one machine with an RTX 3090.
For our model, generation samples the latent with the DiT, and decoding reads the four scans from it.
For OpenDWM, generation samples the latent and maps it to a density field, and decoding reads the four scans from this field with its render head.
These times differ from \tabref{tab:app_gen_time}, which measures OpenDWM's default point cloud output instead.
T2LDM++ is not in the table because each pattern needs its own sample and the Avia pattern has no raster.

\noindent\textbf{Generators.}
Among the generators that keep all four patterns, ours generates the scene and its scans in 1.3\,s, against LidarDM's 16\,s.
LidarDM quickly raycasts its generated mesh with all four patterns on the CPU, but mesh creation takes most of its time.

\noindent\textbf{Per-scene representations.}
GS-LiDAR and LiDAR4D reconstruct one measured scene from its 51 measured scans instead of generating a new one.
Although they read returns about as accurately as our decoder (Appendix~\ref{sec:app_occ_neural}), their optimization to generate each scene takes about 1.8 and 2.1 hours (\figref{fig:app_neural4} and \tabref{tab:app_transfer_time}).
Without such optimization, our model generates a scene in 0.75\,s and encodes a measured scan in 0.023\,s.
Overall, our model produces a scene with all four patterns more than ten times faster than any other method that keeps them.

\subsection{Scan Generation}
\label{sec:app_gen_scan}
\begin{table}[t]
    \centering
    \caption{Time (s) and compute (TFLOPs) per generated scan, and achieved throughput (TFLOP/s), on one RTX 3090. NFE is the number of network evaluations.}
    \label{tab:app_gen_time}
    \scriptsize
    \setlength{\tabcolsep}{2.2pt}
    \renewcommand{\arraystretch}{0.94}
    \resizebox{\linewidth}{!}{\begin{tabular}{@{}lccccccc@{}}
        \toprule
        & & \multicolumn{3}{c}{KITTI-360} & \multicolumn{3}{c}{nuScenes} \\
        \cmidrule(lr){3-5}\cmidrule(l){6-8}
        Method & NFE & s $\downarrow$ & TFLOPs $\downarrow$ & TFLOP/s $\uparrow$ & s $\downarrow$ & TFLOPs $\downarrow$ & TFLOP/s $\uparrow$ \\
        \midrule
        \multicolumn{8}{@{}l}{\textbf{Generators of single scans}} \\
        LiDARGen~\cite{zyrianov2022lidargen} & 1,160 & 90.67 & 1,470 & 16.2 & -- & -- & -- \\
        R2DM~\cite{nakashima2024lidar} & 256 & 3.92 & 60.1 & 15.3 & -- & -- & -- \\
        LiDM~\cite{ran2024lidm} & 50 & \cfirst\textbf{0.76} & \cfirst\textbf{8.7} & 11.4 & -- & -- & -- \\
        R2Flow~\cite{nakashima2025r2flow} & 256 & 5.41 & 40.0 & 7.4 & -- & -- & -- \\
        R3DPA~\cite{sereyjol2026r3dpa} & 256 & 1.40 & 25.2 & \csecond 18.0 & -- & -- & -- \\
        T2LDM~\cite{qu2026t2ldm} & 1,024 & 63.99 & 536 & 8.4 & 31.75 & 266 & 8.4 \\
        T2LDM++~\cite{qu2026t2ldmpp} & 1,024 & 59.60 & 496 & 8.3 & 28.96 & 246 & 8.5 \\
        \midrule
        \multicolumn{8}{@{}l}{\textbf{Sequence generators}} \\
        LidarDM~\cite{zyrianov2025lidardm} & 618 & 16.68 & 120.1 & 7.2 & -- & -- & -- \\
        OpenDWM-DiT~\cite{opendwm} & 5 & 1.69 & \csecond 22.8 & 13.5 & 1.69 & \csecond 22.8 & \csecond 13.5 \\
        UniScene~\cite{li2025uniscene} & 50 & -- & -- & -- & \csecond 1.37 & \cfirst\textbf{13.9} & 10.1 \\
        \ours & 10 & \csecond 0.94 & 39.0 & \cfirst\textbf{41.6} & \cfirst\textbf{0.86} & 36.3 & \cfirst\textbf{42.2} \\
        \bottomrule
    \end{tabular}}
    \vspace{-3mm}
\end{table}

\tabref{tab:app_gen_time} reports the time and compute per scan for the methods in \tabref{tab:uncond_generation}, grouped by whether they generate sequences.
Each method runs on one RTX 3090, and times exclude model loading.
TFLOPs count matrix multiplications, convolutions, and attention per generated scan, and throughput divides compute by unrounded time.
OpenDWM-DiT uses the same voxel grid and networks on both datasets, so its cost is identical.
With 10 network evaluations, our model generates a scan in under 1\,s on both datasets, of which decoding 65,536 rays takes 115\,ms.
Our model is the fastest on nuScenes and second only to the single scan generator LiDM on KITTI-360.
Although it computes more per scan than several baselines, it runs at about 42\,TFLOP/s, more than twice the throughput of any other method.

\section{Ray Decoder Analysis}
\label{sec:app_rays}
This section first examines the hit probability and depth that our decoder predicts.
It then tests two properties on which decoding for heterogeneous \acp{LiDAR} relies.
Each ray's prediction should be independent of the other queried rays, and the decoder should predict returns along directions absent from its input scan.
Finally, we compare the training data and scan pattern augmentation of the autoencoder.

\subsection{Hit Probability and Depth}
\label{sec:app_rays_pred}

\begin{table}[t]
    \centering
    \caption{Hit threshold and depth refinement of our decoder on the reconstruction sets of \tabref{tab:vae_fidelity}.}
    \label{tab:app_raydec}
    \scriptsize
    \setlength{\tabcolsep}{2.2pt}
    \renewcommand{\arraystretch}{0.94}
    \resizebox{\linewidth}{!}{\begin{tabular}{@{}lcccccccc@{}}
        \toprule
        & \multicolumn{4}{c}{KITTI-360} & \multicolumn{4}{c}{nuScenes} \\
        \cmidrule(lr){2-5}\cmidrule(l){6-9}
        Setting & CD $\downarrow$ & $F_{0.1}$ $\uparrow$ & dMAE $\downarrow$ & H-F1 $\uparrow$ & CD $\downarrow$ & $F_{0.1}$ $\uparrow$ & dMAE $\downarrow$ & H-F1 $\uparrow$ \\
        \midrule
        $\tau_{\mathrm{hit}}=0.3$ & \csecond 0.060 & \csecond 0.803 & 0.309 & \csecond 0.979 & 0.065 & 0.755 & 0.306 & 0.955 \\
        $\tau_{\mathrm{hit}}=0.4$ & \cfirst\textbf{0.059} & \cfirst\textbf{0.804} & 0.307 & \cfirst\textbf{0.980} & \csecond 0.064 & 0.759 & 0.303 & \cfirst\textbf{0.959} \\
        $\tau_{\mathrm{hit}}=0.5$ (\ours) & \cfirst\textbf{0.059} & \csecond 0.803 & 0.303 & 0.974 & \csecond 0.064 & \csecond 0.762 & 0.294 & \csecond 0.958 \\
        $\tau_{\mathrm{hit}}=0.6$ & \csecond 0.060 & 0.799 & \csecond 0.288 & 0.955 & \cfirst\textbf{0.063} & \cfirst\textbf{0.764} & \csecond 0.241 & 0.938 \\
        $\tau_{\mathrm{hit}}=0.7$ & 0.061 & 0.788 & \cfirst\textbf{0.252} & 0.917 & \cfirst\textbf{0.063} & 0.755 & \cfirst\textbf{0.122} & 0.861 \\
        \midrule
        \textit{Measured hit mask} & 0.059 & 0.807 & 0.310 & -- & 0.061 & 0.778 & 0.307 & -- \\
        Preliminary depth $d^0_i$ & 0.099 & 0.561 & 0.386 & -- & 0.117 & 0.440 & 0.401 & -- \\
        \bottomrule
    \end{tabular}}
    \vspace{-3mm}
\end{table}
\begin{figure}[t]
    \centering
    \includegraphics[width=\linewidth]{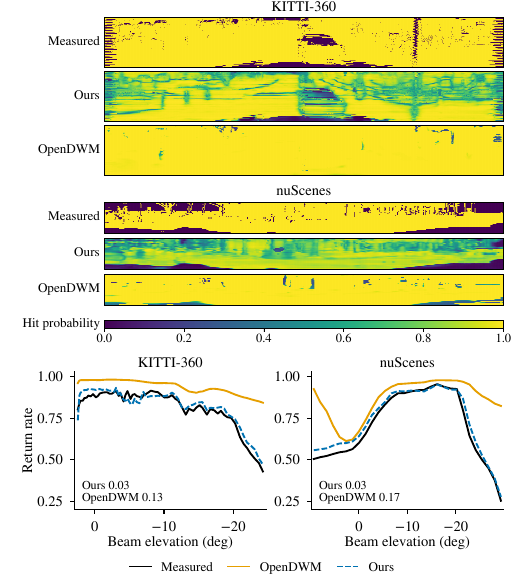}
    \caption{\textbf{Hit probability.}
    The upper panels show the front $180^\circ$ of one scan per dataset with one row per beam.
    The lower panels give the return rate of each beam over all evaluated scans, thresholding our hit probability and the opacity of the OpenDWM render head at 0.5.
    Values give the mean absolute difference from the measured return rate over beams.}
    \label{fig:app_hit}
    \vspace{-3mm}
\end{figure}

\noindent\textbf{Hit probability estimation.}
Our decoder decides whether each ray returns by thresholding its hit probability.
We test how this threshold affects the reconstruction and whether the probabilities match where measured scans return.
\tabref{tab:app_raydec} varies $\tau_{\mathrm{hit}}$, the probability threshold above which our decoder returns a ray.
Across thresholds from 0.3 to 0.7, CD changes only marginally on both datasets.
H-F1 peaks at $\tau_{\mathrm{hit}}=0.4$. We keep the default $\tau_{\mathrm{hit}}=0.5$, whose H-F1 is within 0.01 of the peak and whose dMAE is lower.
Higher thresholds retain fewer returns.
They lower dMAE on the remaining co-returned rays but reduce hit recall.
We also replace our hit decisions with the measured hit mask, which keeps our depths but returns exactly the rays that return in the measured scan.
This replacement leaves CD nearly unchanged, while its dMAE rises slightly because it also scores difficult rays that our decoder leaves without a return.

\begin{figure}[t]
    \centering
    \includegraphics[width=\linewidth]{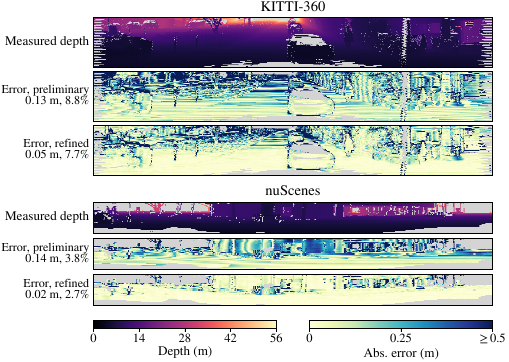}
    \caption{\textbf{Preliminary and refined depth} for the scans of \figref{fig:app_hit}, shown as the measured depth and the absolute error of both depths where our decoder and the measured scan return.
    Gray rays have no return.
    Values give the median absolute error and the fraction of rays with errors above 1\,m.}
    \label{fig:app_depth}
    \vspace{-2mm}
\end{figure}
\begin{table}[t]
    \centering
    \caption{Depth errors of our decoder on the co-returned rays of \tabref{tab:vae_fidelity} by measured range, away from and at depth discontinuities. Disc.\ is the fraction of rays at discontinuities, where a neighboring ray of the same beam differs in depth by more than 1\,m and 10\%.}
    \label{tab:app_depth_breakdown}
    \scriptsize
    \setlength{\tabcolsep}{3pt}
    \renewcommand{\arraystretch}{0.94}
    \resizebox{0.86\linewidth}{!}{\begin{tabular}{@{}lcccccc@{}}
        \toprule
        & & & \multicolumn{2}{c}{dMAE} & \multicolumn{2}{c}{MedAE} \\
        \cmidrule(lr){4-5}\cmidrule(l){6-7}
        Range & Rays (\%) & Disc.\ (\%) & Away & At & Away & At \\
        \midrule
        \multicolumn{7}{@{}l}{\textit{KITTI-360 (2,000 scans)}} \\
        3 to 10\,m & 68.6 & 3.7 & 0.096 & 1.397 & 0.029 & 0.184 \\
        10 to 20\,m & 22.8 & 9.8 & 0.239 & 2.066 & 0.074 & 0.354 \\
        20 to 40\,m & 7.4 & 16.1 & 0.595 & 3.613 & 0.190 & 0.692 \\
        Beyond 40\,m & 1.2 & 23.5 & 1.290 & 5.800 & 0.407 & 1.175 \\
        All & 100 & 6.3 & 0.172 & 2.251 & 0.039 & 0.361 \\
        \midrule
        \multicolumn{7}{@{}l}{\textit{nuScenes (6,019 scans)}} \\
        3 to 10\,m & 61.8 & 2.1 & 0.070 & 2.326 & 0.022 & 0.206 \\
        10 to 20\,m & 23.7 & 7.5 & 0.214 & 2.596 & 0.045 & 0.520 \\
        20 to 40\,m & 13.3 & 14.5 & 0.448 & 3.134 & 0.141 & 0.707 \\
        Beyond 40\,m & 1.3 & 18.3 & 0.739 & 4.274 & 0.262 & 0.895 \\
        All & 100 & 5.2 & 0.156 & 2.804 & 0.031 & 0.572 \\
        \bottomrule
    \end{tabular}}
    \vspace{-3mm}
\end{table}%
In \figref{fig:app_hit}, our probabilities follow measured returns, including beams that rarely return and the region occluded by the ego vehicle.
The return rate of a beam is the fraction of its fired rays that return.
Averaged over beams, our return rates lie much closer to the measured ones than those of the OpenDWM render head.
The render head instead returns almost every ray, as its opacity is close to one on most rays.

\noindent\textbf{Depth estimation.}
Our decoder first selects a depth bin and offset with global attention and then corrects this preliminary depth with local features.
To measure what the local correction adds, \tabref{tab:app_raydec} and \figref{fig:app_depth} compare the preliminary depth $d^0_i$ and the refined depth $\hat{d}_i$ from the same decoding pass.
On both datasets, refinement lowers dMAE in \tabref{tab:app_raydec} and more than halves the median error over all scans, as in the scans of \figref{fig:app_depth}.

\begin{table}[t]
    \centering
    \caption{Returns on masked rays of the 362 KITTI-360 scans of \tabref{tab:occ_oracle}, under its protocol. Only OpenDWM and ours omit these rays from the input.}
    \label{tab:app_masked_repr}
    \scriptsize
    \setlength{\tabcolsep}{2.4pt}
    \renewcommand{\arraystretch}{0.92}
    \resizebox{\linewidth}{!}{\begin{tabular}{@{}llcccc@{}}
        \toprule
        Scene (resolution) & Method & CD $\downarrow$ & $F_{0.1}$ $\uparrow$ & dMAE $\downarrow$ & AbsRel $\downarrow$ \\
        \midrule
        \multicolumn{6}{@{}l}{\textbf{Every other row masked}} \\
        Occ., SSCBench 0.2\,m & Ray marching & 0.169 & 0.115 & 1.43 & 11.83 \\
        Mesh, NKSR 0.15\,m & Raycasting & \cfirst\textbf{0.079} & \cfirst\textbf{0.701} & 0.51 & 4.12 \\
        Mesh, NKSR 0.15\,m & LidarDM~\cite{zyrianov2025lidardm} & 0.124 & 0.399 & 0.74 & 5.38 \\
        Latent, OpenDWM~\cite{opendwm} & Render head & 0.098 & 0.572 & \csecond 0.45 & \csecond 3.58 \\
        Latent, \ours & Ray decoder & \csecond 0.080 & \csecond 0.681 & \cfirst\textbf{0.35} & \cfirst\textbf{2.72} \\
        \midrule
        \multicolumn{6}{@{}l}{\textbf{One of every four rows masked}} \\
        Occ., SSCBench 0.2\,m & Ray marching & 0.180 & 0.083 & 1.43 & 11.82 \\
        Mesh, NKSR 0.15\,m & Raycasting & \csecond 0.075 & \csecond 0.705 & 0.50 & 4.01 \\
        Mesh, NKSR 0.15\,m & LidarDM~\cite{zyrianov2025lidardm} & 0.133 & 0.353 & 0.70 & 5.13 \\
        Latent, OpenDWM~\cite{opendwm} & Render head & 0.098 & 0.578 & \csecond 0.42 & \csecond 3.44 \\
        Latent, \ours & Ray decoder & \cfirst\textbf{0.074} & \cfirst\textbf{0.710} & \cfirst\textbf{0.31} & \cfirst\textbf{2.46} \\
        \midrule
        \multicolumn{6}{@{}l}{\textbf{Rosette pattern kept}} \\
        Occ., SSCBench 0.2\,m & Ray marching & 0.165 & 0.136 & 1.56 & 12.13 \\
        Mesh, NKSR 0.15\,m & Raycasting & \csecond 0.080 & \csecond 0.694 & 0.58 & 4.48 \\
        Mesh, NKSR 0.15\,m & LidarDM~\cite{zyrianov2025lidardm} & 0.113 & 0.489 & 0.85 & 5.76 \\
        Latent, OpenDWM~\cite{opendwm} & Render head & 0.090 & 0.641 & \csecond 0.47 & \csecond 3.63 \\
        Latent, \ours & Ray decoder & \cfirst\textbf{0.075} & \cfirst\textbf{0.709} & \cfirst\textbf{0.36} & \cfirst\textbf{2.83} \\
        \bottomrule
    \end{tabular}}
    \vspace{-4mm}
\end{table}%
However, errors above 1\,m remain on 3 to 8\% of the co-returned rays in \figref{fig:app_depth} and account for most of dMAE.
Many of them occur where a neighboring ray differs sharply in depth, as a slightly misplaced boundary gives a ray the depth of the wrong surface (Appendix~\ref{sec:app_rays_err}).
Nevertheless, the entropy of the bin distribution flags these unreliable depths.
Removing the 10\% of rays with the highest entropy lowers dMAE on the remaining rays by about half.
Therefore, our decoder places most depths close to the measured surface, and the entropy of its bin distribution is high for the few large errors at depth discontinuities.

\begin{figure}[t]
    \centering
    \includegraphics[width=\linewidth]{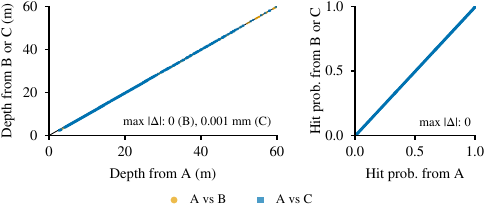}
    \caption{\textbf{Consistency across query sets.}
    One KITTI-360 latent is decoded along the fixed beam grid (A), a random half of its rays (B), and A with the VLP-16 and OS2-128 rays added (C).
    The plots compare B and C with A on their shared rays.}
    \label{fig:app_consistency}
    \vspace{-3mm}
    \vspace{4mm}
    \centering
    \includegraphics[width=\linewidth]{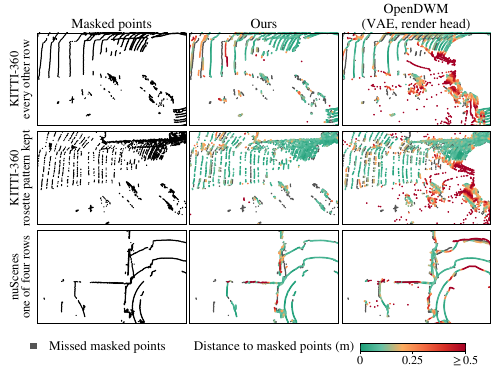}
    \caption{\textbf{Masked rays} in \ac{BEV} crops of $16\times10$\,m.
    Each row applies one mask of \tabref{tab:beam_withholding}.
    Returned points are colored by their distance to the nearest masked point, and a masked point is missed if no returned point lies within 0.1\,m.}
    \label{fig:app_withheld}
    \vspace{-3mm}
\end{figure}%
\subsection{Depth Error by Range and Discontinuity}
\label{sec:app_rays_err}
Appendix~\ref{sec:app_rays_pred} shows that a few large errors dominate dMAE.
\tabref{tab:app_depth_breakdown} shows where they occur.
Away from depth discontinuities, our depths are accurate near the sensor, with a median error of 2 to 3\,cm within 10\,m, and degrade gradually with range.
Rays at discontinuities account for only 5 to 6\% of the rays but carry about half of the error.

\subsection{Consistency across Query Sets}
\label{sec:app_rays_consist}
Our decoder predicts each ray from the latent without attending to the other queried rays.
To verify this, we test 50 KITTI-360 latents in five configurations, three of which \figref{fig:app_consistency} shows for one latent.
We decode the 65,536 rays of the fixed beam grid, a random half, and all rays together with the VLP-16 and OS2-128 scan patterns.
We also decode them in random order and in smaller query chunks.
Across all comparisons, shared rays keep bit-identical hit probabilities, and no hit decision changes.
Depths are bit-identical for the half and random order.
Added scan patterns and smaller chunks change returned depths only by floating-point rounding, which varies with the batch of decoded rays.
Therefore, each scan pattern receives practically the same returns whether it is decoded alone or together with other patterns.
Rendering from an intermediate surface or field shares this property, whereas range image generators draw a separate sample for each scan pattern, so their scans of different patterns do not show one scene.

\subsection{Unseen Ray Directions}
\label{sec:app_rays_unseen}
Our decoder can query ray directions that the input scan does not contain.
Decoding a new scan pattern relies on this ability.
We test it by masking parts of the input and by querying directions off the fixed beam grid.

\newcommand{\beamWithholdingRows}{%
        \multicolumn{7}{@{}l}{\textbf{KITTI-360 0018, every other row masked}} \\
        LiDM (VQ)~\cite{ran2024lidm} & 0.196 & 0.017 & 3.287 & 1.936 & 5.118 & 25.95 \\
        OpenDWM (VQ)~\cite{opendwm} & 0.144 & 0.234 & 1.073 & 0.281 & 2.869 & 10.47 \\
        OpenDWM (VAE, voxel head)~\cite{opendwm} & 0.145 & 0.230 & 1.008 & 0.281 & 2.683 & 9.85 \\
        OpenDWM (VAE, render head)~\cite{opendwm} & \csecond 0.098 & \csecond 0.579 & \csecond 0.538 & \csecond 0.097 & \csecond 2.063 & \csecond 4.58 \\
        R3DPA (VAE)~\cite{sereyjol2026r3dpa} & 0.156 & 0.201 & 0.885 & 0.361 & 2.271 & 7.93 \\
        GEM (VQ)~\cite{wu2026gem} & 0.142 & 0.274 & 0.729 & 0.229 & 2.162 & 6.24 \\
        \ours & \cfirst\textbf{0.081} & \cfirst\textbf{0.679} & \cfirst\textbf{0.411} & \cfirst\textbf{0.063} & \cfirst\textbf{1.674} & \cfirst\textbf{3.47} \\
        \ours, \textit{full input} & 0.063 & 0.773 & 0.302 & 0.043 & 1.471 & 2.63 \\
        \midrule
        \multicolumn{7}{@{}l}{\textbf{KITTI-360 0018, one of every four rows masked}} \\
        LiDM (VQ)~\cite{ran2024lidm} & 0.179 & 0.075 & 1.052 & 0.454 & 2.387 & 9.27 \\
        OpenDWM (VQ)~\cite{opendwm} & 0.157 & 0.195 & 1.201 & 0.298 & 3.162 & 11.38 \\
        OpenDWM (VAE, voxel head)~\cite{opendwm} & 0.158 & 0.185 & 1.086 & 0.302 & 2.875 & 10.42 \\
        OpenDWM (VAE, render head)~\cite{opendwm} & \csecond 0.100 & \csecond 0.563 & \csecond 0.525 & \csecond 0.093 & 2.015 & \csecond 4.50 \\
        R3DPA (VAE)~\cite{sereyjol2026r3dpa} & 0.129 & 0.370 & 0.590 & 0.149 & \csecond 1.970 & 4.70 \\
        GEM (VQ)~\cite{wu2026gem} & 0.138 & 0.284 & 0.593 & 0.129 & 1.983 & 4.87 \\
        \ours & \cfirst\textbf{0.076} & \cfirst\textbf{0.700} & \cfirst\textbf{0.386} & \cfirst\textbf{0.054} & \cfirst\textbf{1.675} & \cfirst\textbf{3.18} \\
        \ours, \textit{full input} & 0.065 & 0.759 & 0.306 & 0.042 & 1.532 & 2.63 \\
        \midrule
        \multicolumn{7}{@{}l}{\textbf{KITTI-360 0018, rosette pattern kept}} \\
        LiDM (VQ)~\cite{ran2024lidm} & 0.172 & 0.112 & 1.587 & 0.636 & 3.282 & 12.33 \\
        OpenDWM (VQ)~\cite{opendwm} & 0.137 & 0.276 & 1.060 & 0.266 & 2.872 & 10.38 \\
        OpenDWM (VAE, voxel head)~\cite{opendwm} & 0.135 & 0.279 & 0.943 & 0.252 & 2.622 & 9.35 \\
        OpenDWM (VAE, render head)~\cite{opendwm} & \csecond 0.092 & \csecond 0.629 & \csecond 0.570 & \csecond 0.090 & \csecond 2.326 & \csecond 4.73 \\
        R3DPA (VAE)~\cite{sereyjol2026r3dpa} & 0.122 & 0.431 & 0.825 & 0.181 & 2.430 & 6.20 \\
        GEM (VQ)~\cite{wu2026gem} & 0.118 & 0.452 & 0.735 & 0.166 & 2.597 & 5.57 \\
        \ours & \cfirst\textbf{0.077} & \cfirst\textbf{0.699} & \cfirst\textbf{0.420} & \cfirst\textbf{0.058} & \cfirst\textbf{1.802} & \cfirst\textbf{3.59} \\
        \ours, \textit{full input} & 0.067 & 0.757 & 0.345 & 0.048 & 1.624 & 2.89 \\
        \midrule
        \multicolumn{7}{@{}l}{\textbf{KITTI Odometry 08, every other row masked}} \\
        LiDM (VQ)~\cite{ran2024lidm} & 0.194 & 0.023 & 3.683 & 2.138 & 6.058 & 24.95 \\
        OpenDWM (VQ)~\cite{opendwm} & 0.151 & 0.202 & 1.466 & 0.343 & 3.811 & 13.95 \\
        OpenDWM (VAE, voxel head)~\cite{opendwm} & 0.151 & 0.203 & 1.336 & 0.337 & 3.487 & 12.98 \\
        OpenDWM (VAE, render head)~\cite{opendwm} & \csecond 0.102 & \csecond 0.547 & \csecond 0.723 & \csecond 0.106 & \csecond 2.679 & \csecond 5.26 \\
        R3DPA (VAE)~\cite{sereyjol2026r3dpa} & 0.162 & 0.169 & 1.212 & 0.472 & 3.013 & 9.14 \\
        GEM (VQ)~\cite{wu2026gem} & 0.147 & 0.245 & 1.049 & 0.274 & 3.022 & 7.57 \\
        \ours & \cfirst\textbf{0.100} & \cfirst\textbf{0.561} & \cfirst\textbf{0.617} & \cfirst\textbf{0.095} & \cfirst\textbf{2.307} & \cfirst\textbf{4.73} \\
        \ours, \textit{full input} & 0.087 & 0.636 & 0.491 & 0.074 & 2.102 & 3.91 \\
        \midrule
        \multicolumn{7}{@{}l}{\textbf{KITTI Odometry 08, one of every four rows masked}} \\
        LiDM (VQ)~\cite{ran2024lidm} & 0.181 & 0.054 & 1.102 & 0.439 & 2.699 & 9.42 \\
        OpenDWM (VQ)~\cite{opendwm} & 0.165 & 0.152 & 1.626 & 0.368 & 4.168 & 15.07 \\
        OpenDWM (VAE, voxel head)~\cite{opendwm} & 0.165 & 0.148 & 1.410 & 0.360 & 3.651 & 13.47 \\
        OpenDWM (VAE, render head)~\cite{opendwm} & \csecond 0.104 & \csecond 0.543 & \csecond 0.683 & \csecond 0.100 & 2.601 & \csecond 5.33 \\
        R3DPA (VAE)~\cite{sereyjol2026r3dpa} & 0.163 & 0.163 & 0.707 & 0.266 & \csecond 2.290 & 6.52 \\
        GEM (VQ)~\cite{wu2026gem} & 0.156 & 0.148 & 0.798 & 0.171 & 2.953 & 6.60 \\
        \ours & \cfirst\textbf{0.100} & \cfirst\textbf{0.559} & \cfirst\textbf{0.590} & \cfirst\textbf{0.090} & \cfirst\textbf{2.249} & \cfirst\textbf{4.60} \\
        \ours, \textit{full input} & 0.090 & 0.617 & 0.482 & 0.073 & 2.069 & 3.91 \\
        \midrule
        \multicolumn{7}{@{}l}{\textbf{KITTI Odometry 08, rosette pattern kept}} \\
        LiDM (VQ)~\cite{ran2024lidm} & 0.196 & 0.012 & 3.718 & 1.958 & 6.104 & 26.85 \\
        OpenDWM (VQ)~\cite{opendwm} & 0.150 & 0.210 & 1.561 & 0.329 & 4.056 & 14.61 \\
        OpenDWM (VAE, voxel head)~\cite{opendwm} & 0.147 & 0.213 & 1.326 & 0.293 & \csecond 3.606 & 12.87 \\
        OpenDWM (VAE, render head)~\cite{opendwm} & \csecond 0.103 & \csecond 0.553 & \csecond 0.990 & \csecond 0.101 & 3.908 & \csecond 7.25 \\
        R3DPA (VAE)~\cite{sereyjol2026r3dpa} & 0.175 & 0.105 & 2.117 & 0.830 & 4.544 & 14.81 \\
        GEM (VQ)~\cite{wu2026gem} & 0.179 & 0.080 & 4.285 & 1.866 & 7.802 & 28.78 \\
        \ours & \cfirst\textbf{0.098} & \cfirst\textbf{0.573} & \cfirst\textbf{0.644} & \cfirst\textbf{0.086} & \cfirst\textbf{2.493} & \cfirst\textbf{5.07} \\
        \ours, \textit{full input} & 0.090 & 0.622 & 0.588 & 0.083 & 2.364 & 4.44 \\
        \midrule
        \multicolumn{7}{@{}l}{\textbf{nuScenes, every other row masked}} \\
        OpenDWM (VQ)~\cite{opendwm} & 0.153 & 0.226 & 0.824 & 0.216 & 2.496 & 7.74 \\
        OpenDWM (VAE, voxel head)~\cite{opendwm} & 0.163 & 0.158 & \csecond 0.779 & 0.260 & \csecond 2.349 & 7.95 \\
        OpenDWM (VAE, render head)~\cite{opendwm} & \csecond 0.115 & \csecond 0.480 & \csecond 0.779 & \cfirst\textbf{0.074} & 2.689 & \csecond 5.70 \\
        \ours & \cfirst\textbf{0.112} & \cfirst\textbf{0.481} & \cfirst\textbf{0.478} & \csecond 0.079 & \cfirst\textbf{1.914} & \cfirst\textbf{3.96} \\
        \ours, \textit{full input} & 0.065 & 0.755 & 0.301 & 0.034 & 1.543 & 2.32 \\
        \midrule
        \multicolumn{7}{@{}l}{\textbf{nuScenes, one of every four rows masked}} \\
        OpenDWM (VQ)~\cite{opendwm} & 0.151 & 0.241 & 0.899 & 0.221 & 2.683 & 8.01 \\
        OpenDWM (VAE, voxel head)~\cite{opendwm} & 0.164 & 0.156 & 0.802 & 0.261 & \csecond 2.415 & 8.11 \\
        OpenDWM (VAE, render head)~\cite{opendwm} & \csecond 0.116 & \csecond 0.476 & \csecond 0.722 & \csecond 0.070 & 2.515 & \csecond 5.37 \\
        \ours & \cfirst\textbf{0.106} & \cfirst\textbf{0.497} & \cfirst\textbf{0.348} & \cfirst\textbf{0.063} & \cfirst\textbf{1.470} & \cfirst\textbf{3.33} \\
        \ours, \textit{full input} & 0.064 & 0.758 & 0.298 & 0.033 & 1.534 & 2.32 \\
}
\begin{table}[t]
    \centering
    \caption{Returns on rays masked from the input of the autoencoders of \ac{LiDAR} generators, under the protocol of \tabref{tab:vae_fidelity}. Only the masked rows or cells are scored.}
    \label{tab:beam_withholding}
    \scriptsize
    \setlength{\tabcolsep}{1.6pt}
    \renewcommand{\arraystretch}{0.94}
    \resizebox{\linewidth}{!}{\begin{tabular}{@{}lcccccc@{}}
        \toprule
        Method & CD $\downarrow$ & $F_{0.1}$ $\uparrow$ & dMAE $\downarrow$ & MedAE $\downarrow$ & RMSE $\downarrow$ & AbsRel $\downarrow$ \\
        \midrule
        \beamWithholdingRows
        \bottomrule
    \end{tabular}}
    \par\vspace{2pt}\parbox{\linewidth}{\scriptsize With full input, our autoencoder encodes the complete scan and is scored on the same masked rows or cells.}
    \vspace{-6mm}
\end{table}%
\noindent\textbf{Against scene representations.}
\tabref{tab:app_masked_repr} extends the masked setting of \tabref{tab:occ_oracle} with three more masks on the same KITTI-360 scans.
Row masks cover every other row or one of every four rows of the range images in Appendix~\ref{sec:app_metrics}.
They remove the input points whose nearest row is masked.
A rosette pattern like that of the Avia, drawn for each scan, keeps some cells and masks the rest, and input points in masked cells are removed.
OpenDWM and our model encode the masked scan, whereas occupancy and meshes keep their accumulated representations built without masks.
Even so, our decoder has the lowest depth errors in every masked setting.
Its CD and $F_{0.1}$ are also the best, except with every other row masked, where the NKSR mesh has a slightly lower CD and a higher $F_{0.1}$.

\noindent\textbf{Against other autoencoders.}
\tabref{tab:beam_withholding} applies masks to the inputs of the autoencoders of \tabref{tab:vae_fidelity}.
We use 2,000 KITTI-360 sequence 0018 scans, 2,000 scans sampled uniformly by pose from KITTI Odometry~\cite{geiger2012kitti} sequence 08, and the 6,019 nuScenes validation scans.
In the case of KITTI Odometry, its sequence 08 lies outside the training data of every compared autoencoder.
Its scans use the KITTI-360 crop, and their range images have one row per beam at the beam's median elevation with 1024 azimuth bins.
Row masks apply to all three datasets.
The rosette applies only to KITTI-360 and KITTI Odometry, retaining about 50\% and 60\% of cells on average, respectively.
Our autoencoder is the most accurate on every metric, except that the OpenDWM render head has a lower MedAE with every other nuScenes row masked.
\figref{fig:app_withheld} shows the masked rays of one KITTI-360 scan under two masks and of one nuScenes scan in \ac{BEV}.
Most of our returns on these rays lie within 0.25\,m of a measured return, whereas the OpenDWM render head places many returns more than 0.5\,m away.
The render head's returns far from the measured ones lie mainly on distant structures.

\begin{table}[t]
    \centering
    \caption{Returns along the exact directions of the measured KITTI-360 returns, off the fixed beam grid, on the 362 scans of \tabref{tab:occ_oracle} with full input. CD and $F_{0.1}$ compare point sets. Every query has a measured return, so false returns are not scored.}
    \label{tab:app_offgrid}
    \scriptsize
    \setlength{\tabcolsep}{2.4pt}
    \renewcommand{\arraystretch}{0.94}
    \resizebox{\linewidth}{!}{\begin{tabular}{@{}llcccccc@{}}
        \toprule
        Scene (resolution) & Method & CD $\downarrow$ & $F_{0.1}$ $\uparrow$ & dMAE $\downarrow$ & MedAE $\downarrow$ & RMSE $\downarrow$ & AbsRel $\downarrow$ \\
        \midrule
        Occ., SSCBench 0.2\,m & Ray marching & 0.158 & 0.171 & 1.552 & 0.570 & 3.318 & 12.44 \\
        Mesh, NKSR 0.15\,m & Raycasting & \csecond 0.065 & \csecond 0.780 & 0.614 & \csecond 0.056 & 2.271 & 4.73 \\
        Mesh, NKSR 0.15\,m & LidarDM~\cite{zyrianov2025lidardm} & 0.091 & 0.635 & 0.803 & 0.118 & 2.511 & 5.48 \\
        Latent, OpenDWM & Render head & 0.074 & 0.745 & \csecond 0.416 & 0.086 & \csecond 1.742 & \csecond 3.50 \\
        Latent, \ours & Ray decoder & \cfirst\textbf{0.053} & \cfirst\textbf{0.833} & \cfirst\textbf{0.332} & \cfirst\textbf{0.038} & \cfirst\textbf{1.578} & \cfirst\textbf{2.47} \\
        \bottomrule
    \end{tabular}}
    \vspace{-3mm}
\end{table}%
\noindent\textbf{Directions off the fixed beam grid.}
Our decoder is trained along the fixed beam grid of Appendix~\ref{sec:app_train} for single scans, whereas the reconstruction tables query it along the rays that the sensor fired.
On KITTI-360, these rays lie up to 0.57$^\circ$ from the nearest ray of this grid.
\tabref{tab:app_offgrid} scores every method along the exact directions of the measured returns, and our decoder is the most accurate on every metric.
As the angle to the nearest grid ray grows, our median depth error rises from 0.020 to 0.056\,m, but it remains below those of occupancy, the NKSR mesh constructed from the static scene, and the OpenDWM render head at every tested angle.
Therefore, our decoder keeps the lowest depth errors even off the fixed beam grid used to train it on single scans.

\afterpage{\begin{table}[t]
    \centering
    \caption{Training data and scan pattern augmentation of the autoencoder. S uses single scans only, S+A adds aggregated scans, and Ours adds the scan pattern augmentation. Masked inputs are scored only on their masked rays, as in \tabref{tab:beam_withholding}, with the Avia mask of \tabref{tab:occ_oracle} applied to the sequence 0018 scans.}
    \label{tab:app_aug}
    \scriptsize
    \setlength{\tabcolsep}{2pt}
    \renewcommand{\arraystretch}{0.94}
    \resizebox{\linewidth}{!}{\begin{tabular}{@{}lccccccccc@{}}
        \toprule
        & \multicolumn{3}{c}{CD $\downarrow$} & \multicolumn{3}{c}{$F_{0.1}$ $\uparrow$} & \multicolumn{3}{c}{H-F1 $\uparrow$} \\
        \cmidrule(lr){2-4}\cmidrule(lr){5-7}\cmidrule(l){8-10}
        Input & S & S+A & Ours & S & S+A & Ours & S & S+A & Ours \\
        \midrule
        \multicolumn{10}{@{}l}{\textit{KITTI-360 (sequence 0018)}} \\
        Complete scan & \cfirst\textbf{0.059} & \csecond 0.061 & \cfirst\textbf{0.059} & \cfirst\textbf{0.808} & 0.798 & \csecond 0.803 & \cfirst\textbf{0.977} & \csecond 0.975 & 0.974 \\
        Every other row masked & \csecond 0.082 & 0.083 & \cfirst\textbf{0.081} & \csecond 0.674 & 0.670 & \cfirst\textbf{0.679} & \csecond 0.937 & \cfirst\textbf{0.947} & \cfirst\textbf{0.947} \\
        One of every four rows masked & \csecond 0.077 & 0.079 & \cfirst\textbf{0.076} & \csecond 0.698 & 0.690 & \cfirst\textbf{0.700} & \cfirst\textbf{0.968} & \cfirst\textbf{0.968} & \cfirst\textbf{0.968} \\
        Rosette pattern kept & 0.079 & \csecond 0.078 & \cfirst\textbf{0.077} & 0.685 & \csecond 0.692 & \cfirst\textbf{0.699} & 0.879 & \csecond 0.912 & \cfirst\textbf{0.923} \\
        Avia cells masked & \csecond 0.087 & \csecond 0.087 & \cfirst\textbf{0.082} & \csecond 0.621 & 0.617 & \cfirst\textbf{0.652} & \csecond 0.867 & 0.834 & \cfirst\textbf{0.889} \\
        \midrule
        \multicolumn{10}{@{}l}{\textit{nuScenes}} \\
        Complete scan & \cfirst\textbf{0.060} & \csecond 0.063 & 0.064 & \cfirst\textbf{0.775} & \csecond 0.767 & 0.762 & \csecond 0.962 & \cfirst\textbf{0.964} & 0.958 \\
        Every other row masked & \csecond 0.119 & 0.121 & \cfirst\textbf{0.112} & 0.411 & \csecond 0.414 & \cfirst\textbf{0.481} & 0.651 & \csecond 0.706 & \cfirst\textbf{0.790} \\
        One of every four rows masked & \cfirst\textbf{0.104} & 0.109 & \csecond 0.106 & \csecond 0.494 & 0.458 & \cfirst\textbf{0.497} & \csecond 0.656 & 0.633 & \cfirst\textbf{0.710} \\
        \bottomrule
    \end{tabular}}
    \vspace{-3mm}
\end{table}}%
\subsection{Training Data and Augmentation}
\label{sec:app_rays_aug}
Our autoencoder learns from aggregated scans and from the scan pattern augmentation of Appendix~\ref{sec:app_train}.
To test which of the two helps on sparse and irregular inputs, \tabref{tab:app_aug} compares three autoencoders.
S trains on single scans only, and S+A adds aggregated scans.
Ours further adds the scan pattern augmentation.

\noindent\textbf{Effect of the augmentation.}
On complete single scans, the three autoencoders stay close, and S has the highest $F_{0.1}$.
With either row mask on KITTI-360, the three also remain close, as the remaining input is still dense.
The augmentation helps most when the input becomes sparse or irregular.
On nuScenes with every other row masked, it clearly raises $F_{0.1}$ and H-F1 over S+A.
With the rosette kept and the Avia cells masked, our autoencoder is best on all three metrics.

\noindent\textbf{Without the augmentation.}
In contrast, adding aggregated scans alone helps only in some settings.
S+A is better than S on the rosette but worse on all three metrics with one of every four nuScenes rows masked.
It also has lower $F_{0.1}$ and H-F1 than S on the Avia cells.
In these two settings, S+A returns fewer of the masked rays than S at a similar precision.
Both settings keep most of the input, whereas the rosette and the mask of every other row keep only about half.
To test whether the amount of remaining input explains the deficit, we mask one of every four nuScenes rows in the front half of the scan and every other row in the rear half.
The sparser rear half raises the returns of S+A and Ours in the front half but changes those of S only marginally.
This suggests that the deficit of S+A depends on the density of the whole scan rather than on the masked rows themselves.
We attribute this to the miss labels of aggregated scans, which often lie next to densely fused points.
From these labels, S+A learns to treat an empty direction in a densely observed scan as empty space.
It returns nothing on many masked rays, mostly on the ground within 20\,m.
Our augmentation shows half of the aggregated scans through scan patterns of varied density and supervises only the labeled cells along these patterns, so fewer of its miss labels lie next to dense input.
In these regions, our autoencoder returns more masked rays than S+A and about as many as S or more.
Therefore, aggregated scans alone do not reliably help on masked inputs, as their effect depends on the density of the rest of the scan.

\afterpage{\begin{table}[t]
    \centering
    \caption{Prediction against warping the last observed scan on KITTI-360 over the five predicted frames. Dyn.\ is the recall at 0.1\,m of the measured points in boxes that moved more than 0.5\,m since the last observed frame.}
    \label{tab:app_pred_horizon}
    \scriptsize
    \setlength{\tabcolsep}{3pt}
    \renewcommand{\arraystretch}{0.94}
    \begin{tabular}{@{}lcccc@{}}
        \toprule
        Method & CD $\downarrow$ & $F_{0.1}$ $\uparrow$ & H-F1 $\uparrow$ & Dyn.\ $\uparrow$ \\
        \midrule
        \multicolumn{5}{@{}l}{\textbf{0.5\,s} horizon} \\
        Ego warp & \cfirst\textbf{0.085} & \cfirst\textbf{0.677} & \csecond 0.775 & 0.401 \\
        Ego and box warp & \csecond 0.086 & \csecond 0.674 & \csecond 0.775 & \cfirst\textbf{0.594} \\
        \ours & 0.087 & 0.649 & \cfirst\textbf{0.975} & \csecond 0.502 \\
        \midrule
        \multicolumn{5}{@{}l}{\textbf{1.0\,s} horizon} \\
        Ego warp & \csecond 0.099 & \csecond 0.593 & \csecond 0.690 & 0.344 \\
        Ego and box warp & 0.100 & 0.589 & 0.689 & \cfirst\textbf{0.471} \\
        \ours & \cfirst\textbf{0.095} & \cfirst\textbf{0.600} & \cfirst\textbf{0.973} & \csecond 0.444 \\
        \midrule
        \multicolumn{5}{@{}l}{\textbf{1.5\,s} horizon} \\
        Ego warp & \csecond 0.111 & \csecond 0.521 & \csecond 0.616 & 0.315 \\
        Ego and box warp & \csecond 0.111 & 0.516 & \csecond 0.616 & \csecond 0.382 \\
        \ours & \cfirst\textbf{0.101} & \cfirst\textbf{0.564} & \cfirst\textbf{0.971} & \cfirst\textbf{0.458} \\
        \midrule
        \multicolumn{5}{@{}l}{\textbf{3.0\,s} horizon} \\
        Ego warp & \csecond 0.136 & \csecond 0.361 & \csecond 0.462 & 0.174 \\
        Ego and box warp & 0.137 & 0.357 & 0.461 & \csecond 0.194 \\
        \ours & \cfirst\textbf{0.115} & \cfirst\textbf{0.478} & \cfirst\textbf{0.967} & \cfirst\textbf{0.383} \\
        \bottomrule
    \end{tabular}
    \vspace{-3mm}
\end{table}}
\noindent\textbf{Summary.}
Aggregated scans serve to supervise directions between the beams of the training sensors, which \tabref{tab:app_aug} does not score.
In all three autoencoders, every supervised return is itself part of the input.
Neither S nor S+A masks the input during training.
The augmentation varies the density and scan pattern of the inputs and of the supervised rays.
Therefore, the additional gains over S+A on sparse and irregular inputs come from the scan pattern augmentation, which observes the aggregated scans along random patterns.

\section{Prediction Analysis}
\label{sec:app_pred}
This section analyzes our predictions on KITTI-360 and also on KITTI Odometry in Appendix~\ref{sec:app_pred_gem}.
It first compares them with warping the last observed scan and with other LiDAR prediction models at horizons up to 3.0\,s.
It then tests how prediction depends on its inputs and breaks down the world model ablation of \tabref{tab:wm_ablation}.
Finally, it shows layout editing during prediction.

\subsection{Comparison with Warping}
\label{sec:app_pred_warp}

A simple way to predict a future scan is to move the last observed scan by the given motion.
To test when our predictions improve on such warping, \tabref{tab:app_pred_horizon} compares them with two warps on KITTI-360.
The ego warp moves the scan by the ego motion.
The ego and box warp also moves the points inside each annotated box with that box.
Each test window contains three observed and five predicted frames spaced by 0.1, 0.2, 0.3, or 0.6\,s, so the last prediction lies 0.5, 1.0, 1.5, or 3.0\,s after the last observation.
At 0.1\,s spacing, these are the 250 windows of \tabref{tab:sequence_generation}.
At the wider spacings, we use the 248 of these windows that have enough earlier frames, and each keeps the last observed frame of the original window.
The KITTI-360 model is trained with all four spacings (Appendix~\ref{sec:app_train}).
At 0.5\,s, both warps give a slightly lower CD and a higher $F_{0.1}$ than our predictions, and the ego and box warp recovers more points of moving objects.
At 1.0\,s, our predictions have lower CD and higher $F_{0.1}$, although the ego and box warp still recovers more points of moving objects.
At the 1.5 and 3.0\,s horizons, our predictions lead in every metric.
Our H-F1 is higher at every horizon and stays nearly constant, whereas that of warping falls steadily.
Warping is a strong baseline at short horizons, as the scene changes little.
However, it only moves observed points and cannot place returns on newly visible surfaces.
Therefore, in these tests, warping competes only at short horizons, and our predictions are more accurate at longer ones.

\subsection{Comparison with LiDAR Prediction Models}
\label{sec:app_pred_gem}
Following the prediction experiments of Sec.~\ref{sec:prediction}, we further compare our method with GEM~\cite{wu2026gem} and ST3DCNN~\cite{mersch2021st3dcnn}, which those experiments do not include because their released models predict from observed scans without boxes and are trained on KITTI Odometry~\cite{geiger2012kitti}.
GEM also receives the future ego translations, whereas ST3DCNN receives no motion.
Our model is trained on KITTI-360 and conditions its prediction on the ego poses and, where available, on boxes.
Therefore, we evaluate these models on both KITTI Odometry and KITTI-360 in \tabref{tab:app_pred_gem}.
As in the evaluation of GEM, each window has five observed scans and five predicted scans 0.6\,s apart, and CD is averaged over the horizons from 0.6 to 3.0\,s.
GEM uses its released 3\,s model, ST3DCNN rolls out its released 10\,Hz model from the five scans 0.1\,s apart that end at the last observed frame, and our model receives the latest three of the five observed scans.
On KITTI Odometry, which has no box labels, windows start every ten frames on sequences 08 to 10.
On KITTI-360, prediction starts at the last observed frame of each of the 250 windows of \tabref{tab:sequence_generation}, and we keep the 248 windows with five observed scans 0.6\,s apart.
We report our model with and without boxes.
Without boxes, the model receives no box layout and warps the references by ego motion alone.
All metrics use the KITTI-360 crop of Appendix~\ref{sec:app_metrics} on both datasets.

\begin{table}[t]
    \centering
    \caption{Comparison with LiDAR prediction models over five predicted frames 0.6\,s apart. Subscripts are frame gaps, and CD compares point sets with the measured future scan.}
    \label{tab:app_pred_gem}
    \scriptsize
    \setlength{\tabcolsep}{3pt}
    \renewcommand{\arraystretch}{0.94}
    \resizebox{\linewidth}{!}{\begin{tabular}{@{}lccccccc@{}}
        \toprule
        Method & CD $\downarrow$ & TTCE$_2$ $\downarrow$ & TTCE$_3$ $\downarrow$ & TTCE$_4$ $\downarrow$ & CTC$_2$ $\downarrow$ & CTC$_3$ $\downarrow$ & CTC$_4$ $\downarrow$ \\
        \midrule
        \multicolumn{8}{@{}l}{\textbf{KITTI Odometry} (sequences 08 to 10, 671 windows)} \\
        GEM~\cite{wu2026gem} & \cfirst\textbf{0.117} & 1.020 & 1.687 & 2.604 & \csecond 0.168 & \csecond 0.180 & \csecond 0.188 \\
        ST3DCNN~\cite{mersch2021st3dcnn} & 0.158 & \csecond 0.869 & \csecond 1.502 & \csecond 2.392 & 0.175 & 0.185 & 0.190 \\
        \ours{} w/o boxes & \csecond 0.130 & \cfirst\textbf{0.641} & \cfirst\textbf{1.133} & \cfirst\textbf{1.869} & \cfirst\textbf{0.167} & \cfirst\textbf{0.179} & \cfirst\textbf{0.186} \\
        \midrule
        \multicolumn{8}{@{}l}{\textbf{KITTI-360} (sequences 0000 and 0002, 248 windows)} \\
        GEM~\cite{wu2026gem} & 0.160 & 2.908 & 4.430 & 6.018 & 0.176 & 0.186 & 0.192 \\
        ST3DCNN~\cite{mersch2021st3dcnn} & 0.162 & 1.769 & 2.818 & 3.940 & 0.176 & 0.186 & 0.191 \\
        \ours{} w/o boxes & \csecond 0.117 & \csecond 0.385 & \cfirst\textbf{0.650} & \cfirst\textbf{1.061} & \csecond 0.160 & \csecond 0.173 & \csecond 0.182 \\
        \ours & \cfirst\textbf{0.115} & \cfirst\textbf{0.375} & \csecond 0.669 & \csecond 1.166 & \cfirst\textbf{0.158} & \cfirst\textbf{0.171} & \cfirst\textbf{0.181} \\
        \bottomrule
    \end{tabular}}
    \vspace{-3mm}
\end{table}
On KITTI Odometry, GEM has the lowest CD, but our predictions lead on every other metric.
Our CTC is close to that of GEM but slightly lower at every frame gap, and our TTCE is markedly lower than those of both baselines.
In contrast, on KITTI-360, our model outperforms GEM and ST3DCNN on every metric, with or without boxes.
With boxes, TTCE$_3$ and TTCE$_4$ rise slightly but remain far below those of GEM and ST3DCNN.
Outside their training dataset, GEM and ST3DCNN have a much higher TTCE on KITTI-360 than on KITTI Odometry, and GEM also has a much higher CD.
There, the scenes predicted by GEM move much less than the given ego motion, which likely raises its TTCE.
Our model instead remains competitive on KITTI Odometry, which is also outside its training dataset.

\subsection{Sensitivity to the Inputs}
\label{sec:app_pred_inputs}
\begin{table}[t]
    \centering
    \caption{Sensitivity of prediction to its inputs on KITTI-360 with a 0.5\,s horizon. Dyn.\ is defined as in \tabref{tab:app_pred_horizon}.}
    \label{tab:app_pred_robust}
    \scriptsize
    \setlength{\tabcolsep}{3pt}
    \renewcommand{\arraystretch}{0.94}
    \resizebox{\linewidth}{!}{\begin{tabular}{@{}lcccc@{}}
        \toprule
        Setting & CD $\downarrow$ & $F_{0.1}$ $\uparrow$ & H-F1 $\uparrow$ & Dyn.\ $\uparrow$ \\
        \midrule
        \multicolumn{5}{@{}l}{\textbf{References and warp} (250 windows)} \\
        1 reference & 0.112 & 0.499 & 0.758 & 0.417 \\
        2 references & 0.096 & 0.594 & 0.926 & \cfirst\textbf{0.503} \\
        3 references (default) & \cfirst\textbf{0.087} & \cfirst\textbf{0.649} & \cfirst\textbf{0.975} & \csecond 0.502 \\
        5 references & 0.094 & 0.606 & \csecond 0.974 & 0.459 \\
        3 references, every other row masked & \csecond 0.092 & \csecond 0.622 & 0.970 & 0.474 \\
        Ego motion only, no box warp & \cfirst\textbf{0.087} & \cfirst\textbf{0.649} & \cfirst\textbf{0.975} & 0.456 \\
        \midrule
        \multicolumn{5}{@{}l}{\textbf{Noise} (250 windows)} \\
        No noise & \cfirst\textbf{0.087} & \cfirst\textbf{0.649} & \cfirst\textbf{0.975} & \cfirst\textbf{0.502} \\
        Pose, 0.25\,m and $1^\circ$ & 0.101 & 0.560 & 0.973 & 0.430 \\
        Pose, 0.5\,m and $2^\circ$ & 0.115 & 0.476 & 0.969 & 0.375 \\
        Pose, 1.0\,m and $5^\circ$ & 0.130 & 0.382 & 0.962 & 0.322 \\
        Boxes, 0.25\,m and $1^\circ$ & \csecond 0.089 & \csecond 0.637 & \csecond 0.974 & \csecond 0.479 \\
        Boxes, 0.5\,m and $2^\circ$ & 0.090 & 0.626 & \csecond 0.974 & 0.435 \\
        Boxes, 1.0\,m and $5^\circ$ & 0.091 & 0.621 & 0.973 & 0.414 \\
        \bottomrule
    \end{tabular}}
    \vspace{-3mm}
\end{table}

In our method, prediction receives reference scans, ego poses, and object boxes.
If the world model relies on each of these inputs, changing or perturbing it should affect the prediction.
\tabref{tab:app_pred_robust} tests this at the 0.5\,s horizon by varying the references, the warp, and the noise in the poses and boxes.

\noindent\textbf{References and warp.}
Prediction is most accurate with three reference scans, the number used in training.
Fewer references raise CD and lower $F_{0.1}$ and H-F1, as less of the scene and its past motion has been observed.
Five references also reduce accuracy, as such longer contexts fall outside training.
Masking every other row of the references degrades most metrics less than using fewer references, as the autoencoder still recovers most returns on the masked rows from the remaining ones (Appendix~\ref{sec:app_rays_unseen}).
In the setting with ego motion only (\tabref{tab:app_pred_robust}), the boxes are still given as layout, but the warp ignores them and moves every reference cell by the ego motion.
This leaves CD, $F_{0.1}$, and H-F1 unchanged at the reported precision.
However, this lowers the recall of points on moving objects, as the warped references then keep moving objects at their observed positions.
Therefore, the world model draws on both the observed history in the references and the box motion in the warp.

\noindent\textbf{Noise in poses and boxes.}
We also add Gaussian noise to the ego poses and boxes that prediction receives.
The listed translations and rotations are standard deviations of the horizontal position and yaw, drawn independently for the pose of every frame or for every box of every frame.
Box noise perturbs both the box warp and the box attributes of the layout.
Growing pose noise raises CD and lowers $F_{0.1}$ and H-F1.
In contrast, box noise mainly lowers the recall of points on moving objects.
This reflects the roles of the two inputs, as the ego poses place the whole scene while the boxes move only the objects inside them.
Therefore, prediction follows the given ego and object motion, and its accuracy depends most on accurate ego poses.

\begin{figure*}[t]
    \centering
    \includegraphics[width=\textwidth]{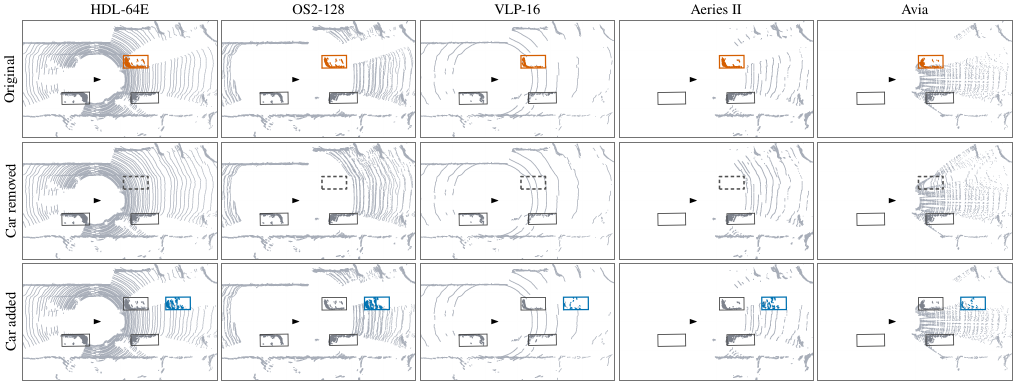}
    \caption{\textbf{Layout editing on KITTI-360 across scan patterns.}
    All rows show the first scan predicted 0.6\,s after the last of three observed scans, with the same initial noise.
    Orange, dashed, and blue boxes mark the car to remove, its former location, and the added car.}
    \label{fig:layout_edit}
    \vspace{-3mm}
\end{figure*}
\subsection{World Model Ablation}
\label{sec:app_pred_ablation}
\begin{table}[t]
    \centering
    \caption{KITTI-360 world model ablation in \tabref{tab:wm_ablation} by frame and by window, with frames 0.1\,s apart. Heading change runs from the last observed to the fifth predicted frame, and each heading group contains about a third of the 250 windows. The static CTC is also given for the measured scans.}
    \label{tab:app_wm_why}
    \scriptsize
    \setlength{\tabcolsep}{2.6pt}
    \renewcommand{\arraystretch}{0.86}
    \resizebox{\linewidth}{!}{\begin{tabular}{@{}lcccccc@{}}
        \toprule
        & Full & w/o Warp & w/o $\mathcal{L}_{\mathrm{ray}}$ & Full & w/o Warp & w/o $\mathcal{L}_{\mathrm{ray}}$ \\
        \midrule
        \multicolumn{7}{@{}l}{\textbf{Per predicted frame}} \\
        Frame & \multicolumn{3}{c}{CD $\downarrow$} & \multicolumn{3}{c}{H-F1 $\uparrow$} \\
        \cmidrule(lr){2-4}\cmidrule(l){5-7}
        1 & \cfirst\textbf{0.077} & \csecond 0.078 & \csecond 0.078 & \cfirst\textbf{0.977} & \csecond 0.976 & 0.972 \\
        2 & \cfirst\textbf{0.084} & 0.087 & \csecond 0.085 & \cfirst\textbf{0.976} & \csecond 0.975 & 0.970 \\
        3 & \cfirst\textbf{0.088} & 0.093 & \csecond 0.090 & \cfirst\textbf{0.975} & \csecond 0.974 & 0.969 \\
        4 & \cfirst\textbf{0.091} & 0.097 & \csecond 0.092 & \cfirst\textbf{0.974} & \csecond 0.973 & 0.968 \\
        5 & \cfirst\textbf{0.094} & 0.101 & \csecond 0.095 & \cfirst\textbf{0.974} & \csecond 0.972 & 0.967 \\
        \midrule
        \multicolumn{7}{@{}l}{\textbf{By heading change of the ego vehicle}} \\
        Heading & \multicolumn{3}{c}{CD $\downarrow$} & \multicolumn{3}{c}{dMAE $\downarrow$} \\
        \cmidrule(lr){2-4}\cmidrule(l){5-7}
        0.0--0.4$^\circ$ & \cfirst\textbf{0.085} & 0.088 & \csecond 0.087 & \cfirst\textbf{0.498} & 0.534 & \csecond 0.501 \\
        0.4--1.8$^\circ$ & \cfirst\textbf{0.087} & 0.091 & \csecond 0.088 & \cfirst\textbf{0.547} & 0.590 & \csecond 0.550 \\
        1.8--18.9$^\circ$ & \cfirst\textbf{0.088} & 0.094 & \csecond 0.090 & \cfirst\textbf{0.529} & 0.609 & \csecond 0.530 \\
        \midrule
        \multicolumn{7}{@{}l}{\textbf{Distance between static points of predicted frames}} \\
        Gap & \multicolumn{3}{c}{CTC, static $\downarrow$} & \multicolumn{3}{c}{Measured scans (Ref.)} \\
        \cmidrule(lr){2-4}\cmidrule(l){5-7}
        1 & \cfirst\textbf{0.099} & \csecond 0.100 & \csecond 0.100 & \multicolumn{3}{c}{0.074} \\
        2 & \cfirst\textbf{0.109} & 0.112 & \csecond 0.111 & \multicolumn{3}{c}{0.085} \\
        3 & \cfirst\textbf{0.117} & 0.120 & \csecond 0.118 & \multicolumn{3}{c}{0.093} \\
        4 & \cfirst\textbf{0.122} & 0.127 & \csecond 0.124 & \multicolumn{3}{c}{0.099} \\
        \bottomrule
    \end{tabular}}
    \vspace{-3mm}
\end{table}
\tabref{tab:wm_ablation} scores each variant over all predicted frames and windows.
To show where each component matters, \tabref{tab:app_wm_why} breaks down this ablation by predicted frame, ego heading change, and frame gap.
Without the warp, the CD gap to the full model grows with the predicted frame.
The CTC gap on static points is also largest for the frames farthest apart.
The CD and dMAE gaps without the warp are also largest at the largest heading changes.
The warp carries the reference content along the given motion.
Without it, the model must infer this motion, which becomes harder as the scene moves farther from the references.
Without $\mathcal{L}_{\mathrm{ray}}$, the CD and CTC gaps stay nearly constant across frames and frame gaps.
However, this variant has the lowest H-F1 at every predicted frame.
As $\mathcal{L}_{\mathrm{ray}}$ supervises the decoded returns during training, it improves which rays return.
Therefore, the warp matters most for later frames and larger turns, whereas $\mathcal{L}_{\mathrm{ray}}$ improves hit decisions at every frame.

\subsection{Layout Editing}
\label{sec:app_pred_edit}

If prediction follows its boxes, editing them should change only the edited objects and the regions they occlude.
\figref{fig:layout_edit} illustrates this on a KITTI-360 scene with the observed scans and the initial noise held fixed.
Both edits change only the boxes of the five predicted frames.
Each predicted latent is decoded along the fixed beam grid (HDL-64E) and with the scan patterns of the four HeLiPR sensors.
Removing the oncoming car from the layout eliminates its returns in all five patterns.
Ground returns then continue through the region it occluded.
Conversely, adding a box behind this car in the original layout makes a new car appear inside that box in every pattern.
Outside the edited boxes and their occlusion shadows, the scan stays close to the original.
As the edit changes the latent before decoding, every pattern observes the same edited scene.
Therefore, in this example, one layout edit yields an alternative future that all five scan patterns observe consistently.

\section{Downstream Tasks}
\label{sec:app_downstream}
\label{sec:downstream}
Our model can produce scans for sensors that it never saw in training.
This section tests whether such scans help perception models on these sensors.
It gives the full results and setups of the three tasks in Sec.~\ref{sec:hetero_lpr}.
For all three tasks, we also train on autoencoded source scans.
These need a measured source scan at every training frame and serve as a reference built from measured scans.
Predicted scans need only three observed frames and the motion, so the same model can also predict sequences for edited layouts that no sensor recorded (Appendix~\ref{sec:app_pred_edit}).
In all three tasks, the predicted scans lie at most 2.5\,s after the last observed frame.

\subsection{Heterogeneous Place Recognition}
\label{sec:app_down_lpr}
Place recognition across \acp{LiDAR} locates a query scan of one sensor among the database scans of another.
If our predicted scans reproduce the patterns of the query sensors, training on them should improve this task.

\noindent\textbf{Setup.}
We train HeLiOS~\cite{jung2025helios} on nuScenes data and evaluate it on the HeLiPR Roundabout and Town sequences.
The database contains OS2-128 scans, and Aeries~II, Avia, and VLP-16 scans serve as queries.
Besides the raw nuScenes scans, we train on two sets decoded along the rays of the four HeLiPR sensors.
These patterns come from a HeLiPR sequence other than Roundabout and Town, and no point or label of HeLiPR is used in training.
\afterpage{\begin{table}[t]
    \centering
    \caption{Place recognition with HeLiOS on HeLiPR, reported as R@1, maximum F1 score (F1), and area under the precision--recall curve (AUC) in \%. Mean averages the three query sensors.}
    \label{tab:down_lpr}
    \scriptsize
    \setlength{\tabcolsep}{2.2pt}
    \renewcommand{\arraystretch}{0.94}
    \resizebox{\linewidth}{!}{\begin{tabular}{@{}llcccc@{\hspace{2.2pt}}c@{\hspace{1.6pt}}!{\color{black}\vrule}@{\hspace{1.6pt}}c@{\hspace{2.2pt}}cccc@{}}
        \toprule
        & & \multicolumn{4}{c}{Roundabout} & \multicolumn{2}{@{}c@{}}{} & \multicolumn{4}{c}{Town} \\
        \cmidrule(lr){3-6}\cmidrule(l){9-12}
        Metric & Training data & Aeries~II & Avia & VLP-16 & Mean & \multicolumn{2}{@{}c@{}}{} & Aeries~II & Avia & VLP-16 & Mean \\
        \midrule
        \multirow{3}{*}{R@1} & Raw nuScenes & 56.5 & \csecond 27.6 & 69.0 & 51.0 & & & 64.2 & 46.7 & 58.6 & 56.5 \\
        & Our autoencoder & \cfirst\textbf{72.2} & \cfirst\textbf{53.5} & \cfirst\textbf{80.5} & \cfirst\textbf{68.7} & & & \cfirst\textbf{67.9} & \cfirst\textbf{60.6} & \cfirst\textbf{67.7} & \cfirst\textbf{65.4} \\
        & Our world model & \csecond 71.3 & \cfirst\textbf{53.5} & \csecond 78.5 & \csecond 67.8 & & & \csecond 65.4 & \csecond 56.5 & \csecond 65.1 & \csecond 62.3 \\
        \midrule
        \multirow{3}{*}{F1} & Raw nuScenes & 52.2 & 23.7 & 64.3 & 46.8 & & & 59.9 & 42.7 & 54.0 & 52.2 \\
        & Our autoencoder & \cfirst\textbf{67.5} & \csecond 46.3 & \cfirst\textbf{77.0} & \cfirst\textbf{63.6} & & & \cfirst\textbf{62.6} & \cfirst\textbf{55.1} & \cfirst\textbf{62.4} & \cfirst\textbf{60.0} \\
        & Our world model & \csecond 66.4 & \cfirst\textbf{47.4} & \csecond 74.7 & \csecond 62.8 & & & \csecond 60.1 & \csecond 51.7 & \csecond 60.0 & \csecond 57.3 \\
        \midrule
        \multirow{3}{*}{AUC} & Raw nuScenes & 40.2 & 10.2 & 56.5 & 35.6 & & & \csecond 48.1 & 26.1 & 38.3 & 37.5 \\
        & Our autoencoder & \cfirst\textbf{60.2} & \csecond 32.2 & \cfirst\textbf{71.5} & \cfirst\textbf{54.6} & & & \cfirst\textbf{51.7} & \cfirst\textbf{42.0} & \cfirst\textbf{49.2} & \cfirst\textbf{47.6} \\
        & Our world model & \csecond 59.3 & \cfirst\textbf{33.9} & \csecond 68.7 & \csecond 54.0 & & & 46.3 & \csecond 36.3 & \csecond 45.1 & \csecond 42.6 \\
        \bottomrule
    \end{tabular}}
    \vspace{-3mm}
\end{table}}%

Each keyframe is used for training in all four patterns, with positive pairs following the HeLiOS overlap rule across patterns.
Autoencoded scans are the raw nuScenes scans encoded and decoded by our autoencoder.
Predicted scans come from our world model, conditioned on three measured frames and the recorded ego and object motion.
The raw and autoencoded sets use 12,014 keyframes of the training scenes sampled every 4\,m.
The predicted set replaces each keyframe with its prediction from a window in the same scene.
The 1,168 keyframes near scene starts lack three preceding keyframes, leaving 10,846 predicted frames from 4,293 windows.
We use the released HeLiOS code's architecture, input preprocessing, and loss~\cite{jung2025helios}.
Every model trains for 1,800 updates with a batch of 2,048 using Adam, a learning rate of $5\times10^{-4}$ reduced tenfold after 1,440 updates, and weight decay of $10^{-4}$.
Evaluation follows the official Roundabout and Town protocols and script, including temporal exclusion for queries from the database sequence.
On Roundabout, the database contains Roundabout01 OS2-128 scans, and queries comprise Roundabout01 to Roundabout03 Aeries~II, Avia, and VLP-16 scans.
Town uses Town01 to Town03 in the same way.
A correct database match has official query overlap above 0.5.

\begin{figure}[t]
    \centering
    \includegraphics[width=\linewidth]{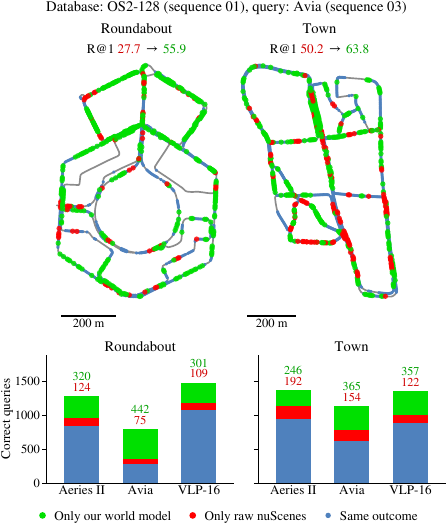}
    \caption{\textbf{Where predicted scans change the top-1 match} on the Roundabout03 and Town03 queries.
    On the Avia query trajectory (gray), green and red dots mark queries matched correctly only with predicted scans or only with raw nuScenes scans, and blue dots mark queries that both models match correctly or both miss.
    Bars count queries matched correctly by both (blue) or by only one (green, red) for each query sensor.}
    \label{fig:app_down_lpr}
    \vspace{-3mm}
    \vspace{4mm}
    \centering
    \includegraphics[width=\linewidth]{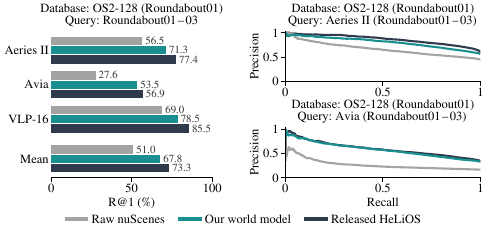}
    \caption{\textbf{Place recognition on Roundabout.}
    The released HeLiOS model is trained on measured HeLiPR scans of other sequences.
    Values of the other two models match \tabref{tab:down_lpr}.}
    \label{fig:lpr}
    \vspace{-2mm}
\end{figure}%
\noindent\textbf{Gains for unseen query sensors.}
In \tabref{tab:down_lpr}, both decoded sets raise the mean R@1, F1, and AUC over the raw scans on both sequences.
With both sets, R@1 rises for every query sensor, and the Avia queries gain the most.
We attribute this to the difference between the non-repetitive Avia pattern and the rotating pattern of the nuScenes sensor.
Training on the predicted scans, which cover about 10\% fewer keyframes, nearly matches training on the autoencoded ones on Roundabout and scores lower on Town.
\figref{fig:app_down_lpr} compares the top-1 matches of the models trained on raw and predicted scans.
On these queries, training on predicted scans corrects more errors than it introduces for every query sensor on both sequences.
For Avia, these corrected matches are distributed along the entire Roundabout03 and Town03 routes.
\figref{fig:lpr} compares these two models with the released HeLiOS weights, which are trained on measured scans of all four HeLiPR sensors in other sequences.
On Roundabout, training on predicted scans achieves 92.5\% of the mean R@1 of the released weights.
Furthermore, its precision--recall curve for Avia queries nearly overlaps that of the released weights.
For these reasons, this subsection shows that our world model provides training data for sensors whose measured scans are absent from HeLiOS training.
\afterpage{\begin{table}[t]
    \centering
    \caption{Moving IoU and AP (\%) of 4DMOS on the HeLiMOS test split. Raw SemanticKITTI denotes the released weights. The other rows fine-tune them on the same windows, with raw scans, autoencoded measured frames, or our predictions.}
    \label{tab:down_mos}
    \scriptsize
    \setlength{\tabcolsep}{3pt}
    \renewcommand{\arraystretch}{0.94}
    \begin{tabularx}{\linewidth}{@{}ll*{5}{>{\centering\arraybackslash}X}@{}}
        \toprule
        Metric & Training data & \mbox{Aeries~II} & Avia & \mbox{OS2-128} & \mbox{VLP-16} & Mean \\
        \midrule
        \multirow{4}{*}{IoU, online} & Raw SemanticKITTI & 56.3 & 53.1 & 65.3 & 24.3 & 49.7 \\
        & + Raw sequences & 42.5 & 52.5 & 53.8 & 22.5 & 42.8 \\
        & + Our autoencoder & \csecond 72.0 & \csecond 58.7 & \csecond 71.8 & \cfirst\textbf{31.5} & \csecond 58.5 \\
        & + Our world model & \cfirst\textbf{73.1} & \cfirst\textbf{59.4} & \cfirst\textbf{73.3} & \csecond 31.2 & \cfirst\textbf{59.3} \\
        \midrule
        \multirow{4}{*}{IoU, delayed} & Raw SemanticKITTI & 60.2 & 60.9 & 72.0 & 34.9 & 57.0 \\
        & + Raw sequences & 45.4 & 60.0 & 61.5 & 30.0 & 49.2 \\
        & + Our autoencoder & \cfirst\textbf{79.4} & \csecond 65.8 & \cfirst\textbf{78.4} & \cfirst\textbf{40.6} & \cfirst\textbf{66.1} \\
        & + Our world model & \csecond 78.4 & \cfirst\textbf{66.0} & \csecond 78.1 & \csecond 40.5 & \csecond 65.8 \\
        \midrule
        \multirow{4}{*}{AP, online} & Raw SemanticKITTI & 68.7 & 76.5 & 83.8 & 46.9 & 69.0 \\
        & + Raw sequences & 47.7 & 76.5 & 76.0 & 35.7 & 59.0 \\
        & + Our autoencoder & \csecond 89.9 & \cfirst\textbf{79.8} & \csecond 89.0 & \cfirst\textbf{52.3} & \cfirst\textbf{77.7} \\
        & + Our world model & \cfirst\textbf{90.3} & \csecond 79.4 & \cfirst\textbf{89.5} & \csecond 50.1 & \csecond 77.3 \\
        \midrule
        \multirow{4}{*}{AP, delayed} & Raw SemanticKITTI & \csecond 77.0 & 79.9 & 89.1 & 56.1 & 75.5 \\
        & + Raw sequences & 54.0 & 78.2 & 78.9 & 45.2 & 64.1 \\
        & + Our autoencoder & \cfirst\textbf{92.4} & \cfirst\textbf{82.2} & \csecond 91.2 & \cfirst\textbf{59.4} & \cfirst\textbf{81.3} \\
        & + Our world model & \cfirst\textbf{92.4} & \csecond 81.9 & \cfirst\textbf{91.7} & \csecond 58.7 & \csecond 81.2 \\
        \bottomrule
    \end{tabularx}
    \vspace{-3mm}
\end{table}}%

\subsection{Moving Object Segmentation}
\label{sec:app_down_mos}
Moving object segmentation labels each point of a scan as moving or static.
Labeled moving points are costly to collect for every new sensor.
If our predicted sequences carry realistic motion in the patterns of new sensors, fine-tuning a segmenter on them should improve it on these sensors.

\noindent\textbf{Setup.}
We test this on HeLiMOS~\cite{lim2024helimos}, which labels moving points in scans of the four HeLiPR sensors.
We start from the released 4DMOS~\cite{mersch2022receding} weights, which are trained on SemanticKITTI~\cite{behley2019semantickitti} HDL-64E scans.
We then fine-tune them on SemanticKITTI together with sequences generated by our KITTI-360 and nuScenes models.
Each of these sequences has ten frames, as 4DMOS takes ten scans as input.
The frames are sampled at 10\,Hz on KITTI-360 and at 12\,Hz on nuScenes.
Our autoencoder reconstructs the first five frames from measured scans.
Our world model predicts the last five from the three latest measured frames and the recorded ego and object motion.
All ten frames are decoded with the scan pattern of each HeLiPR sensor, as in Appendix~\ref{sec:app_down_lpr}.
Points inside boxes that move faster than 0.5\,m/s are labeled as moving.
The fine-tuning data have two sources.
The first is the training set of the released 4DMOS weights, 19,130 labeled SemanticKITTI scans.
The second is our sequences, built from 922 non-overlapping windows of KITTI-360 sequences 0003 to 0007, 0009, and 0010 and 1,400 windows of the 700 nuScenes training scenes.
Their predicted frames in the four patterns give 46,440 training scans.
Fine-tuning samples uniformly from the scans of both sources and runs for 48k updates with Adam, a learning rate of $10^{-4}$ with cosine decay, and a batch of 16.
For comparison, raw sequences use the measured scans of the same windows in their native patterns, labeled and sampled by the same rule as our autoencoder.
We evaluate on the 514 labeled test scans of each HeLiMOS sensor.

\begin{figure}[t]
    \centering
    \includegraphics[width=\linewidth]{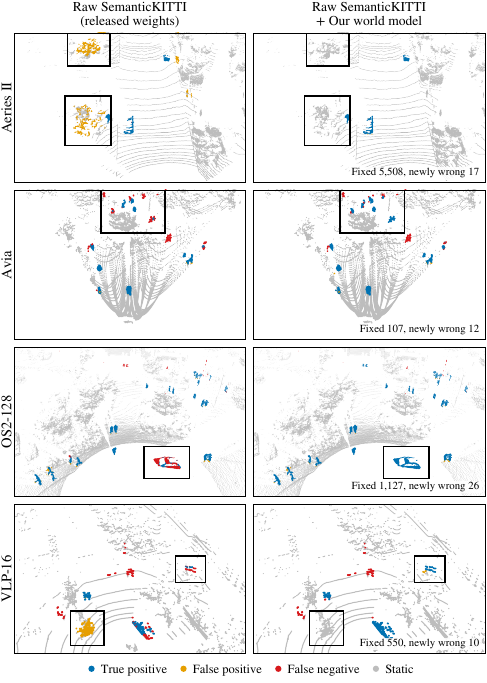}
    \caption{\textbf{Moving object segmentation on HeLiMOS.}
    One test scan per sensor in the delayed mode.
    Boxes mark moving objects segmented more accurately after fine-tuning on our predictions, and static regions that only the released weights mark as moving.
    The numbers count the points that this fine-tuning corrects and those it newly misclassifies.
    Gray points are static or unlabeled.}
    \label{fig:app_down_mos}
    \vspace{-3mm}
\end{figure}%
\noindent\textbf{Gains on the HeLiMOS sensors.}
In \tabref{tab:down_mos}, both models fine-tuned on our sequences exceed the released weights in moving IoU and AP on all four sensors, with the largest gains on Aeries~II.
When the last five frames are autoencoded from measured scans instead of predicted, the scores are similar, so our predicted frames are as useful for training as autoencoded measured ones.
Fine-tuning on these raw sequences lowers every mean score below the released weights.
Thus, the additional windows do not help by themselves, and the gain appears when they are decoded with the HeLiPR patterns.
With our sequences, scores rise in both the online mode, which uses the current and past scans, and the delayed mode, which also uses later scans.
AP, which does not depend on a decision threshold, also rises, so the gains do not come from the choice of threshold.
\figref{fig:app_down_mos} shows test scans with many moving objects.
In the marked regions, the released weights miss many points of moving objects or mark static regions as moving, and the model fine-tuned on our predictions classifies most of these points correctly.
In every scan, it corrects far more points than it newly misclassifies.
For these reasons, this subsection shows that our world model provides training data that improves moving object segmentation on unseen sensors.

\subsection{Object Detection}
\label{sec:app_down_det}
\afterpage{\begin{table}[t]
    \centering
    \caption{Car detection with Voxel R-CNN on PandarGT test scans, for cars with at least five points within 56\,m. 3D and BEV AP are AP$_{\mathrm{R40}}$ (\%) at an IoU of 0.5, and nuScenes mAP (\%) is the devkit car AP averaged over thresholds of 0.5, 1, 2, and 4\,m on the center distance.}
    \label{tab:down_det}
    \scriptsize
    \setlength{\tabcolsep}{2.2pt}
    \renewcommand{\arraystretch}{0.94}
    \resizebox{\linewidth}{!}{\begin{tabular}{@{}lccccc@{}}
        \toprule
        & \multicolumn{3}{c}{3D AP $\uparrow$} & BEV AP $\uparrow$ & nuScenes mAP $\uparrow$ \\
        \cmidrule(lr){2-4}\cmidrule(lr){5-5}\cmidrule(l){6-6}
        Training data & All & 0--30\,m & 30--50\,m & All & All \\
        \midrule
        \multicolumn{6}{@{}l}{\textit{Zero-shot}} \\
        Raw nuScenes & 10.5 & 8.9 & 13.2 & 13.1 & 2.1 \\
        Ray interpolation & 48.9 & 79.0 & 29.0 & 49.3 & 43.0 \\
        Our autoencoder & \cfirst\textbf{56.1} & \cfirst\textbf{86.2} & \cfirst\textbf{42.9} & \cfirst\textbf{58.8} & \cfirst\textbf{52.9} \\
        Our world model & \csecond 53.5 & \csecond 83.9 & \csecond 38.1 & \csecond 56.1 & \csecond 48.8 \\
        \midrule
        \multicolumn{6}{@{}l}{\textit{Self-training on unlabeled PandarGT scans}} \\
        Raw nuScenes & 30.0 & 23.2 & 36.5 & 31.4 & 23.2 \\
        Ray interpolation & 68.7 & \cfirst\textbf{88.7} & 56.0 & 71.5 & 66.2 \\
        Our autoencoder & \cfirst\textbf{72.6} & \cfirst\textbf{88.7} & \cfirst\textbf{66.9} & \cfirst\textbf{75.4} & \cfirst\textbf{70.4} \\
        Our world model & \csecond 71.9 & \csecond 88.2 & \csecond 64.6 & \csecond 75.0 & \csecond 69.4 \\
        \bottomrule
    \end{tabular}}
    \vspace{-3mm}
\end{table}\begin{figure}[t]
    \centering
    \includegraphics[width=\linewidth]{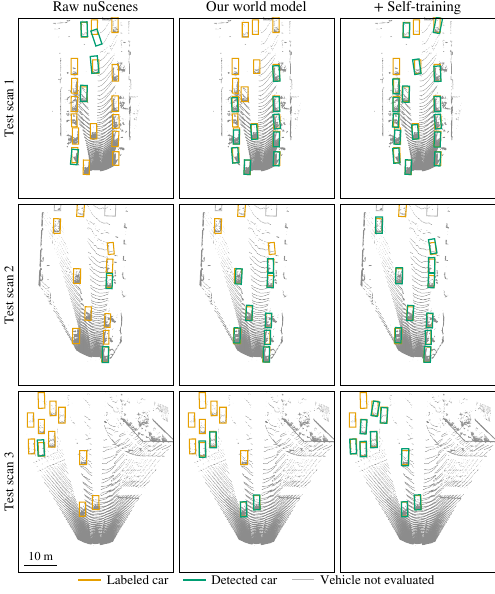}
    \caption{\textbf{Detections on PandarGT test scans} in \ac{BEV}, 0 to 56\,m ahead of the sensor.
    Columns show detectors trained on raw nuScenes scans, on our predicted scans, and on our predicted scans with self-training.
    Detections scoring at least 0.2 are drawn.}
    \label{fig:app_down_det}
    \vspace{-3mm}
\end{figure}
\begin{figure*}[t]
    \centering
    \includegraphics[width=\textwidth]{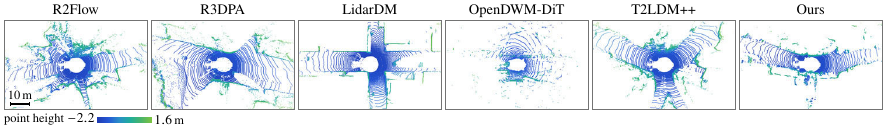}
    \caption{\textbf{Unconditional samples on KITTI-360}, one per method with a fixed seed.
    Heights are relative to the sensor.}
    \label{fig:app_gen_samples}
    \vspace{-3mm}
\end{figure*}

\begin{figure*}[t]
    \centering
    \includegraphics[width=\textwidth]{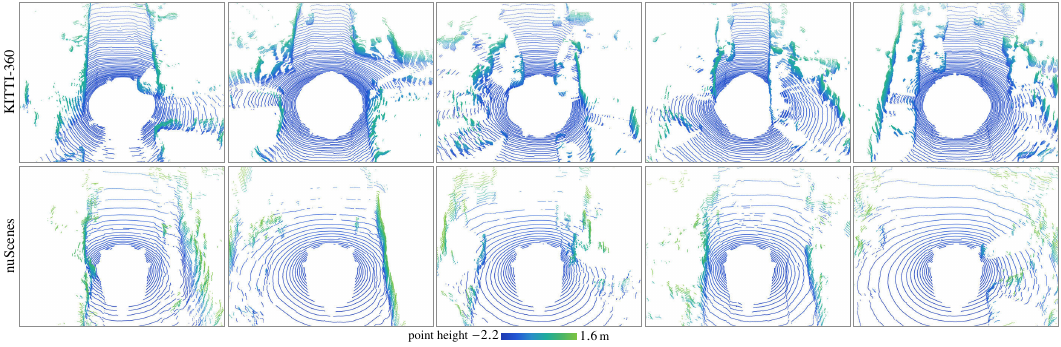}
    \caption{\textbf{More unconditional samples} from our model, viewed from behind and above the sensor.}
    \label{fig:app_uncond_more}
    \vspace{-3mm}
\end{figure*}}%
Object detection locates objects such as cars in a scan.
Labeled boxes are costly to collect for every new sensor.
If our decoder reproduces the scan pattern of a new sensor, a detector trained on the decoded scans should transfer to that sensor without its labels.

\noindent\textbf{Setup.}
We test this on PandaSet~\cite{xiao2021pandaset}, whose PandarGT faces forward and has a scan pattern unlike those of all our training sensors.
We train Voxel R-CNN~\cite{deng2021voxelrcnn} detectors on 9,723 nuScenes training keyframes, taking every third keyframe of each scene.
We evaluate them on the PandarGT scans of 35 PandaSet test scenes, taking every third of the 103 scenes in ID order.
The other 68 scenes are used without their labels, only for the PandarGT rays and for self-training.
A CenterHead~\cite{yin2021center} proposes the boxes of each detector from predicted object centers.
All detectors are trained from scratch with the same settings for every training set, for eight epochs with Adam, a one-cycle schedule with a maximum learning rate of 0.003, and a batch of 4.
Every detector takes only point coordinates, and we exclude intensity from the raw nuScenes scans because the \acp{LiDAR} used in nuScenes and PandaSet report intensity on different scales.
Besides the raw nuScenes scans, we train on three sets whose scans follow the PandarGT rays.
These rays are the measured ray directions of 256 PandaSet scans outside the test scenes, without their points or labels, repeated at six headings $60^\circ$ apart to cover all directions.
Related methods for adaptation across sensors complete source scans or reduce their beams to match the target sensor~\cite{yi2021complete,wei2022lidar}.
As a simple method of this kind, ray interpolation fills each PandarGT ray from the two nuScenes rows around its elevation in the same $0.35^\circ$ azimuth bin.
Because linear interpolation between rows blurs depth edges~\cite{kwon2022iln}, the two ranges are interpolated only when both rows return and agree within 0.5\,m or 5\%.
Otherwise, the ray takes the range of the nearer row within half a row spacing or returns nothing.
Autoencoded scans encode the raw scans and decode them along the PandarGT rays.
Predicted scans come from our nuScenes world model, which predicts the 9,023 of these keyframes that have three preceding keyframes in their scene, conditioned on the recorded ego and object motion.
All training scans are leveled like the PandaSet test scans.
Only PandaSet Car cuboids count as ground truth.
Car detections that match a cuboid of another vehicle class, such as a pickup truck, truck, or bus, count as neither true nor false positives.

\noindent\textbf{Transfer without PandaSet labels.}
In the upper block of \tabref{tab:down_det}, the detector trained on the raw nuScenes scans detects few cars.
Cars in the raw nuScenes scans contain far fewer points than in PandarGT scans, with medians of 13 against 421 points at 20 to 40\,m.
Autoencoded scans raise both 3D and BEV AP far above the raw scans and above interpolation to the same rays.
The gain over interpolation is largest at 30 to 50\,m and also appears under the nuScenes metric.
Predicted scans come close to the autoencoded scans and also exceed interpolation.
Thus, our predicted scans train a detector for a new sensor nearly as well as nuScenes scans autoencoded along its rays.

\noindent\textbf{Self-training on unlabeled PandarGT scans.}
A zero-shot detector can also label PandarGT scans without using their annotations, so a better zero-shot detector should provide a better initialization for adaptation without labels.
We use the zero-shot detector of each training set to label the 5,440 PandarGT scans of the other 68 scenes with the cars it detects.
We then fine-tune a copy of it for eight epochs on these labels together with its own training scans.
In the lower block of \tabref{tab:down_det}, self-training raises every detector, and the detectors that start from our predicted or autoencoded scans remain the most accurate overall.
Their gain over interpolation lies beyond 30\,m.
We attribute this gain to the pseudo-labels, as the detector from autoencoded scans detects many more distant cars in these scans than the detector from interpolated scans.
The detector from predicted scans nearly matches the one from autoencoded scans, although it is trained on fewer keyframes.

\figref{fig:app_down_det} shows detections on three test scans.
The detector trained on raw scans detects only a few cars in these scans, and the detector trained on predicted scans detects most nearby cars.
Self-training recovers most of the cars that this detector missed.
For these reasons, this subsection shows that our world model provides detection training data for a sensor without labeled scans and a better initialization for adaptation without labels.

\section{More Qualitative Results}
\label{sec:app_qual}

\noindent\textbf{Unconditional generation.}
\figref{fig:app_gen_samples} shows one KITTI-360 sample from several generators of \tabref{tab:uncond_generation} in \ac{BEV}.
\figref{fig:app_uncond_more} adds more samples from our model on both datasets.

\begin{figure*}[t]
    \centering
    \includegraphics[width=0.92\textwidth]{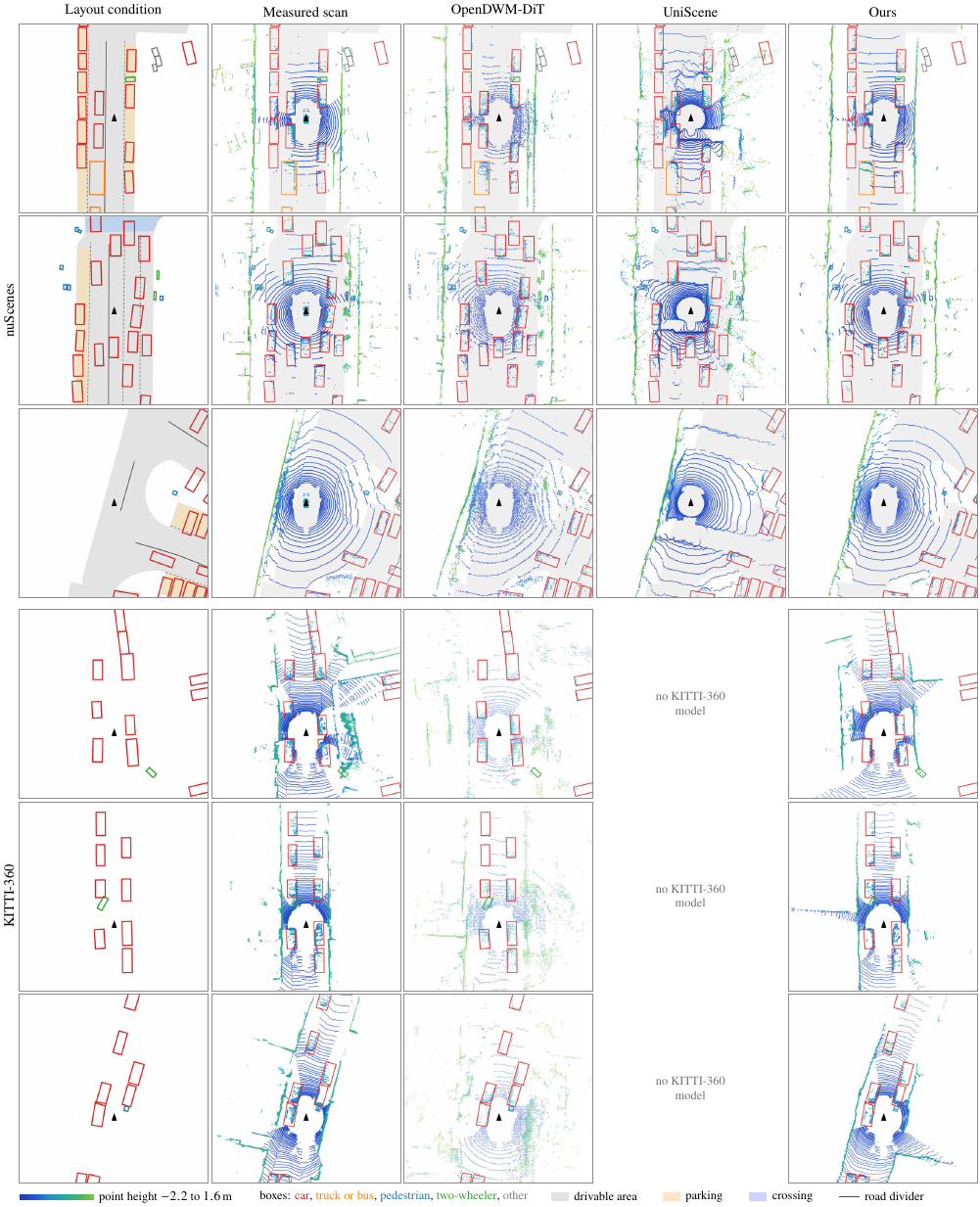}
    \caption{\textbf{Layout-conditioned generation} without observed scans.
    All panels show the boxes with the sensor facing up, and the nuScenes panels also show the map areas.}
    \label{fig:cond_gen}
    \label{fig:app_cond_more}
    \vspace{-3mm}
\end{figure*}

\noindent\textbf{Layout-conditioned generation.}
\figref{fig:cond_gen} shows generation from the layouts of \tabref{tab:object_generation} without observed scans.
The nuScenes layouts also contain an HD map, and only nuScenes includes UniScene, which has no KITTI-360 model.

\begin{figure*}[t]
    \centering
    \includegraphics[width=\textwidth]{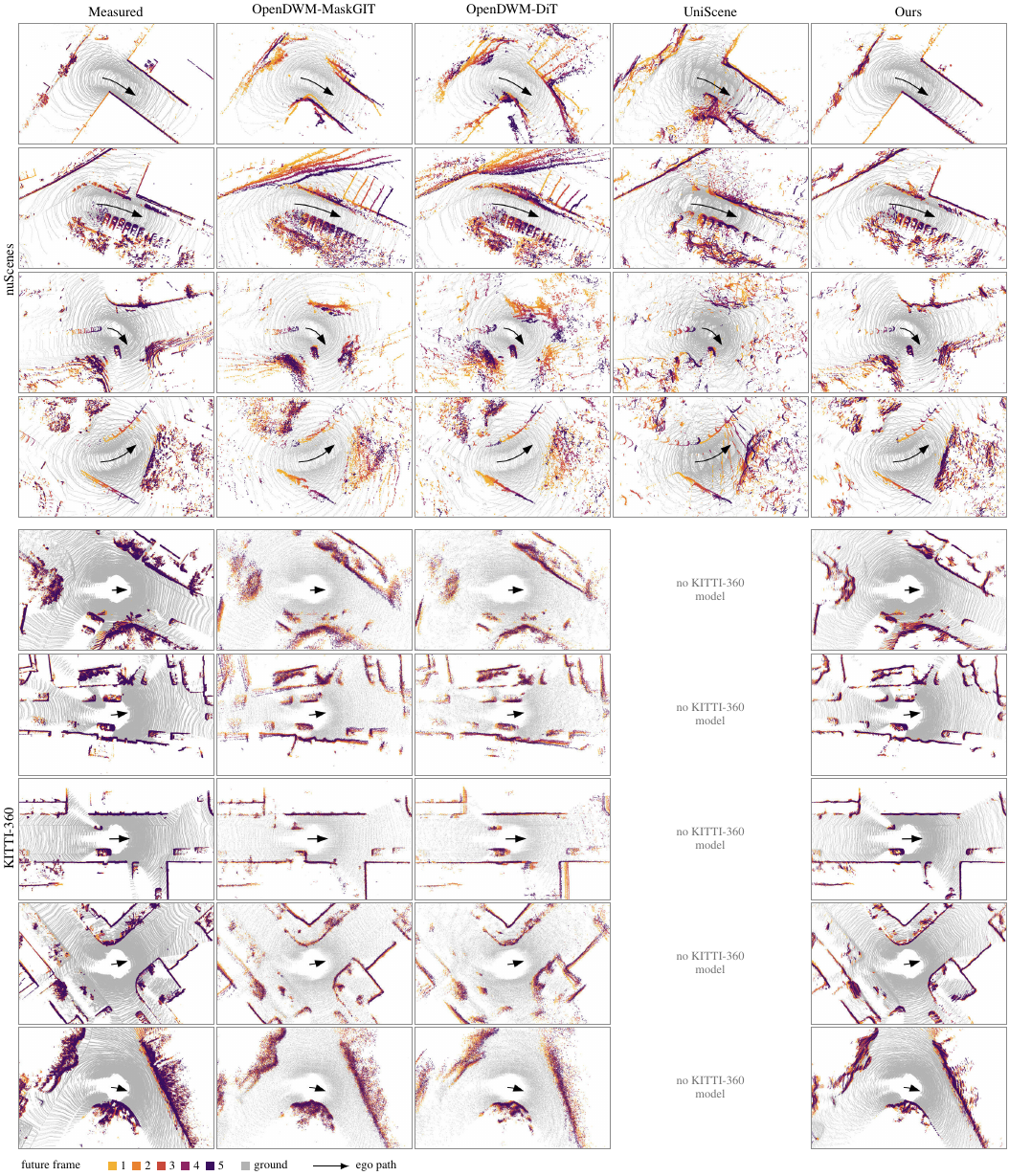}
    \caption{\textbf{More sequence predictions}, as in \figref{fig:seq_pred}.
    UniScene generates from layouts alone and has no KITTI-360 model.}
    \label{fig:app_pred_more}
    \vspace{-3mm}
\end{figure*}

\begin{figure*}[t]
    \centering
    \includegraphics[width=\textwidth]{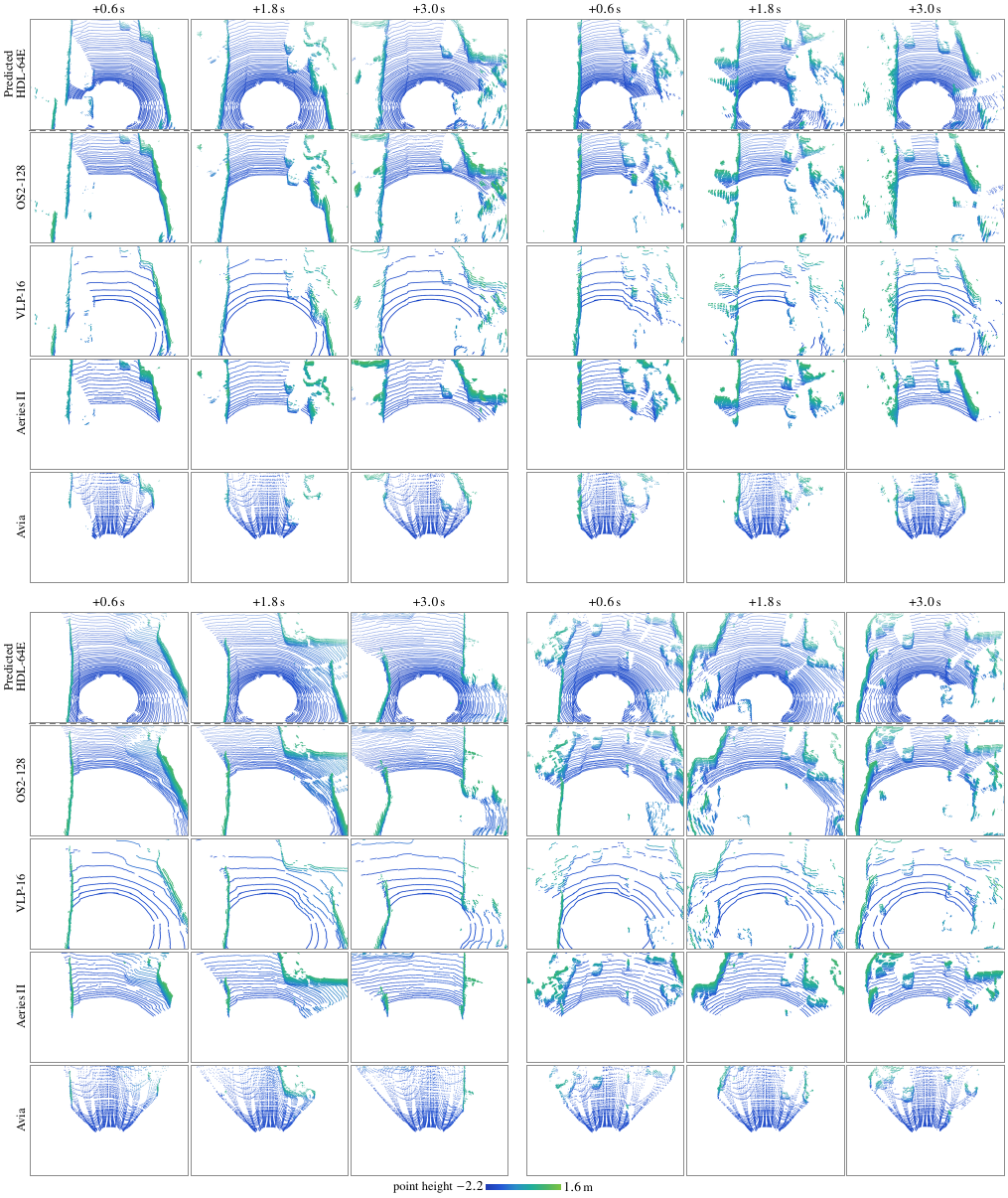}
    \caption{\textbf{Predicted sequences for unseen sensors.}
    Each block is one KITTI-360 test window with three observed and five predicted frames 0.6\,s apart.
    Each column decodes one predicted latent along the fixed beam grid (HDL-64E) in the first row and with the four HeLiPR patterns below, seen from behind each sensor.}
    \label{fig:app_pred_hetero}
    \vspace{-3mm}
\end{figure*}

\begin{figure*}[t]
    \centering
    \includegraphics[width=\textwidth]{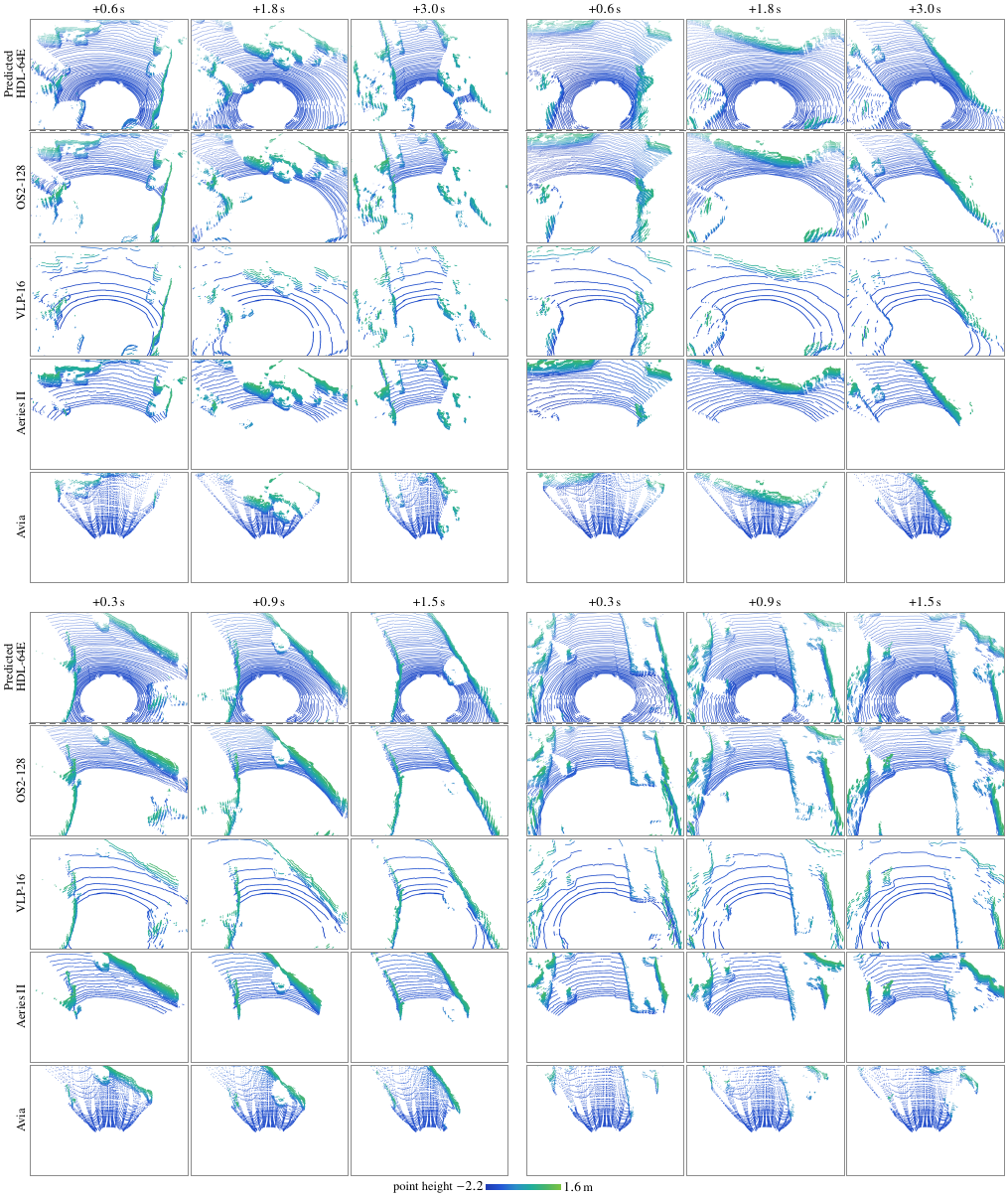}
    \caption{\textbf{More predicted sequences for unseen sensors} on four other KITTI-360 test windows, shown as in \figref{fig:app_pred_hetero}.
    The bottom two use frames 0.3\,s apart.}
    \label{fig:app_pred_hetero2}
    \vspace{-3mm}
\end{figure*}

\noindent\textbf{Prediction.}
\figref{fig:app_pred_more} shows more predicted sequences in the format of \figref{fig:seq_pred}, with frames 0.5\,s apart on nuScenes and 0.1\,s apart on KITTI-360.
Figs.~\ref{fig:app_pred_hetero} and~\ref{fig:app_pred_hetero2} decode predicted KITTI-360 sequences with the four HeLiPR scan patterns, as \figref{fig:transfer} does for generated scenes.
As all patterns decode the same predicted latents, they show the same scene in every frame.

\section{Limitations}
\label{sec:app_limits}
Our model has three main limitations.
First, the decoder reads returns only along rays from the sensor origin of the latent, so it cannot place a sensor at another mounting position.
A different position does not simply shift the rays. It also changes which surfaces are visible, whereas the latent describes the first surface along each direction from the original origin.
Likewise, a mesh or field rendered from another position misses or only extrapolates the surfaces the measured scans did not observe.
Even at the original origin, rendering from occupancy, fused meshes, or a learned render head gives higher depth errors than our decoder in \tabref{tab:occ_oracle}.
A new mounting position resembles a change of ego pose, which our world model already handles in prediction by warping the latent and generating the target frame.
Treating the mount offset in this way could extend our model to other mounting positions.

Second, the decoder predicts no return intensity.
Each manufacturer reports intensity on its own scale and calibration, so the intensities of different sensors give no common target for the shared latent.
The downstream models in Appendix~\ref{sec:app_downstream} use only point geometry, and the decoded scans improve all three tasks without intensity.
Heads for individual sensors on the shared latent could add intensity for sensors with measured data.

Third, the decoder rarely completes large regions without observations, and it can leave most rays there without a return.
Even so, the returns it does predict on masked rays remain accurate, with the lowest depth errors among the compared methods for every mask in Tables~\ref{tab:occ_oracle} and~\ref{tab:app_masked_repr}.
Supervising the decoder on rays masked from the encoder input, which our scan pattern augmentation does not do, could train it to complete such regions.
Furthermore, the KITTI-360 aggregated scans label as misses the regions above the sensor's vertical field of view, such as the upper parts of nearby structures, so the decoder may predict no return there.
Most of these directions lie outside all quantitative KITTI-360 evaluations.
Training with sensors of wider vertical fields of view would help supervise these directions.

\clearpage
\twocolumn

\end{document}